\documentclass{article} 
\usepackage{iclr2027_conference_rebuttal,times}
\iclrfinalcopy

\usepackage{amsmath,amsfonts,bm}

\def\eqref#1{equation~\ref{#1}}

\def\1{\bm{1}}

\DeclareMathAlphabet{\mathsfit}{\encodingdefault}{\sfdefault}{m}{sl}
\SetMathAlphabet{\mathsfit}{bold}{\encodingdefault}{\sfdefault}{bx}{n}

\usepackage{hyperref}
\usepackage{url}
\usepackage{tcolorbox}
\usepackage{enumitem}
\usepackage{xcolor}

\usepackage{xcolor,tikz}
\usepackage{amsmath}
\usepackage{amssymb}
\usepackage{mathtools}
\usepackage{amsthm}
\theoremstyle{plain}
\newtheorem{theorem}{Theorem}[section]
\newtheorem{proposition}[theorem]{Proposition}

\theoremstyle{definition}

\theoremstyle{remark}

\usepackage[percent]{overpic}
\usepackage{graphicx}
\usepackage{tikz}
\usetikzlibrary{positioning}
\usepackage{booktabs} 
\usepackage{multirow}
\usepackage{xcolor}

\let\cite\citep

\title{Understanding Clipping in \\Zeroth-order Optimization}

\author{
Saket Gollapudi$^{1}$ \quad
Elisa Bertino$^{2}$ \quad
Sewoong Oh$^{1}$
\\
$^{1}$University of Washington, Seattle
\quad
$^{2}$Purdue University
\thanks{\texttt{saket312@cs.washington.edu},
\texttt{bertino@purdue.edu},
\texttt{sewoong@cs.washington.edu}}
}

\renewcommand{\eqref}[1]{Eq.~(\ref{#1})}

\newcommand{\sg}[1]{\textcolor{black}{#1}}

\makeatletter
\def\thanks#1{\protected@xdef\@thanks{\@thanks \protect\footnotetext{#1}}}
\makeatother

\begin{document}

\maketitle
\begin{abstract} 
    Zeroth-order optimization has emerged as a resource-efficient alternative to Stochastic Gradient Descent (SGD) when fine-tuning LLMs, thanks to the inherent low dimensionality of the loss landscape. It relies on a two-point estimate of the gradient that only accesses the function value and not the gradient, bypassing the resource-heavy backpropagation. Recently, it was accidentally discovered that per-sample clipping can improve the performance of the trained model, while running a differentially private version of zeroth-order optimization \cite{zhang2024dpzero} with a very large privacy parameter $\epsilon$ (corresponding to very little privacy). In this paper, we systematically investigate this phenomenon and demonstrate that ($i$) the optimal choice of learning rate $\eta$ is inversely proportional to the clipping threshold $c$, satisfying $\eta c =$ constant; ($ii$) clipping drives the zeroth-order optimization to a different solution with higher test accuracy; and ($iii$) this phenomenon is only observed with per-sample clipped zeroth-order optimization and not other zeroth-order alternatives that also provide robust estimates of the gradient. Further analyses with artificially injected label noise reveals that clipping leads to more robust predictions, which stems from the implicit down weighting of hard-to-fit training examples. Our code is available 
    \href{https://github.com/SewoongLab/ClipZero}{here} \footnote{\url{https://github.com/SewoongLab/ClipZero}. 
}

\end{abstract} 

\section{Introduction}

Zeroth-Order (ZO) optimization methods allow one to minimize an objective function without access to its gradients. This is necessary ($i$) when the learner has access to the function evaluation only, as in finding adversarial examples for a black-box model with API access \cite{chen2017zoo,tu2019autozoom,liu2020primer}, or ($ii$) when computing the gradient is prohibitively expensive, as in back-propagation on a model with tens of billions of parameters \cite{malladi2023fine,zhang2024dpzero}. Compared to First-Order (FO) methods that use gradient information to iteratively update the model, a major bottleneck in adopting ZO optimization in practice is the dimension-dependent convergence rate. ZO methods typically search over all dimensions, wasting a lot of iterations on unnecessary directions with small improvements, resulting in a factor of $d$  slower convergence compared to FO counterparts for  $d$-dimensional optimization problems  \cite{duchi2015optimal}. 

The first practical breakthrough, therefore, of ZO methods came in a relatively lower dimensional problem of adversarial attacks, where the optimization is over the input pixels of, say, the Inception-v3 network \cite{szegedy2016rethinking}: $d=299\times 299\times 3 = 268,203$ \cite{chen2017zoo}. Empirically, ZO methods proved to be successful in finding  adversarial examples for a black-box model where we only have API access to the function output  \cite{ye2018hessian}. The next breakthrough in applying ZO methods came in LLM fine-tuning \cite{malladi2023fine,zhang2024dpzero}. \citet{malladi2023fine} theoretically showed that the convergence of ZO methods only scales as the effective rank of the Hessian matrix instead of the number of parameters to be optimized. Empirical evidence suggests that in LLM fine-tuning, the effective rank is significantly smaller than the number of parameters. Since ZO methods do not require back-propagation, which is memory heavy, this makes ZO fine-tuning a memory-efficient alternative for LLM fine-tuning. 

\noindent{\bf Problem formulation.} 
Consider fine-tuning  model weights $w\in{\mathbb R}^d$ given $n$ samples,  
$
    \min_{w\in{\mathbb R}^d} (1/n) \sum_{i=1}^n f_i(w)$, where $f_i$ is the loss on the $i$-th fine-tuning example. Given access to a FO oracle,  Stochastic Gradient Descent (SGD) updates the model as per mini-batch gradient $g(w_t)=(1/|B_t|)\sum_{i\in B_t} \nabla f_i (w_t)$ with learning rate $\eta$:  
$
w_{t+1} \gets  w_t - \eta\ g(w_t)$. {\em Zeroth-Order SGD (ZO-SGD)} only accesses the ZO oracle that can only evaluate the function value, and commonly uses a two-point estimate of the mini-batch gradient:  
\begin{eqnarray}
     g_\lambda(w_t) \;=\; \frac{1}{|B_t|}\sum_{i\in B_t} \frac{f_i(w_t+\lambda u_t)-f_i(w_t - \lambda u_t)}{2\lambda} u_t  \;, \label{eq:zo}    
\end{eqnarray}
where $\lambda>0$ is the {\em smoothing parameter} and  $u_t\in{\mathbb R}^{d}$ is a random vector with i.i.d.~standard Gaussian entries. This approach can also be applied to SVRG and momentum methods \cite{liu2018zeroth, chen2019zo}. While empirically measuring the performance of ZO-SGD under differential privacy,   \cite{zhang2024dpzero} first reported that adding {\em per-sample} clipping to ZO-SGD, which we call {\em clipped ZO-SGD}, can improve performance of LLM fine-tuning. The clipped estimate is: 
\begin{eqnarray}
   g_{\lambda,c}(w_t) \; =\; \frac{1}{|B_t|} \sum_{i\in B_t} \mathrm{Clip}_c\Big( \frac{f_i(w_t+\lambda u_t)-f_i(w_t - \lambda u_t)}{2\lambda} \Big)  u_t \;,
   \label{eq:clip}    
\end{eqnarray} 
where $\mathrm{Clip}_c:{\mathbb R}\to{\mathbb R}$ is defined as $\mathrm{Clip}_c(a)=a$ if $|a|\leq c$ and $\mathrm{Clip}_c(a)=\mathrm{sign}(a)\cdot c$ otherwise.

\noindent{\bf Contributions.} 
We systematically investigate the gain of clipping in ZO-SGD in four fine-tuning tasks of SNLI \cite{bowman2015large}, MNLI \cite{williams2018broad}, SST-2 \cite{socher2013recursive}, and Trec \cite{li2002learning} on two pretrained base models of RoBERTa \cite{liu2019roberta} and OPT \cite{zhang2022opt}. 

In all these settings, we identify a {\em clipping trade-off}: optimal choice of the learning rate inversely scales with the clipping threshold $c$ (Section~\ref{sec:3tradeoff}) for a wide range of $c$ that we refer to as \sg{\em the clipping-dominated regime}.
More importantly, we demonstrate that the models trained in this regime of clipping converge to a different solution compared to those trained with no clipping (Section~\ref{sec:basin}). This new type of solution achieves higher accuracy but suffers from higher loss in both train and test data, indicating that clipping introduces implicit bias for higher accuracy predictions (Section~\ref{sec:3bias}).
We show that this is unique to per-sample clipping, and other robust estimates of the gradient estimate, including batch clipping and Winsorized mean, do not exhibit this phenomenon (Section~\ref{sec:robust}). 
We also show that this trade-off phenomenon is unique to zeroth-order optimization 
(Section~\ref{sec:first_order}).

To investigate the origin of higher accuracy, we analyze the calibration of resulting predictions and demonstrate that clipping leads to overconfident and miscalibrated predictions (Section~\ref{sec:calibrate}). Further analysis with artificially injected label noise reveals that clipping effectively down weights samples that are harder to fit. This in turn makes the model robust to training examples that are mislabeled or ambiguous, leading to higher accuracy (Section~\ref{sec:noise}).
This hypothesis is further  supported by our results showing  that clipped ZO-SGD performs similarly with data pruned ZO-SGD, where we select a subset of data with smallest loss to retrain a model from scratch with unclipped ZO-SGD (Section~\ref{sec:prune}).

\section{Preliminary} 
\label{sec:pre} 

The two-point estimate in \eqref{eq:zo} converges to an unbiased estimate of the gradient  as $\lambda\to0$, i.e.,~ $\lim_{\lambda\to0}{\mathbb E}_{u_t}[g_{\lambda}(w_t)] =  (1/|B_t|)\sum _{i\in B_t} \nabla f_i(w_t)$ \cite{duchi2015optimal}. This acts as a surrogate for the mini-batch gradient and justifies the use of ZO methods. With per-sample clipping, the gradient estimate is no longer unbiased, and the clipped ZO-SGD is optimizing for a different objective function. It is challenging to interpret what that objective function is, but one can reason about the gradient estimate as a {\em weighted average} of the individual gradients, with the weight depending only on the gradient norm, $\|\nabla f_i(w)\|$, of the individual gradient.   A proof is in Appendix~\ref{app:proof}.

\begin{proposition}
\label{propo:clippedgradient}
For the per-sample clipped gradient estimate  defined in \eqref{eq:clip}, 
\begin{eqnarray}
    \label{eq:clippedgradient} 
    \lim_{\lambda \to 0} \;{\mathbb E}_u \big[ \,g_{\lambda,c}(w)\,\big] \; = \; \frac{1}{|B_t|} \sum_{i\in B_t} h_c(\|\nabla f_i(w) \|) \, \nabla f_i(w)\;,
\end{eqnarray}
where $ h_c(\|\nabla f_i(w) \|) = 2 \Phi(c/\|\nabla f_i(w)\|)-1$, and $\Phi(\cdot)$ is the standard normal CDF. \end{proposition}

When the clipping threshold $c$ is large enough such that no clipping happens, then this recovers the (asymptotically) unbiased gradient estimate of \eqref{eq:zo}. We refer to this as an {\em unclipped} regime. When $c$ is small enough that clipping always happens, the weight function converges to $\sqrt{2/\pi} (c/\|\nabla f_i(w)\|)$  and the gradient estimate simplifies to a voting algorithm where each sample contributes equally regardless of its gradient magnitude. The resulting model update in this case is 
\begin{eqnarray}
   {\mathbb E}_{u_t} \big[\, w_{t+1}|w_t
   \,\big] \;\gets \; w_t \;+ \; \sqrt\frac2\pi\,c\,\eta \, \frac{1}{|B_t|} \sum_{i\in B_t}  \frac{\nabla f_i(w_t)}{\|\nabla f_i(w_t)\|}  \;. 
   \label{eq:fullyclipped}
\end{eqnarray} 
We refer to this as a \sg{{\em clipping-dominated}} regime. 
The clipped ZO-SGD can be interpreted as a smooth  interpolation between these two extremes. We investigate which regime dominates for real data, depending on the choice of the clipping threshold $c$, and what the consequences are for the fine-tuned LLMs.

\section{The trade-off between clipping threshold and learning rate} 
\label{sec:tradeoff}

In the \sg{clipping-dominated} regime, \eqref{eq:fullyclipped} suggests that the clipping threshold $c$ and the learning rate $\eta$ can be seamlessly exchanged as long as we keep $c\eta$ fixed. To test this idea, we perform a geometric grid search over $(c,\eta)$ and track the resulting model performance. We demonstrate that ($i$) clipped ZO-SGD typically operates as if it is either in the \sg{clipping-dominated} regime or the unclipped regime; ($ii$) the model converges to different solutions depending on the regime; ($iii$) in the \sg{clipping-dominated} regime, the model suffers higher loss but achieves better accuracy \sg{(both test and train)}; ($iv$) this phenomenon is unique to per-sample clipping; and ($v$) also unique to ZO-SGD. A survey of related work is provided in App.~\ref{app:related} and a detailed experimental set-up in  App.~\ref{sec:experimental_setup}. 

\subsection{The {\em clipping trade-off} at $\eta\,c=\text{constant}$}  \label{sec:3tradeoff}
\begin{figure}[h]
\begin{center}
    \begin{tikzpicture}
        \node[anchor=south west, inner sep=0, outer sep=0]
            (image) at (0,0)
    {\includegraphics[width=0.31\columnwidth]{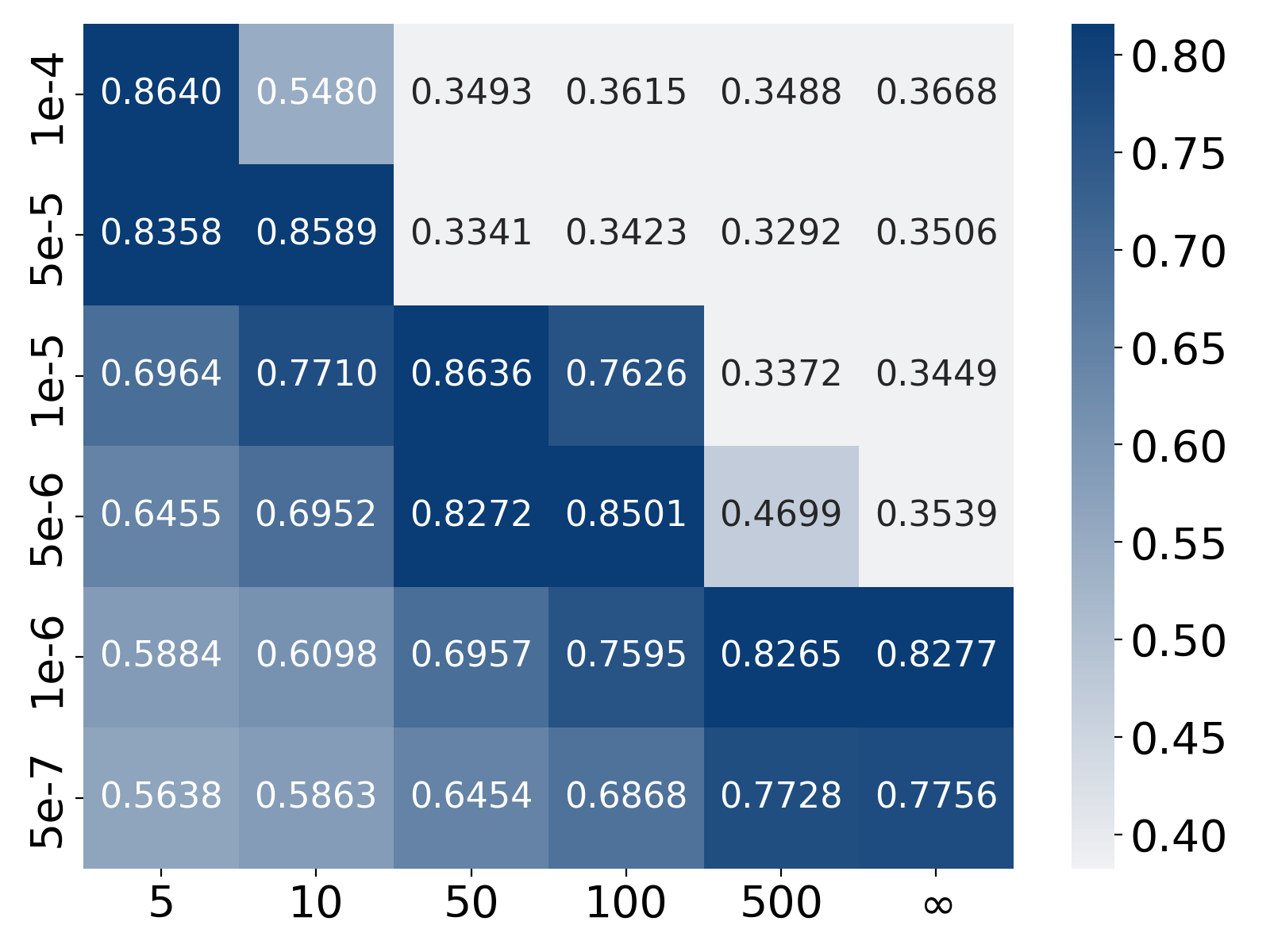}};
        \begin{scope}[x={(image.south east)}, y={(image.north west)}]
            \draw[orange, line width=1.5pt] (0.185,0.968) rectangle (0.071,0.835);
        \end{scope} 
\begin{scope}[x={(image.south east)}, y={(image.north west)}]      \draw[orange, line width=1.5pt] (0.307,0.687) rectangle(0.194,0.820); \end{scope} 
\begin{scope}[x={(image.south east)}, y={(image.north west)}]      \draw[orange, line width=1.5pt] (0.428,0.537) rectangle(0.315,0.674); \end{scope} 
\begin{scope}[x={(image.south east)}, y={(image.north west)}]      \draw[orange, line width=1.5pt] (0.553,0.392) rectangle(0.438,0.525); \end{scope} 
\begin{scope}[x={(image.south east)}, y={(image.north west)}]      \draw[orange, line width=1.5pt] (0.675,0.242) rectangle(0.560,0.375); \end{scope} 
\begin{scope}[x={(image.south east)}, y={(image.north west)}]      \draw[orange, line width=1.5pt] (0.793,0.242) rectangle(0.684,0.375); \end{scope}
    \end{tikzpicture}
    \put(-93,96){\small test accuracy}
    \put(-109,-7){\small clipping threshold $c$}
    \put(-133,47){\rotatebox[origin=c]{90}{\small learning rate $\eta$}}
        \quad
\includegraphics[width=0.31\columnwidth]{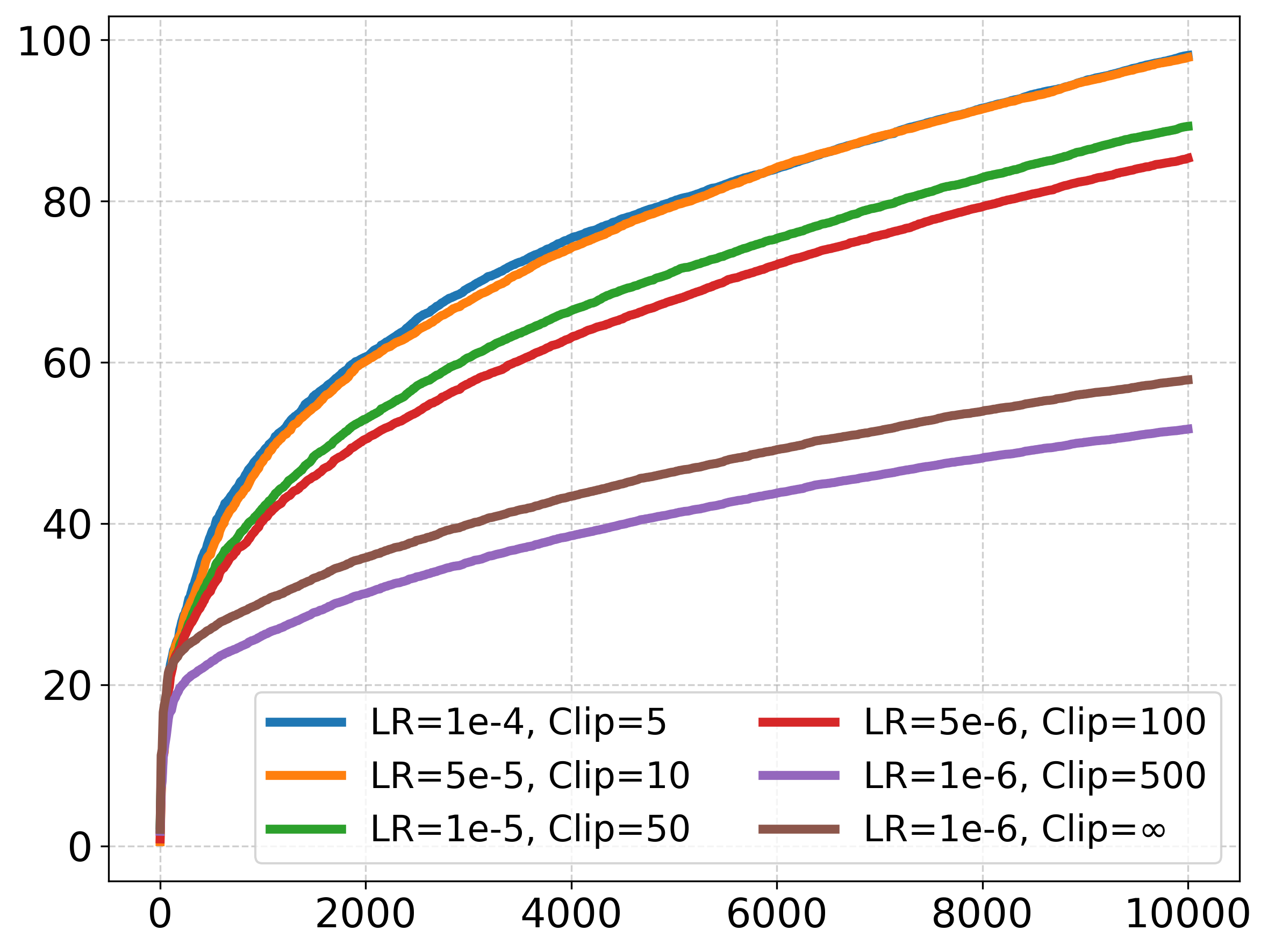} 
\put(-103,-8){\small fine-tuning iterations $t$}
\put(-130,45){\rotatebox[origin=c]{90}{\small distance $\|w_t-w_0\|$}}
    \quad
\includegraphics[width=0.31\columnwidth]{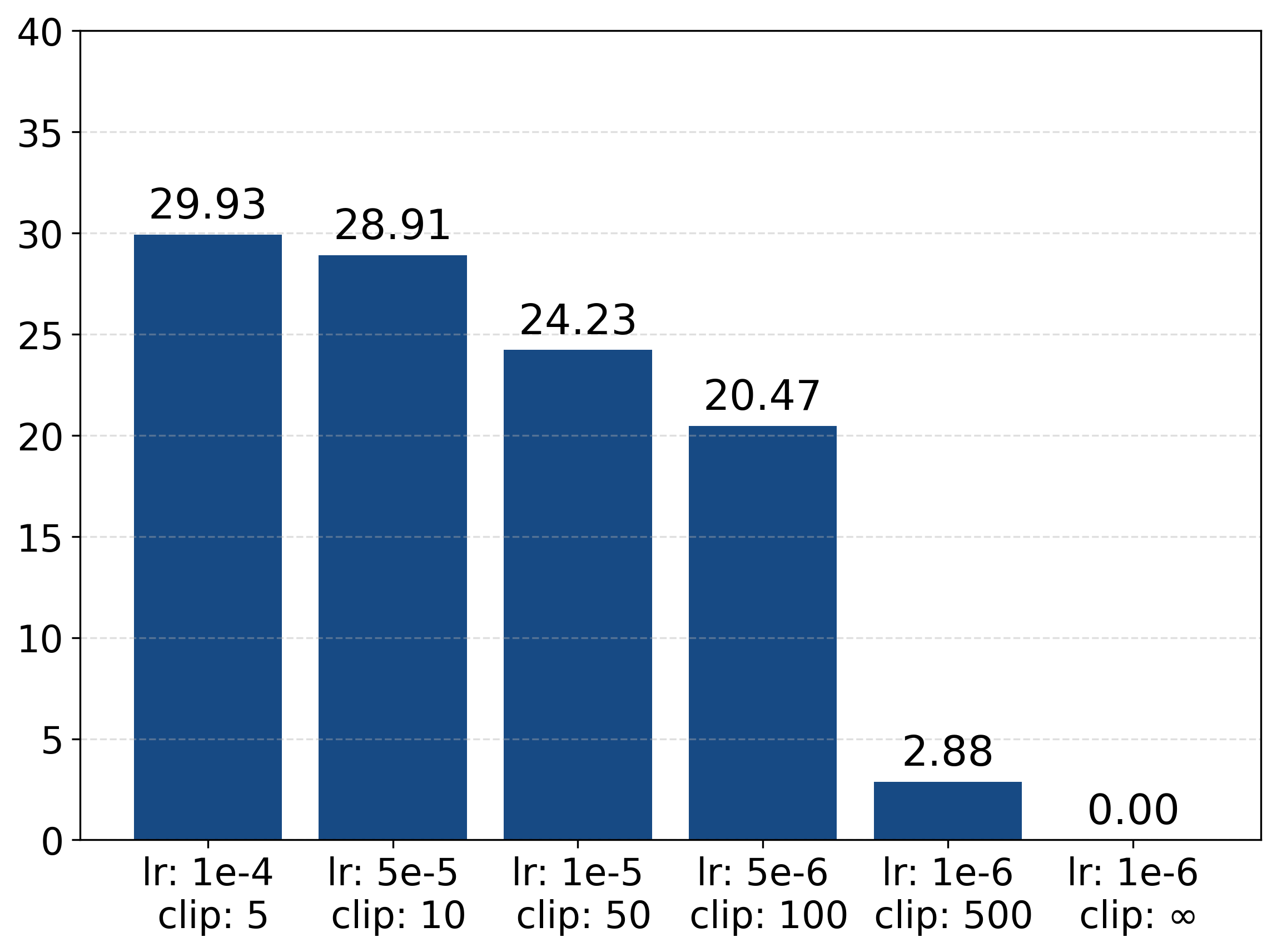}
    \put(-111,96){\small avg. \# of clipped gradients}
\end{center}
\caption{(left) Sweeping through the learning rate $\eta$ and the clipping threshold $c$, we observe that the optimal learning rate for each clipping threshold lies on the curve $\eta\times c = 5\times 10^{-4}$. 
Each cell shows the test accuracy averaged over four models fine-tuned from RoBERTa \cite{liu2019roberta} on SNLI \cite{bowman2015large}, MNLI \cite{williams2018broad}, SST-2 \cite{socher2013recursive} and Trec \cite{li2002learning}, respectively.  $c=\infty$ is the no clipping baseline. (middle) For those orange cells, tracking how fast SNLI fine-tuned weights $w_t$ move away from the pre-trained checkpoint of RoBERTa $w_0$, we observe two distinct behaviors depending on the clipping threshold: $c\leq 100$ and $c\geq 500$. (right) For the same orange cells, the average number of clipped examples per iteration (batch size = $64$) exhibits similar two distinct regimes. \sg{Note that the clip count in Figure~\ref{fig:grid_avg_roberta} (right) is averaged over all $10{,}000$ steps. As shown in Appendix~\ref{app:roberta-expand} (Figure~\ref{fig:snli_small_c_clips}, left), during the first $\sim 4{,}000$ steps the vast majority of gradients exceed $c$ and are clipped, driving the sharp separation in weight trajectories. As the model adapts, fewer gradients exceed the threshold, pulling the $10{,}000$-step average down even when $c$ is reduced to $0.1$ (Figures~\ref{fig:snli_small_c}--\ref{fig:snli_small_c_clips}).}} 
\label{fig:grid_avg_roberta}
\end{figure}

To understand how the optimal learning rate $\eta$ scales with the clipping threshold $c$, we sweep through a range of parameters and measure the test accuracy averaged over four fine-tuned models trained on four separate tasks 
(Figure~\ref{fig:grid_avg_roberta}, left). 
We make two important observations: $(i)$ the optimal learning rate $\eta^*(c)$ for each clipping threshold $c$ lies on the line satisfying $\eta c = 0.0005$ highlighted by orange outline in the diagonal except for $c=\infty$; and ($ii$) the best average test accuracy of $0.8640$ is achieved with clipping, specifically with $c=5$. This phenomenon, which we call the {\em clipping trade-off}, is quite robust. The same behavior is demonstrated on each individual fine-tuning task (Figures~\ref{fig:roberta_snli_grid}-\ref{fig:roberta_trec_grid_off} in the appendix) and across two model families (OPT family in Figure~\ref{fig:sst_opt} in the appendix) in LM fine-tuning. 

One explanation of this inverse relation might be that the clipping in \eqref{eq:clip} is effectively scaling down the gradient estimate $g_\lambda$ in \eqref{eq:zo} and the learning rates need to compensate for that in order to reach the similar optimal point. 
This would mean that  models trained with the same effective speed of  $\eta\, c=5\times 10^{-4}$ should be similar. In Figure \ref{fig:grid_avg_roberta} (middle), we 
track the Euclidean distance from the initial pre-trained checkpoint: $\|w_t-w_0\|$. 
This reveals two distinct paths, possibly towards two distinct basins, indicating that clipping is not just rescaling the gradient estimates but rather driving the model towards potentially a better minima, one that can only be reached with clipping.

This dichotomy suggests that 
when $c\leq 100$, the dynamics of clipped ZO-SGD is driven by the \sg{\em{clipping-dominated} regime}, and when $c>100$, it is driven by the {\em unclipped regime}. 
In the \sg{clipping-dominated} regime, significantly more samples are clipped as shown in Figure~\ref{fig:grid_avg_roberta} (right), and hence the optimal learning rate is inversely proportional to the clipping threshold $c$ as suggested by  \eqref{eq:fullyclipped}. 
This phenomenon of convergence to two distinct solutions is also consistent with the fact that the dynamics of \eqref{eq:fullyclipped} in the \sg{clipping-dominated} regime is distinct from the gradient descent (ZO or FO), due to each sample being weighted differently. 
We provide detailed analyses of the two types of solutions in Section~\ref{sec:basin} and the statistics of clipped gradients in 
Appendix~\ref{sec:aestimator_distribution}.

 \subsection{Two types of solutions along the clipping trade-off line}
\label{sec:basin}

Previous experiments suggest that hyperparameter choices along the \textit{clipping trade-off line} (i.e., $\eta c = 5\times10^{-4}$) converge to two different types of solutions depending on whether clipping is significant enough ($c\leq100$). To visualize the geometry of these solutions, similar to \citep{li2018visualizing,garipov2018loss}, we project the high-dimensional train loss and train accuracy landscapes onto the 2D plane spanned by three model checkpoints along this clipping trade-off line: $w_1$ ($\eta = 5e{-5}$, $c=10$), $w_2$ ($\eta = 1e{-5}$, $c=50$), and $w_3$ ($\eta = 1e{-6}$, $c=\infty$).

\begin{figure}[h]
\begin{center}
\includegraphics[width=0.4\columnwidth]{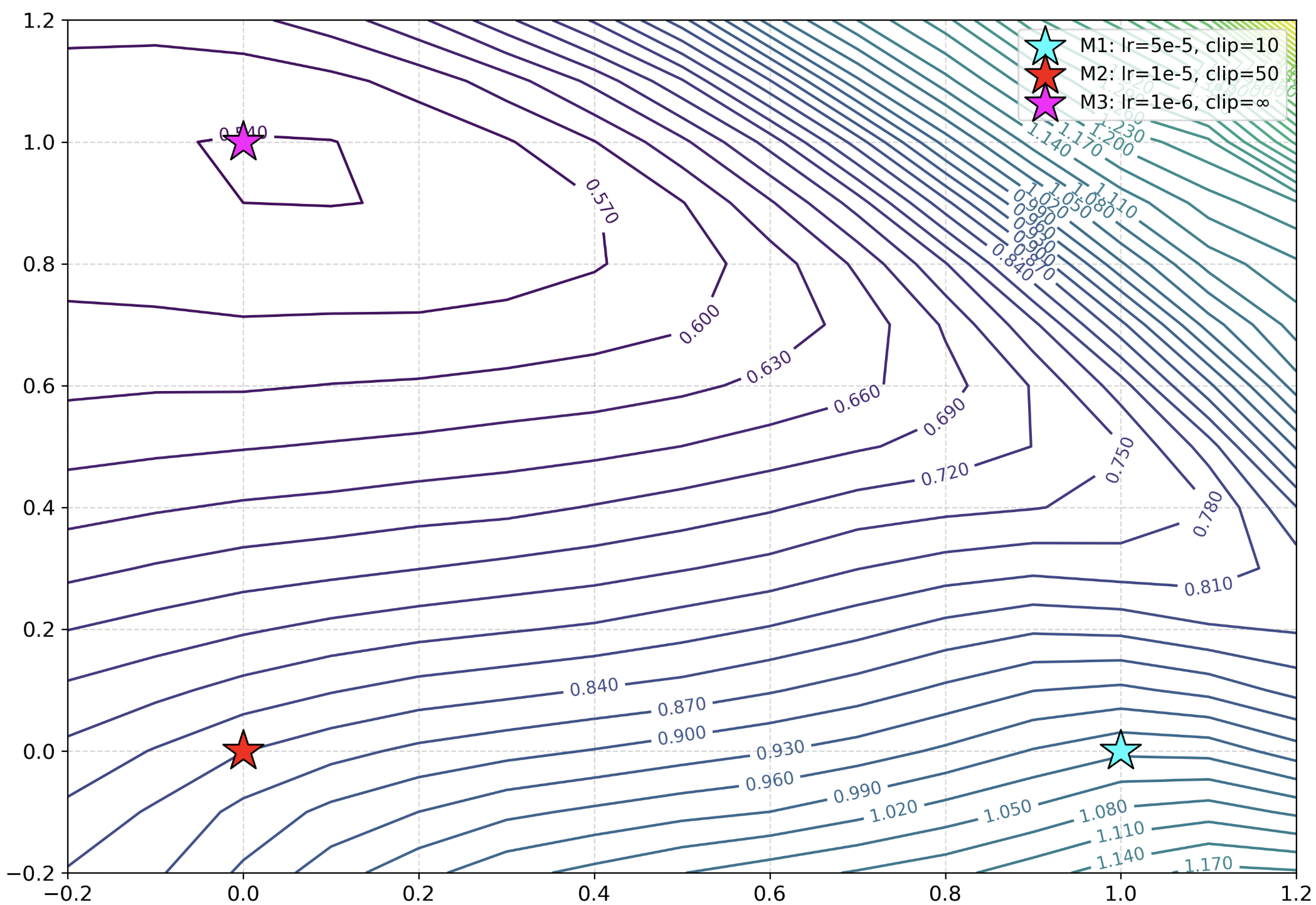}
    \put(-79,-5){$x$}
    \put(-170,60){$y$}
    \;\;\;\;\;\;\;
\includegraphics[width=0.4\columnwidth]{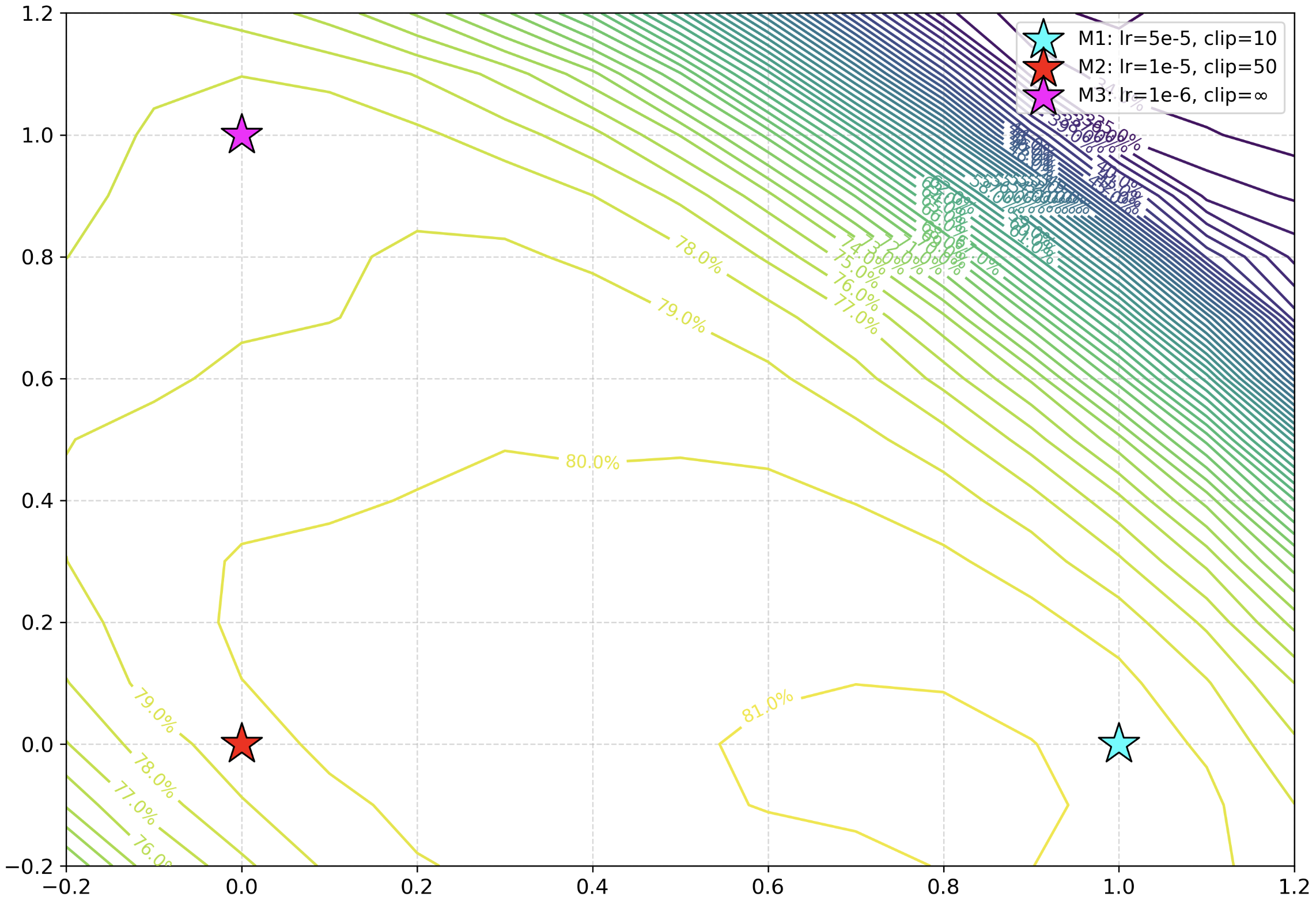}
    \put(-79,-5){$x$}
    \put(-170,60){$y$}    
\end{center}
\caption{Pretrained RoBERTa fine-tuned on SNLI. (left) Train loss landscape projected down to $(x,y)$-plane evaluated at $w_2+(w_1-w_2)x+(w_3-w_2)y$ for three  model checkpoints $w_1,w_2,w_3$. (right) Train Accuracy landscape projected down to $(x,y)$-plane evaluated at $w_2+(w_1-w_2)x+(w_3-w_2)y$ for three  model checkpoints $w_1,w_2,w_3$. This suggests that per-sample clipping introduces an implicit bias towards higher accuracy solutions.
}
\label{fig:mismatches}
\end{figure}

\sg{We plot the training surface in Figure~\ref{fig:mismatches} to directly inspect the objective landscape accessed by the optimizer.} The train loss landscape (Figure~\ref{fig:mismatches}, left) clearly shows that unclipped ZO-SGD, $w_3$, converged close to a local minima of the train loss, whereas  clipped ZO-SGD, $w_1$ and $w_2$, converged to solutions with higher train loss. On the other hand, the train accuracy landscape (Figure~\ref{fig:mismatches}, right) illustrates that $w_1$ and $w_2$ converged to a solution with higher train accuracy compared to $w_3$. This suggests that per-sample clipping introduces an implicit bias in the training that favors higher train accuracy at the cost of higher train loss. These checkpoints generalize well and the same trend holds for test loss and test accuracy, illustrated in Figure~\ref{fig:mismatches_test} in the appendix.

\begin{figure}[h]
\begin{center}
    \includegraphics[width=0.31\columnwidth]{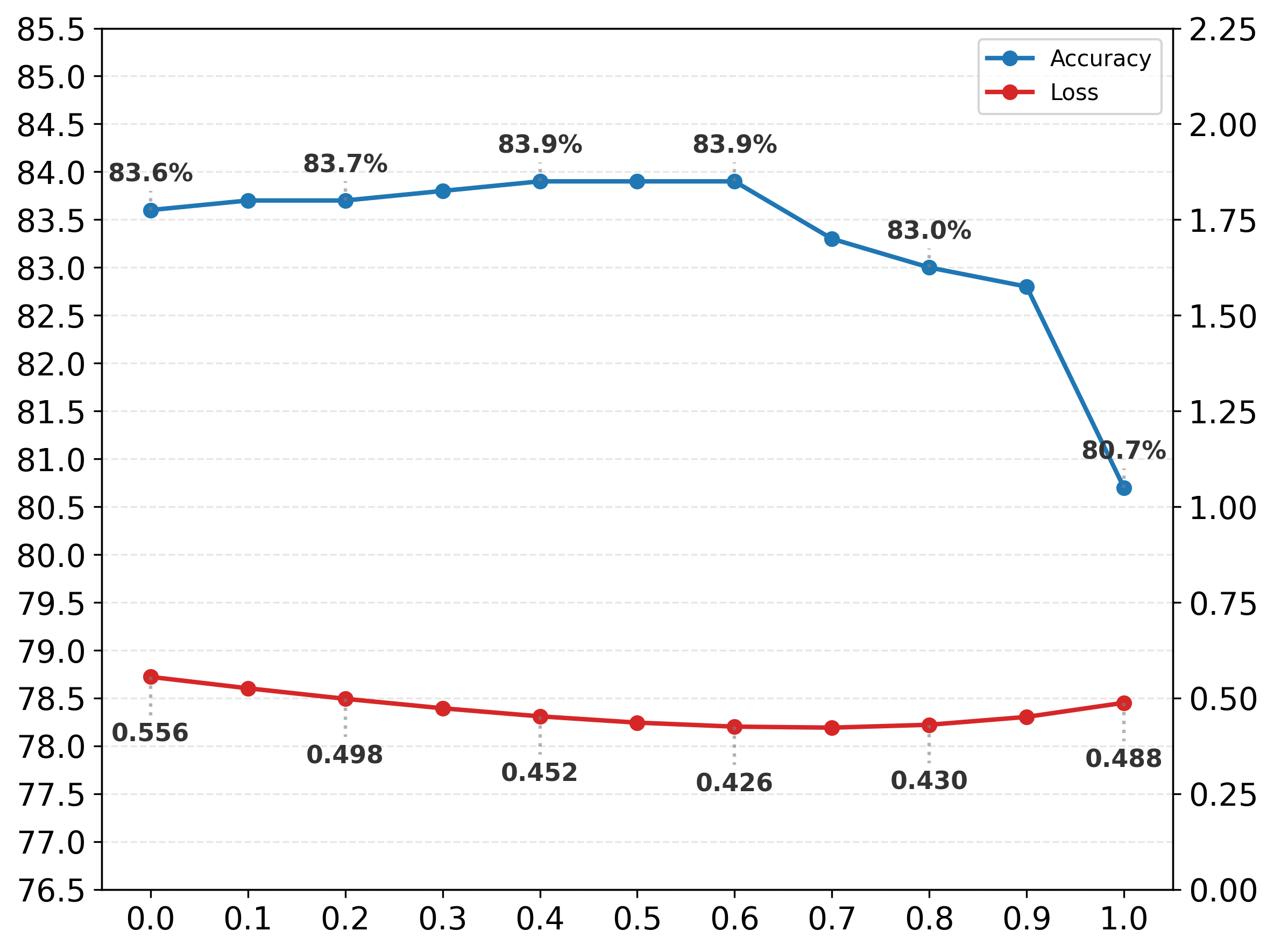}
    \put(-65,-7){$\alpha$}
    \put(-135,43){\rotatebox[origin=c]{90}{test accuracy}}
    \includegraphics[width=0.31\columnwidth]{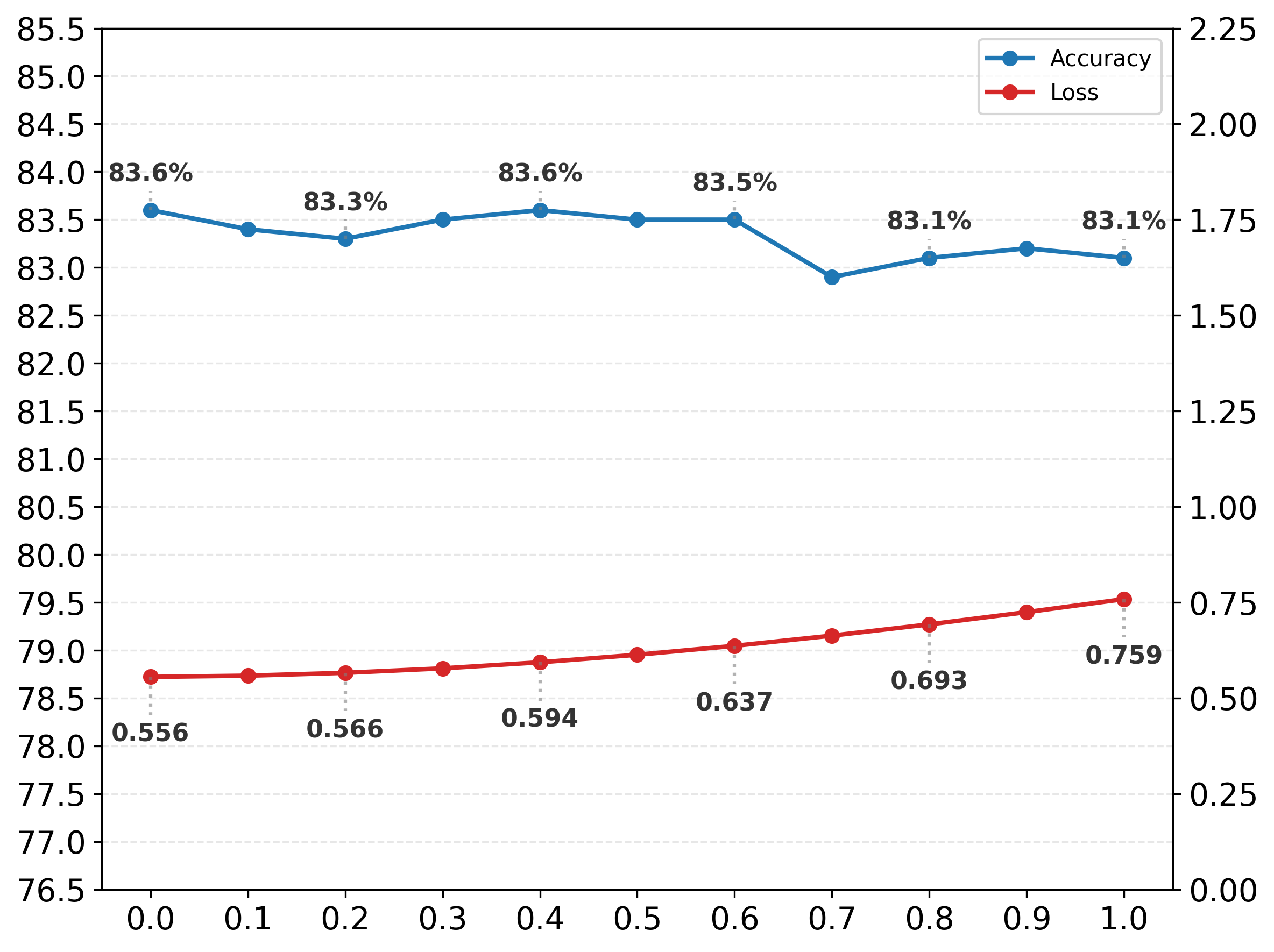}
    \put(-65,-7){$\alpha$}
    \includegraphics[width=0.31\columnwidth]{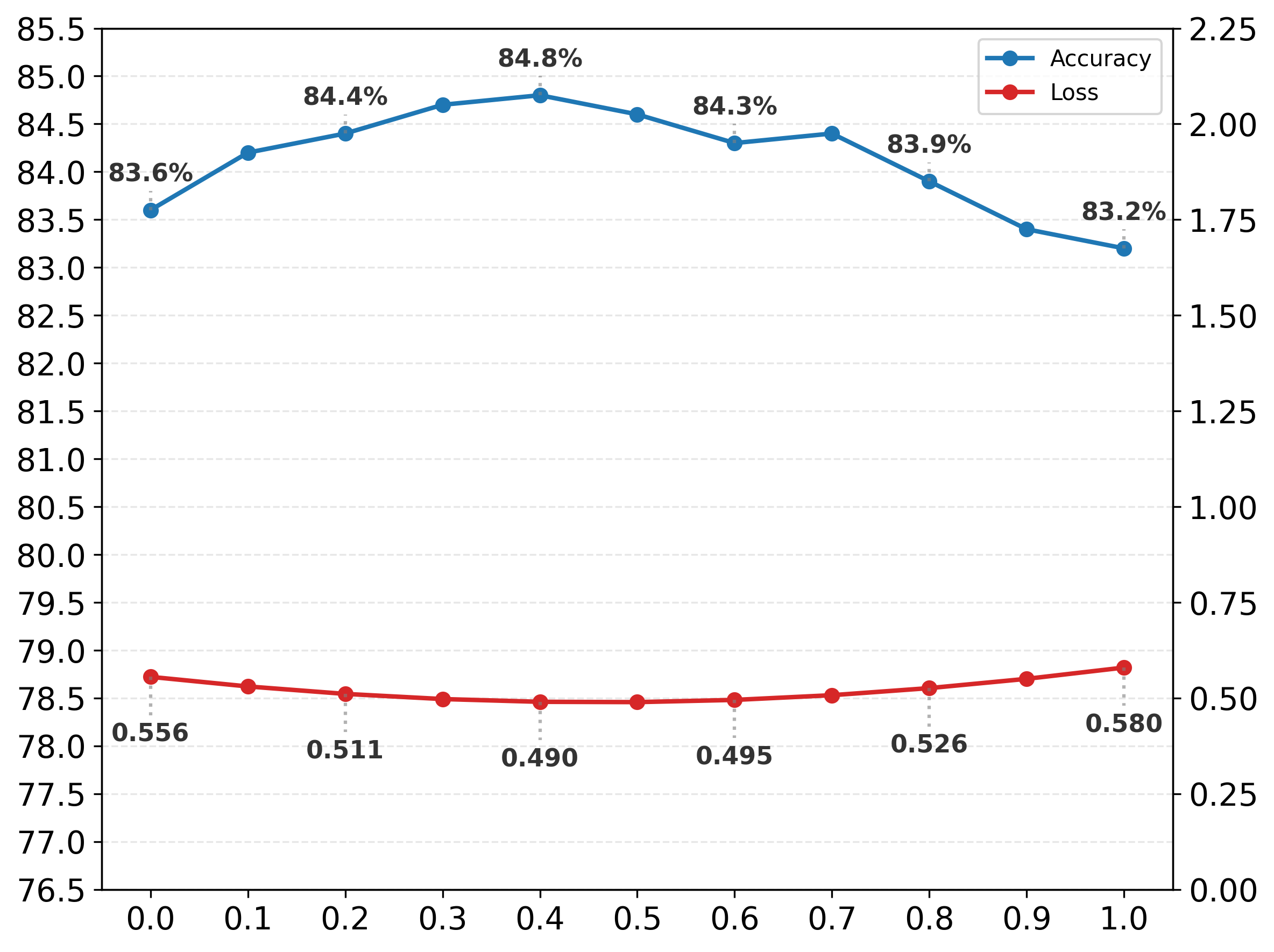}
    \put(-65,-7){$\alpha$}
    \put(2,45){\rotatebox[origin=c]{270}{test loss}}
\end{center}
\caption{For models $w_1$, $w_2$, and $w_3$ from Figure~\ref{fig:mismatches}, ensemble model of ${\rm SoftMax}((1-\alpha)\,h(x;w_2)+\alpha\,h(x;w_3))$, where $h$ is the logit output, achieves better SNLI \sg{test} accuracy than the two original models (left) but 
two clipped models, $w_2$ and $w_1$, lack diversity and do not show ensemble gain (middle). Two clipped models can still get ensemble gain, if they use two different random seeds (right). }
\label{fig:logit_ensembles}
\end{figure}

Another way to measure how two model weights are related is to visualize the ensemble model performances \cite{ainsworthgit,nasery2025pleas,wortsman2022model}. Averaging the logit output of two models and using the softmax to predict, known as an ensemble model, has been observed to give better performance than the two original models when the two model weights are distinct, typically due to different random seeds or different pre-trained base models. Figure~\ref{fig:logit_ensembles} (middle) demonstrates how the ensemble of two clipped models $w_1$ and $w_2$ provides no gain as expected, since we use the same random seeds and initialization of RoBERTa; they only differ in the hyper parameter $(c,\eta)$. On the other hand,  the ensemble of a clipped model $w_1$ and an unclipped model $w_3$ enjoys the ensemble performance gain in both test loss and accuracy (Figure~\ref{fig:logit_ensembles}, left). This suggests that ensemble model is exploiting the diversity between the two models found in the \sg{clipping-dominated} and unclipped regimes. Figure~\ref{fig:logit_ensembles} (right) shows as a sanity check that  such a diversity can also be exploited by training two clipped models with different random seeds. 

\sg{Together, these observations rule out the alternative explanation that $c \le 100$ and $c \ge 500$ merely reflect different optimization speeds along the same trajectory. First, if unclipped ZO-SGD ($w_3$) were simply progressing more slowly toward the same minimum as clipped ZO-SGD ($w_1, w_2$), $w_3$ would exhibit a \emph{higher} training loss than $w_1$ and $w_2$; instead, $w_3$ converges to a strictly \emph{lower} training and test loss ($0.45$ vs.\ $0.70$ at $c=10$, Figure~\ref{grid_loss_roberta} and Figure~\ref{fig:roberta_snli_grid} right) while achieving lower test accuracy. Second, varying $(\eta, c)$ from $w_1$ to $w_2$ yields zero ensemble gain (Figure~\ref{fig:logit_ensembles}, middle), whereas ensembling clipped and unclipped checkpoints $w_2$ and $w_3$ improves both test accuracy and test loss (Figure~\ref{fig:logit_ensembles}, left). This empirical separation matches Proposition~\ref{propo:clippedgradient}: at any stationary point $w^*$ of the unclipped objective satisfying $\sum_{i=1}^n \nabla f_i(w^*) = 0$, the expected clipped update $\frac{1}{n}\sum_{i=1}^n h_c(\|\nabla f_i(w^*)\|)\nabla f_i(w^*)$ remains non-zero whenever per-sample gradient norms vary across examples, driving clipped ZO-SGD to a distinct stationary point.}

\subsection{Implicit bias towards higher accuracy} 
\label{sec:3bias} 

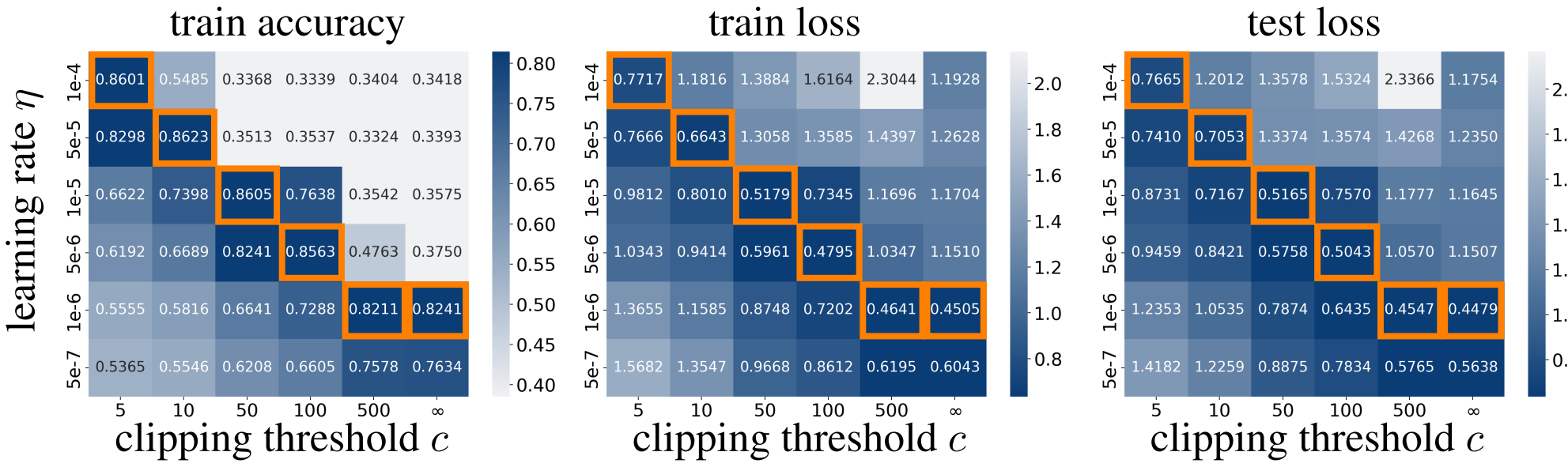
\begin{figure}[ht]
\begin{center}
\begin{tikzpicture}
        \node[anchor=south west, inner sep=0, outer sep=0]
            (image) at (0,0)
    {\includegraphics[width=0.31\columnwidth]{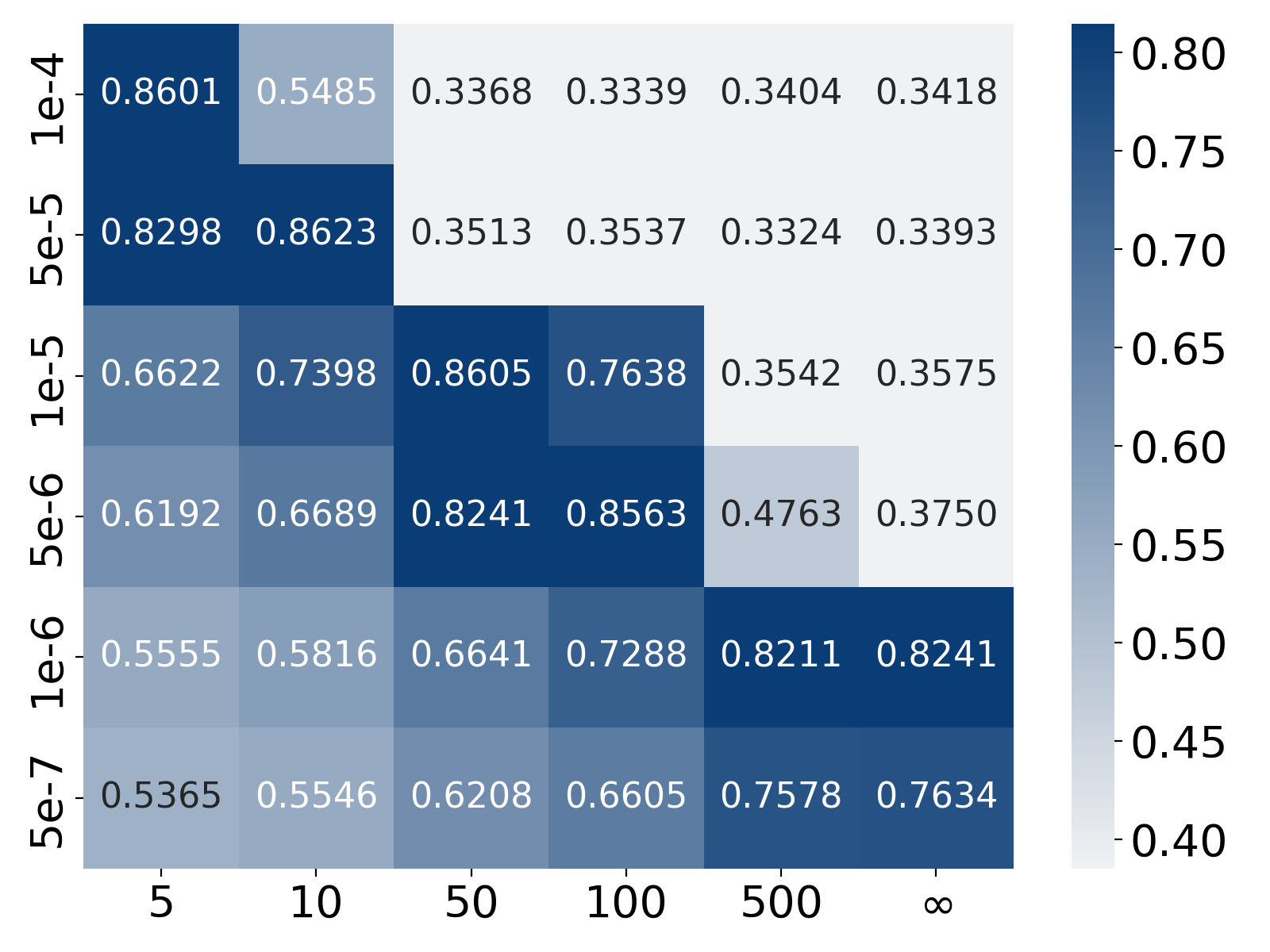}};
        \begin{scope}[x={(image.south east)}, y={(image.north west)}]
            \draw[orange, line width=1.5pt] (0.185,0.968) rectangle (0.071,0.835);
        \end{scope} 
\begin{scope}[x={(image.south east)}, y={(image.north west)}]      \draw[orange, line width=1.5pt] (0.307,0.687) rectangle(0.194,0.820); \end{scope} 
\begin{scope}[x={(image.south east)}, y={(image.north west)}]      \draw[orange, line width=1.5pt] (0.428,0.537) rectangle(0.315,0.674); \end{scope} 
\begin{scope}[x={(image.south east)}, y={(image.north west)}]      \draw[orange, line width=1.5pt] (0.553,0.392) rectangle(0.438,0.525); \end{scope} 
\begin{scope}[x={(image.south east)}, y={(image.north west)}]      \draw[orange, line width=1.5pt] (0.675,0.242) rectangle(0.560,0.375); \end{scope} 
\begin{scope}[x={(image.south east)}, y={(image.north west)}]      \draw[orange, line width=1.5pt] (0.793,0.242) rectangle(0.684,0.375); \end{scope}
    \end{tikzpicture}
    \put(-109,-5){clipping threshold $c$}
    \put(-96,94){train accuracy}
    \put(-135,50){\rotatebox[origin=c]{90}{learning rate $\eta$}}
\begin{tikzpicture}
        \node[anchor=south west, inner sep=0, outer sep=0]
            (image) at (0,0)
    {\includegraphics[width=0.31\columnwidth]{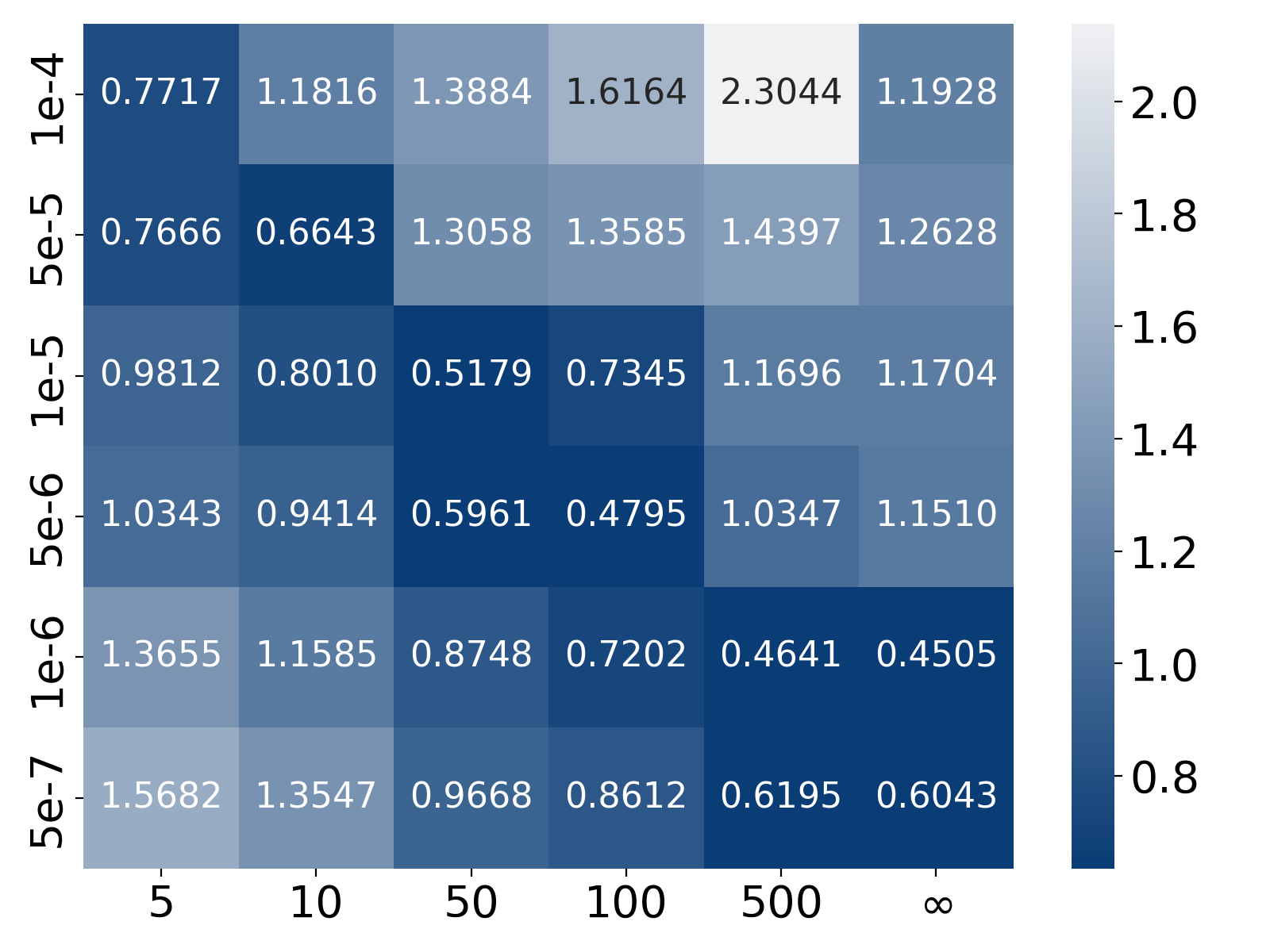}};
        \begin{scope}[x={(image.south east)}, y={(image.north west)}]
            \draw[orange, line width=1.5pt] (0.185,0.968) rectangle (0.071,0.835);
        \end{scope} 
\begin{scope}[x={(image.south east)}, y={(image.north west)}]      \draw[orange, line width=1.5pt] (0.307,0.687) rectangle(0.194,0.820); \end{scope} 
\begin{scope}[x={(image.south east)}, y={(image.north west)}]      \draw[orange, line width=1.5pt] (0.428,0.537) rectangle(0.315,0.674); \end{scope} 
\begin{scope}[x={(image.south east)}, y={(image.north west)}]      \draw[orange, line width=1.5pt] (0.553,0.392) rectangle(0.438,0.525); \end{scope} 
\begin{scope}[x={(image.south east)}, y={(image.north west)}]      \draw[orange, line width=1.5pt] (0.675,0.242) rectangle(0.560,0.375); \end{scope} 
\begin{scope}[x={(image.south east)}, y={(image.north west)}]      \draw[orange, line width=1.5pt] (0.793,0.242) rectangle(0.684,0.375); \end{scope}
    \end{tikzpicture} 
    \put(-109,-5){clipping threshold $c$}
    \put(-91,94){train loss}
\begin{tikzpicture}
        \node[anchor=south west, inner sep=0, outer sep=0]
            (image) at (0,0)
    {\includegraphics[width=0.31\columnwidth]{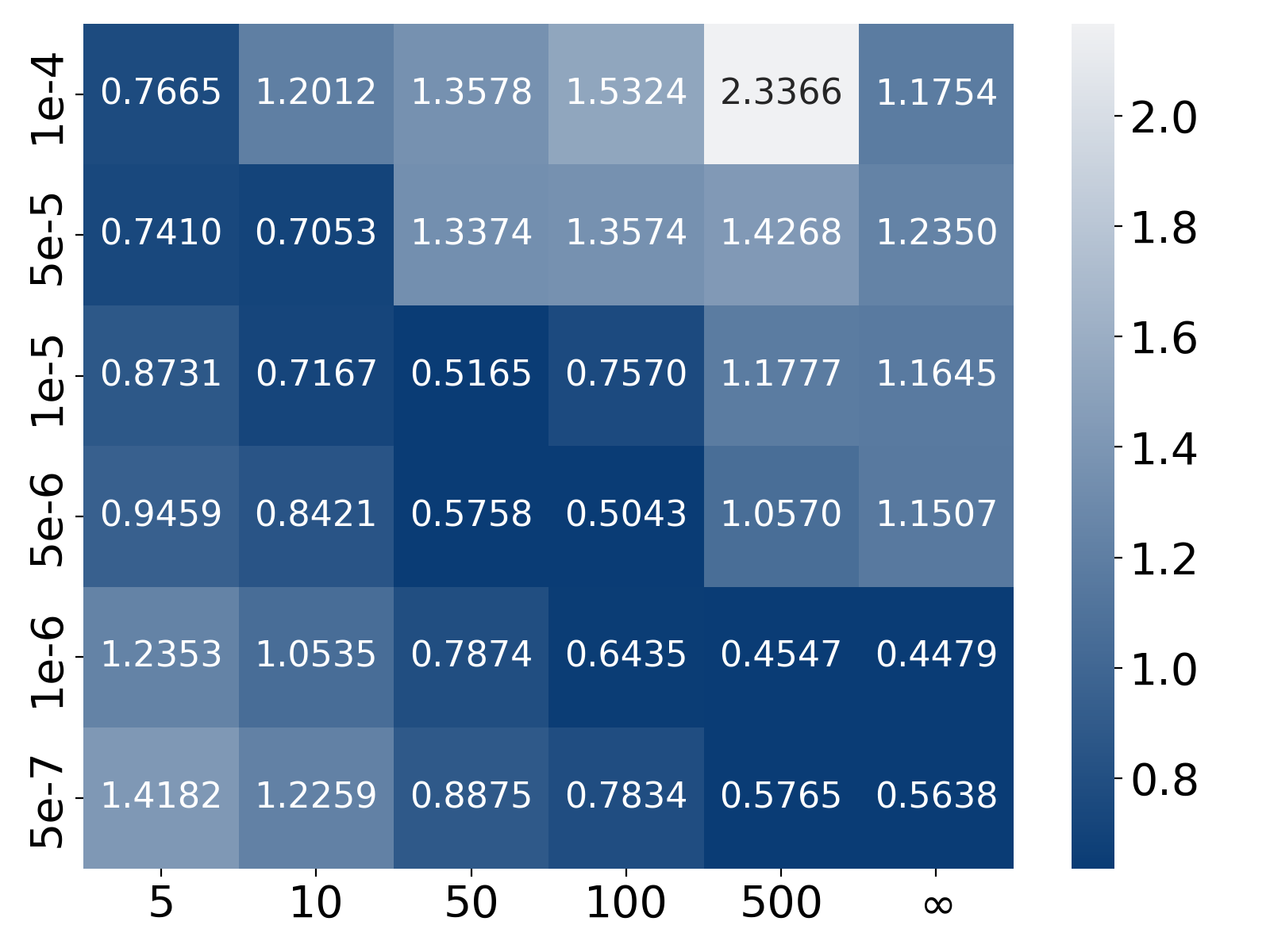}};
        \begin{scope}[x={(image.south east)}, y={(image.north west)}]
            \draw[orange, line width=1.5pt] (0.185,0.968) rectangle (0.071,0.835);
        \end{scope} 
\begin{scope}[x={(image.south east)}, y={(image.north west)}]      \draw[orange, line width=1.5pt] (0.307,0.687) rectangle(0.194,0.820); \end{scope} 
\begin{scope}[x={(image.south east)}, y={(image.north west)}]      \draw[orange, line width=1.5pt] (0.428,0.537) rectangle(0.315,0.674); \end{scope} 
\begin{scope}[x={(image.south east)}, y={(image.north west)}]      \draw[orange, line width=1.5pt] (0.553,0.392) rectangle(0.438,0.525); \end{scope} 
\begin{scope}[x={(image.south east)}, y={(image.north west)}]      \draw[orange, line width=1.5pt] (0.675,0.242) rectangle(0.560,0.375); \end{scope} 
\begin{scope}[x={(image.south east)}, y={(image.north west)}]      \draw[orange, line width=1.5pt] (0.793,0.242) rectangle(0.684,0.375); \end{scope}
    \end{tikzpicture}
    \put(-109,-5){clipping threshold $c$}
    \put(-86,94){test loss}
\caption{The same setting as Figure~\ref{fig:grid_avg_roberta} but showing the average train accuracy, train loss, and test loss. Both clipped and unclipped ZO-SGD generalizes well, but the clipped ZO-SGD has an implicit bias towards higher accuracy (and higher loss) models. 
}
\label{grid_loss_roberta}
\end{center}
\end{figure}

We first ask if clipped ZO-SGD is implicitly regularized and the resulting prediction generalizes better than unclipped ZO-SGD. Comparing the accuracy and loss between train data and test data (Figure~\ref{fig:grid_avg_roberta} left and Figure~\ref{grid_loss_roberta}), all the models in the orange trade-off line generalize well; there is no difference in generalization between the two regimes. Next, we ask if clipped ZO-SGD has implicit bias towards certain types of solutions. The prominent distinction for the models trained in the \sg{clipping-dominated} regime ($c\leq100$) is that their \sg{train and test} accuracy and loss are both higher than the unclipped counterparts ($c\geq 500$). Comparing the train loss on the orange cells in the $c\leq 100$ regime, more clipping leads to more bias, resulting in significantly larger train loss for $c=5$ compared to $c=100$. However, despite achieving significantly different train losses, all those  models trained with clipping achieve similar \sg{train and test} accuracy that is significantly better than the models trained without clipping. The gain of clipping is quite robust to the choice of hyperparameters, as long as we are on the clipping trade-off line, i.e. $\eta c = $ constant. We provide the training loss and accuracy over the training iterations in Figures~\ref{fig:roberta_snli_grid}-\ref{fig:roberta_trec_grid_off} in the appendix for each task: SNLI, MNLI, SST-2, and Trec.

\subsection{Connections to robust gradient estimation} 
\label{sec:robust}

One natural conjecture on why the clipped method converges to a solution with higher accuracy is that it provides a {\em robust estimate} of the gradient, in which case other robust gradient estimators might achieve a similar gain. We test this hypothesis on the following two robust estimators.  

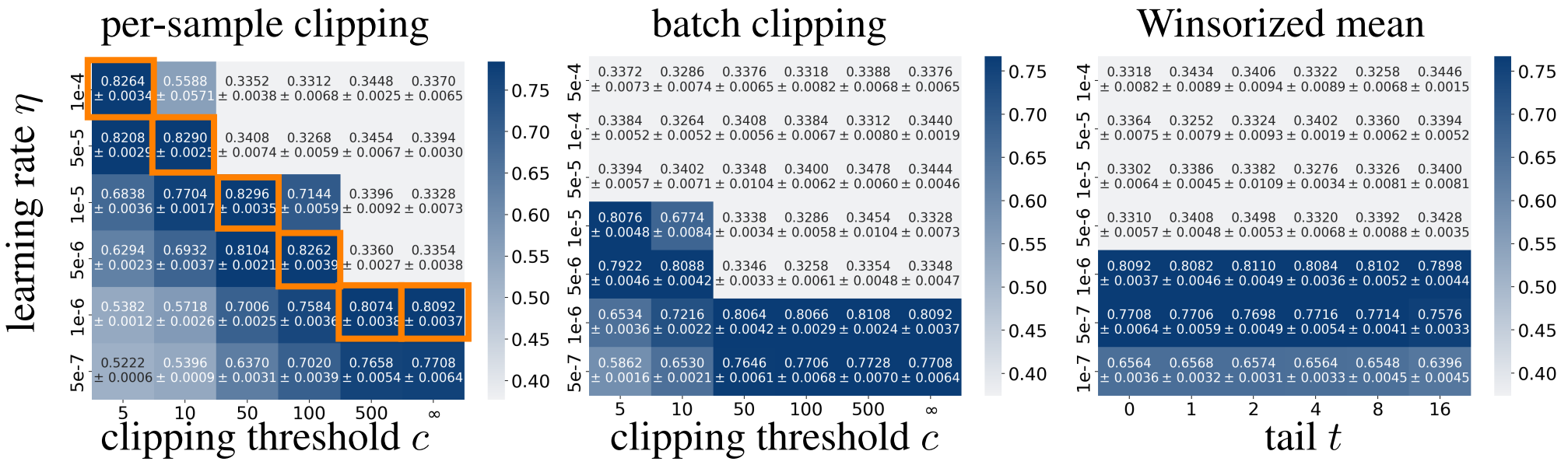
\begin{figure}[h]
\begin{center}
\begin{tikzpicture}
        \node[anchor=south west, inner sep=0, outer sep=0]
            (image) at (0,0)
    {\includegraphics[width=0.3\columnwidth]{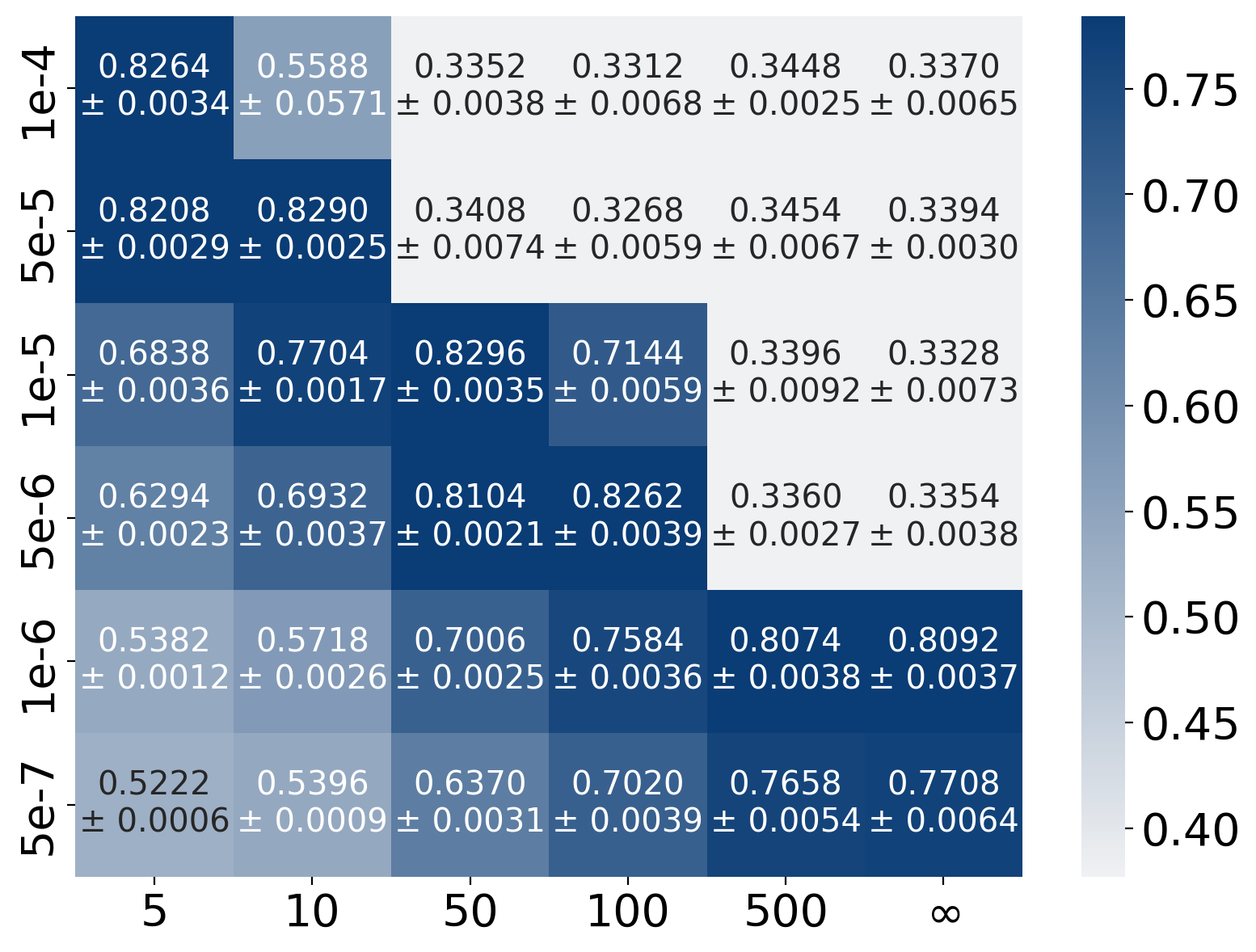}};
        \begin{scope}[x={(image.south east)}, y={(image.north west)}]
            \draw[orange, line width=1.5pt] (0.185,0.988) rectangle (0.051,0.835);
        \end{scope} 
\begin{scope}[x={(image.south east)}, y={(image.north west)}]      \draw[orange, line width=1.5pt] (0.307,0.687) rectangle(0.182,0.835); \end{scope} 
\begin{scope}[x={(image.south east)}, y={(image.north west)}]      \draw[orange, line width=1.5pt] (0.438,0.537) rectangle(0.315,0.674); \end{scope} 
\begin{scope}[x={(image.south east)}, y={(image.north west)}]      \draw[orange, line width=1.5pt] (0.563,0.382) rectangle(0.438,0.525); \end{scope} 
\begin{scope}[x={(image.south east)}, y={(image.north west)}]      \draw[orange, line width=1.5pt] (0.687,0.242) rectangle(0.560,0.385); \end{scope} 
\begin{scope}[x={(image.south east)}, y={(image.north west)}]      \draw[orange, line width=1.5pt] (0.823,0.242) rectangle(0.687,0.385); \end{scope}
    \end{tikzpicture}
\put(-110,95){per-sample clipping}
    \put(-110,-5){clipping threshold $c$}
    \put(-133,50){\rotatebox[origin=c]{90}{learning rate $\eta$}}
\includegraphics[width=0.31\columnwidth]{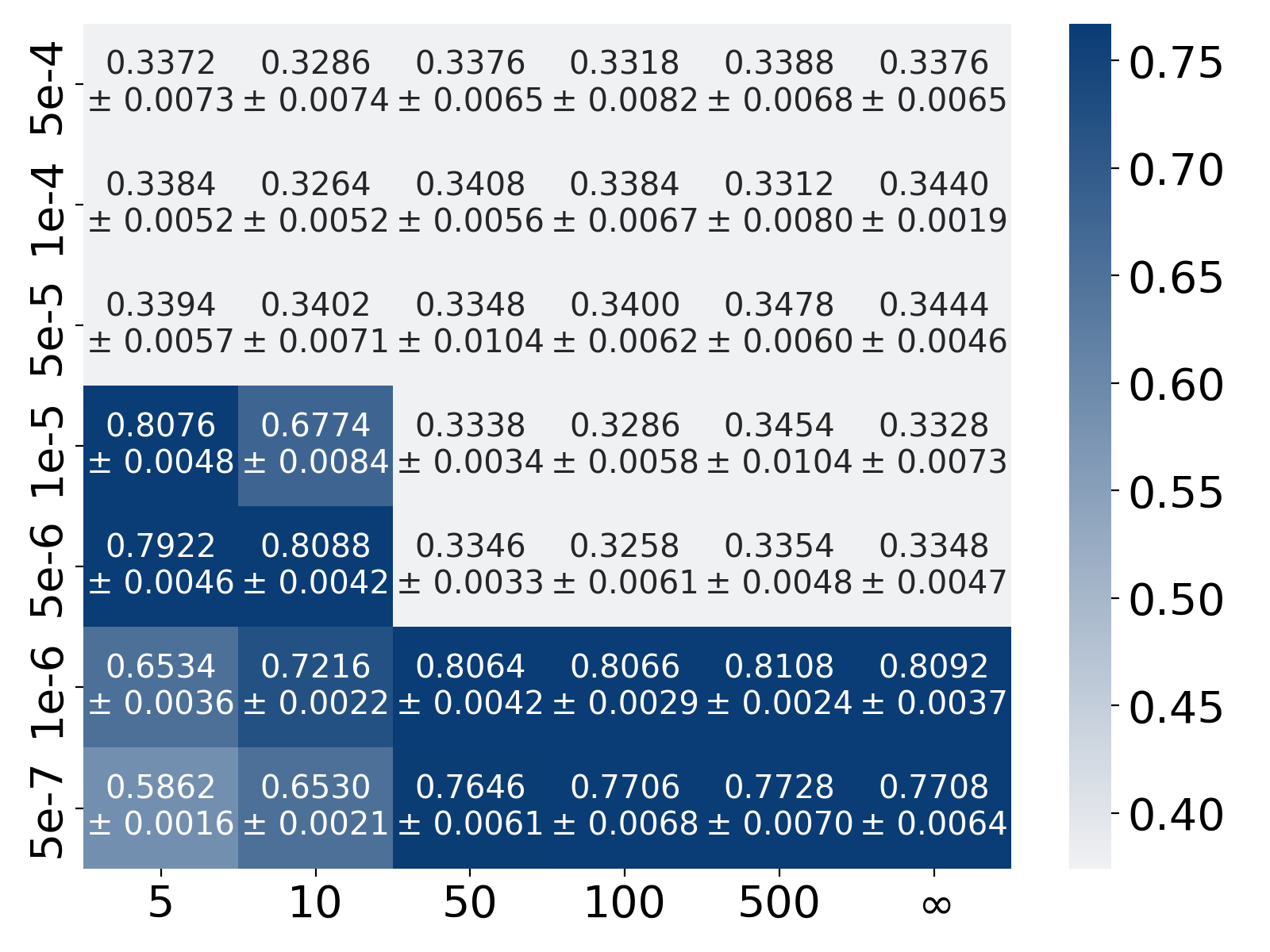}
    \put(-100,95){batch clipping}
    \put(-110,-5){clipping threshold $c$}
\includegraphics[width=0.31\columnwidth]{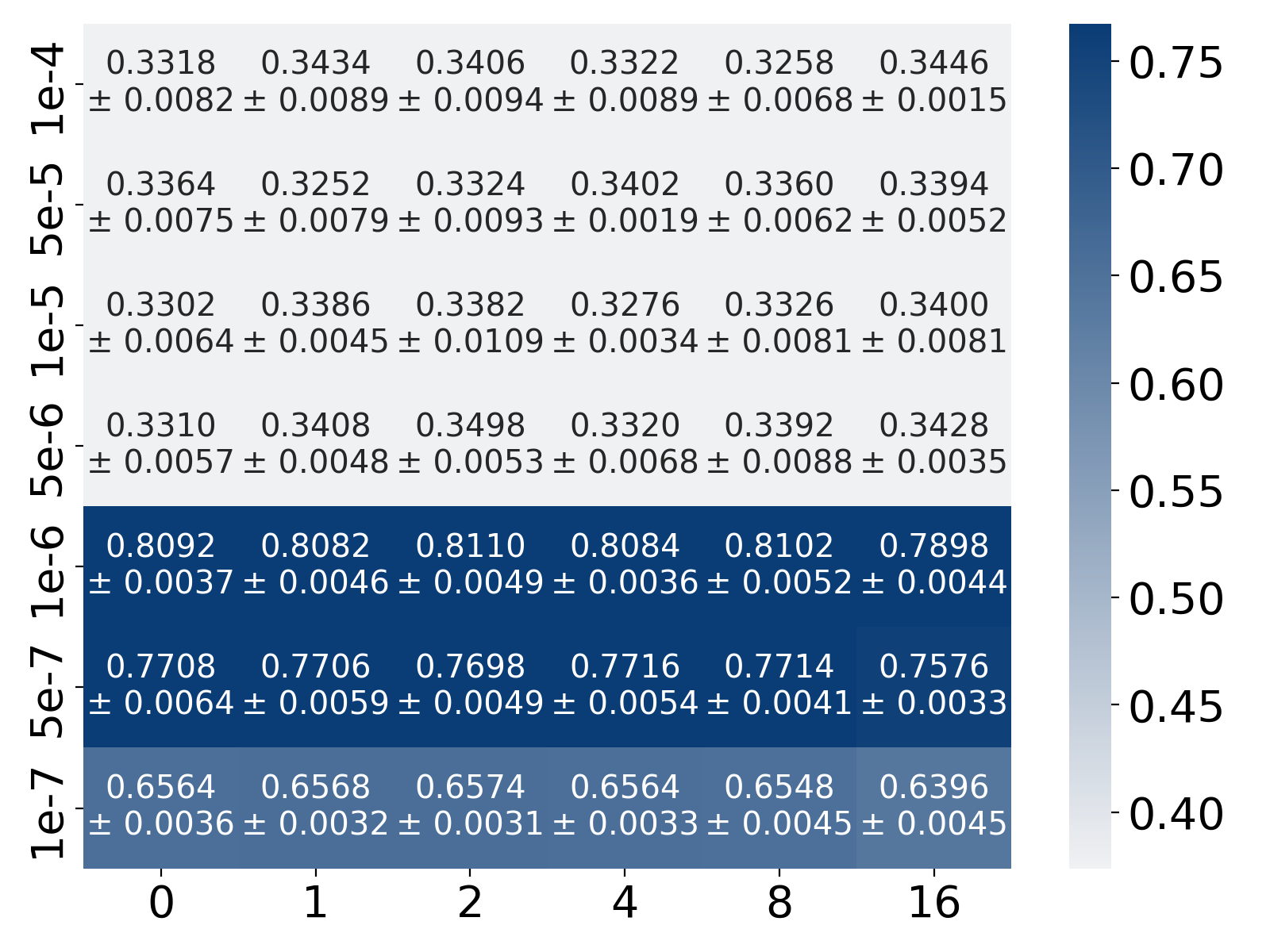} 
    \put(-106,95){Winsorized mean}
    \put(-75,-5){tail $t$}
\end{center}
\vspace{-3mm}
\caption{\sg{SNLI test accuracy (mean $\pm$ standard error across $5$ random seeds) for per-sample clipped ZO-SGD (left), batch clipping $g_{\lambda,\bar c}$ in \eqref{eq:clip_out} (middle), and Winsorized mean $g_{\lambda,k}$ in \eqref{eq:winsor} (right). The $\eta c = 5\times 10^{-4}$ clipping trade-off and test accuracy gains ($82.62\%$--$82.96\%$ along $c \in [5, 100]$ vs.\ $80.92\% \pm 0.37\%$ unclipped) occur exclusively under per-sample clipping; neither batch clipping (peaking at $81.08\% \pm 0.24\%$) nor Winsorized mean (peaking at $81.10\% \pm 0.49\%$) exhibits an inverse scaling relationship with $\eta$ or comparable test accuracy gains.}}
\vspace{8pt}
\label{fig:other_methods}
\vskip -0.1in
\end{figure}

ZO optimization with {\em batch-clipping} applies clipping to the average gradient: 
\begin{eqnarray}
   g_{\lambda,\bar{c}}(w_t) \;=\; {\rm Clip}_{\bar c}\Big( \frac{1}{|B_t|}\sum_{i\in B_t} \frac{f_i(w_t+\lambda u_t)-f_i(w_t - \lambda u_t)}{2\lambda}    \Big) u_t\;.
   \label{eq:clip_out}    
\end{eqnarray} 
When applied to the standard first-order SGD, batch-clipping has been empirically shown to achieve better generalization by converging to a wider and flatter basin for FashionMNIST, SVHN, and CIFAR10, when the noise in the gradient is heavy-tailed \cite{wangeliminating,wang2025global}. Their theoretical analysis shows that batch-clipping makes SGD with heavy-tailed noise exponentially slower to escape wider basins. Further, \cite{zhanggradient} showed that this batch-clipped SGD can achieve faster convergence under $(L_0,L_1)$- smoothness, where smoothness is proportional to the gradient norm: $\|\nabla^2{\cal L}(w)\|\leq L_0 + L_1 \|\nabla{\cal L}(w)\|$. It was further empirically demonstrated that the training loss of LSTM on  Penn Treebank and ResNet20 on CIFAR 10  exhibit $(L_0,L_1)$-smoothness. In our case of ZO optimization, however, Figure~\ref{fig:other_methods} (middle) shows that batch-clipping cannot drive the optimization to the better solution, and hence cannot achieve SNLI test accuracy above 82\%. 

{\em Winsorized mean} is a 1D robust mean estimator which adaptively clips the data according to $k$-th order statistics (see Appendix~\ref{app:winsorized_estimator} for details). We apply this to zeroth-order optimization: 
\begin{eqnarray}
   g_{\lambda,k}(w_t) = \frac{1}{|B_t|} \sum_{i\in B_t} {\rm Clip}_{[L_t, U_t]}\Big( \frac{f_i(w_t+\lambda u_t)-f_i(w_t - \lambda u_t)}{2\lambda} \Big) u_t \;,
   \label{eq:winsor}  
\end{eqnarray} 
where $L_t$ and $U_t$ are the $(k+1)$-th and $(|B_t|-k)$-th order statistics of the mini-batch scalar finite differences, and ${\rm Clip}_{[L_t, U_t]}(a) = \max(L_t, \min(a, U_t))$. Similar to clipping outside the mean, Figure~\ref{fig:other_methods} (right) shows that Winsorized mean cannot achieve the test accuracy of clipped ZO-SGD. 


\subsection{Connections to First-Order Clipping Methods}
\label{sec:first_order}

\sg{To isolate what is specific to zeroth-order (ZO) per-sample clipping versus standard first-order (FO) gradient clipping, we evaluate first-order SGD on SNLI under two distinct clipping regimes: \emph{FO per-sample clipping} (clipping each individual gradient norm $\|\nabla f_i(w_t)\|$ to threshold $c$ prior to aggregation) and \emph{FO batch clipping} (clipping the averaged mini-batch gradient norm $\|(1/|B_t|)\sum_{i\in B_t} \nabla f_i(w_t)\|$ to $c$).}

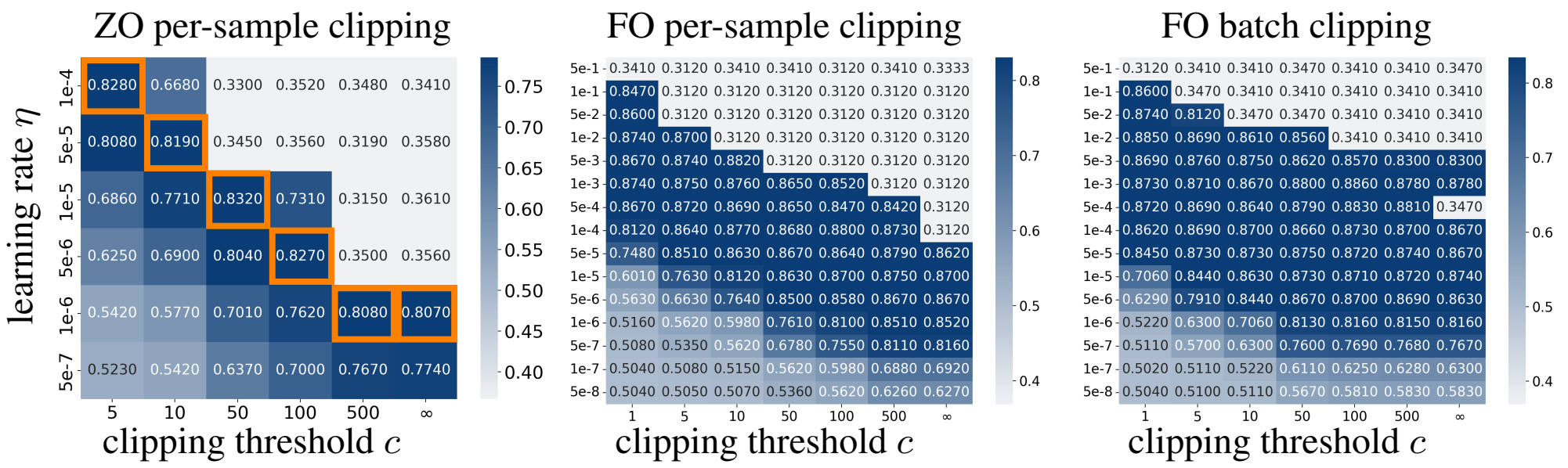
\begin{figure}[h]
\begin{center}
\begin{tikzpicture}
        \node[anchor=south west, inner sep=0, outer sep=0]
            (image) at (0,0)
    {\includegraphics[width=0.31\columnwidth]{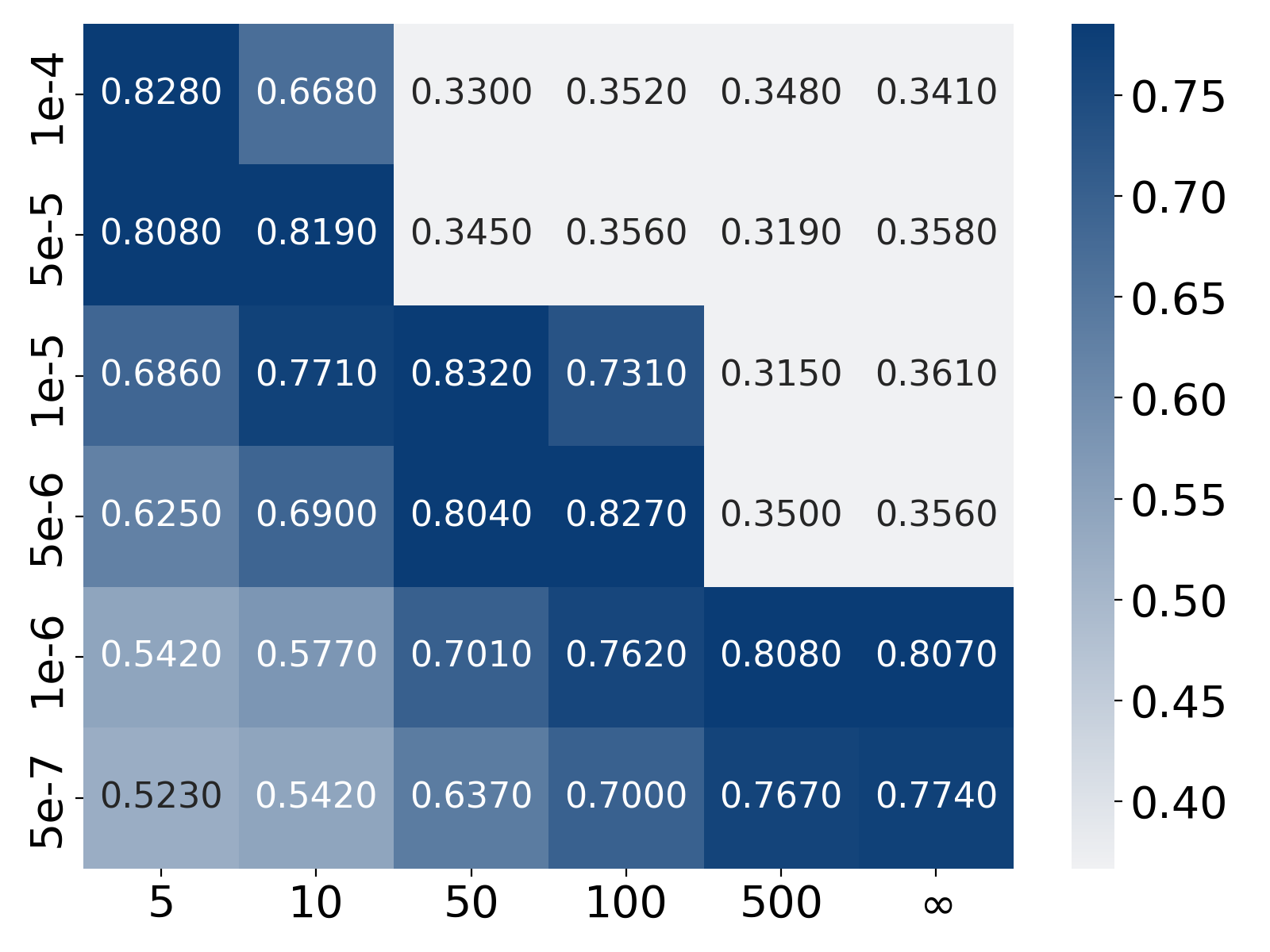}};
        \begin{scope}[x={(image.south east)}, y={(image.north west)}]
            \draw[orange, line width=1.5pt] (0.185,0.968) rectangle (0.071,0.835);
        \end{scope} 
\begin{scope}[x={(image.south east)}, y={(image.north west)}]      \draw[orange, line width=1.5pt] (0.307,0.687) rectangle(0.194,0.820); \end{scope} 
\begin{scope}[x={(image.south east)}, y={(image.north west)}]      \draw[orange, line width=1.5pt] (0.428,0.537) rectangle(0.315,0.674); \end{scope} 
\begin{scope}[x={(image.south east)}, y={(image.north west)}]      \draw[orange, line width=1.5pt] (0.553,0.392) rectangle(0.438,0.525); \end{scope} 
\begin{scope}[x={(image.south east)}, y={(image.north west)}]      \draw[orange, line width=1.5pt] (0.675,0.242) rectangle(0.560,0.375); \end{scope} 
\begin{scope}[x={(image.south east)}, y={(image.north west)}]      \draw[orange, line width=1.5pt] (0.793,0.242) rectangle(0.684,0.375); \end{scope}
    \end{tikzpicture}
    \put(-112,95){\small ZO per-sample clipping}
    \put(-110,-5){\small clipping threshold $c$}
    \put(-133,50){\rotatebox[origin=c]{90}{\small learning rate $\eta$}}
\includegraphics[width=0.31\columnwidth]{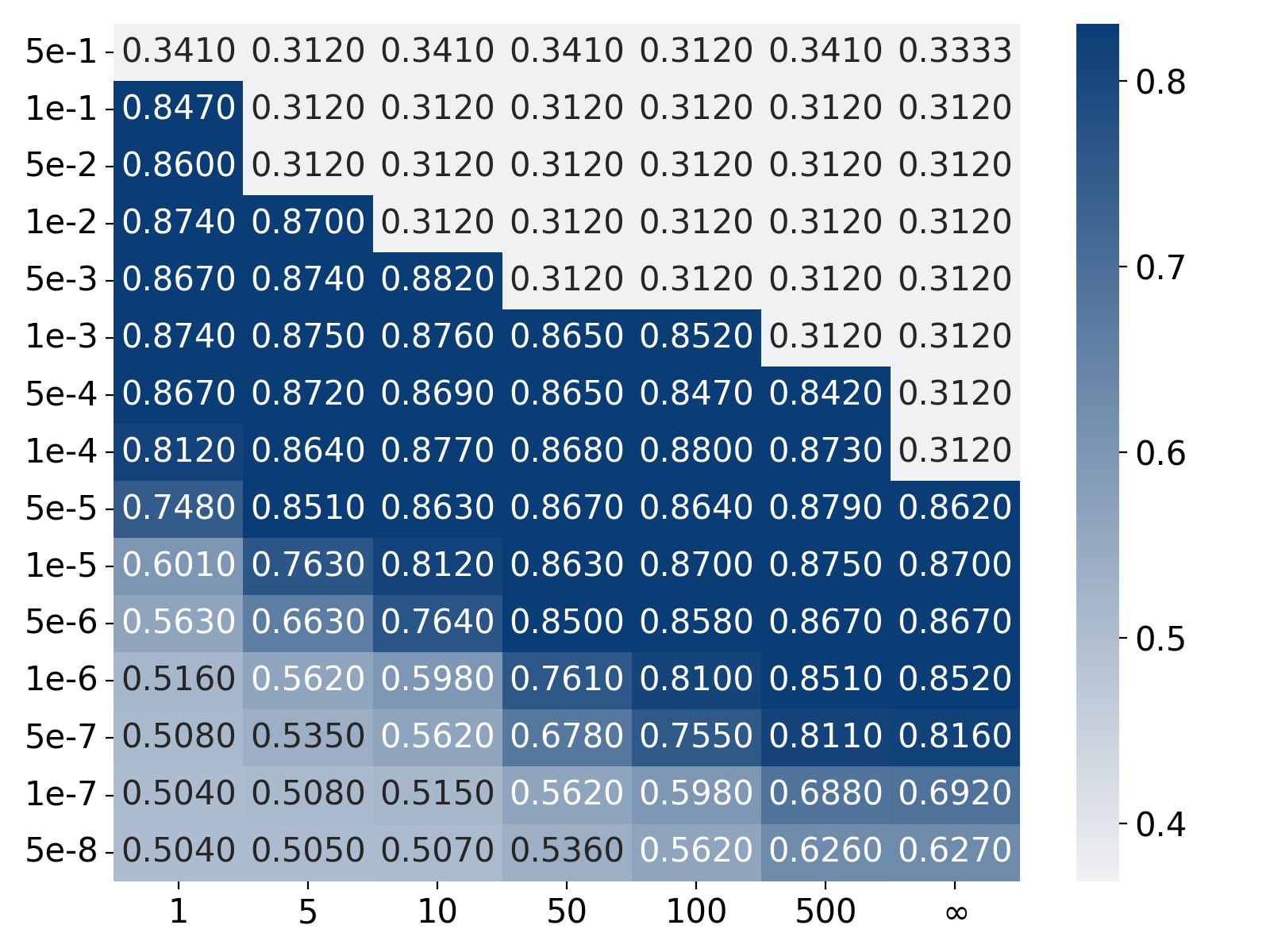}
    \put(-112,95){\small FO per-sample clipping}
    \put(-110,-5){\small clipping threshold $c$}
\includegraphics[width=0.31\columnwidth]{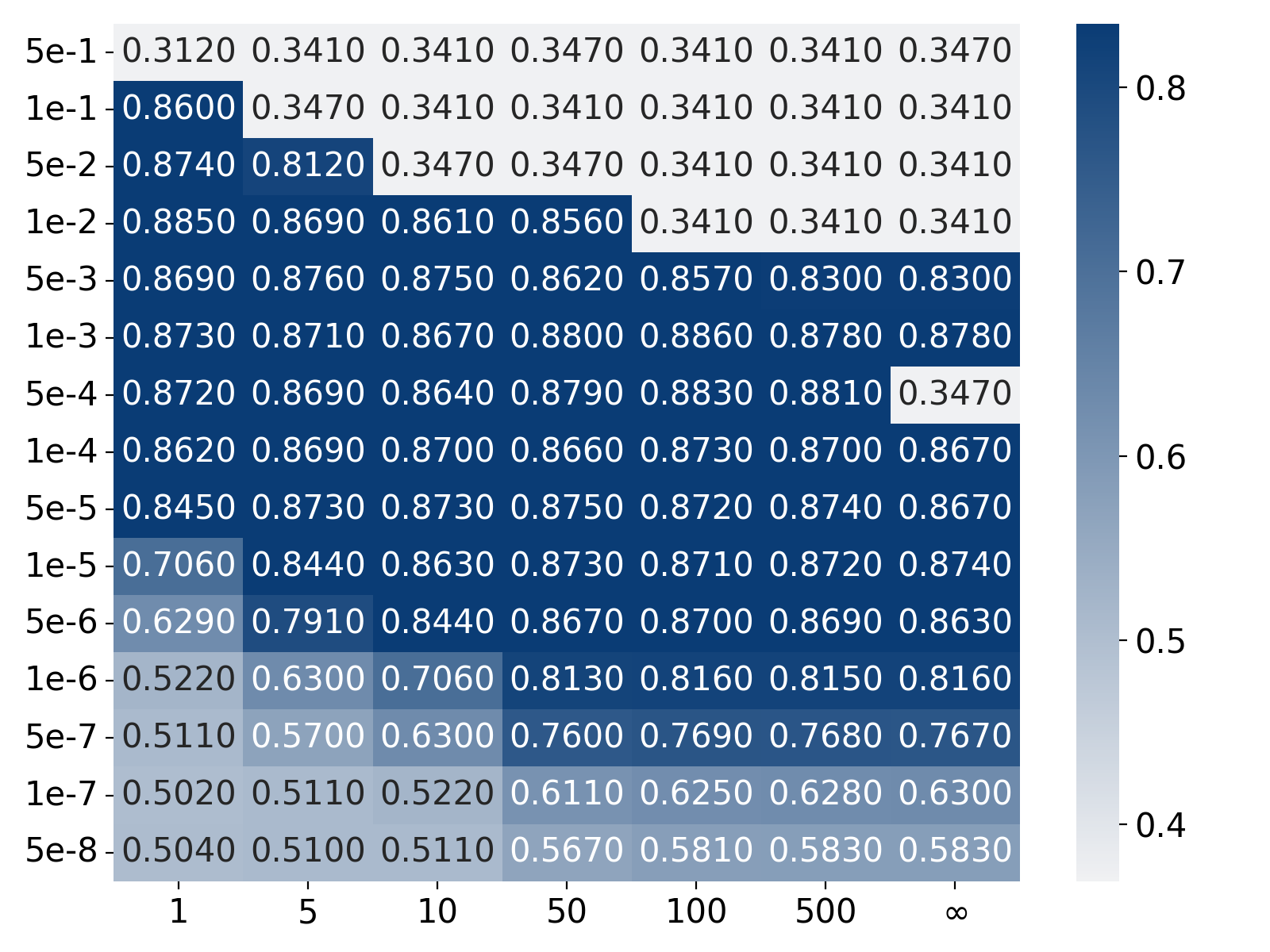} 
    \put(-102,95){\small FO batch clipping}
    \put(-110,-5){\small clipping threshold $c$}
\end{center}
\vspace{-3mm}
\caption{SNLI test accuracy (single seed) across learning rates $\eta$ and clipping thresholds $c$ for (left) Zeroth-Order per-sample clipping, (middle) First-Order per-sample clipping, and (right) First-Order batch clipping. Unlike FO methods, which exhibit a wide band of high test accuracy ($86\%$--$88\%$) across multiple orders of magnitude of $\eta$ and $c$, ZO per-sample clipping concentrates peak performance along a strict diagonal ridge ($\eta c = 5\times 10^{-4}$) and yields a much larger relative gain over its unclipped baseline ($c=\infty$).}
\label{fig:first_order_comp}
\end{figure}

\sg{Comparing the three regimes in Figure~\ref{fig:first_order_comp} shows that the clipping trade-off is specific to per-sample clipped ZO-SGD. In both FO per-sample clipping (middle) and FO batch clipping (right), a wide range of $(\eta, c)$ pairs achieves high test accuracy ($86.0\%$--$88.6\%$) without following a strict $\eta c = \text{constant}$ line; for example, at $c=100$, FO per-sample clipping stays above $85.0\%$ test accuracy across $\eta \in [5\times 10^{-6}, 10^{-3}]$, and at $\eta = 10^{-5}$, varying $c \in [50, \infty]$ changes test accuracy very little ($86.3\%$--$87.5\%$). In contrast, ZO per-sample clipping (left) only achieves its best test accuracy ($82.7\%$--$83.2\%$) along the diagonal $\eta c = 5\times 10^{-4}$, while larger $\eta$ values diverge to random guessing ($\sim 31\%$--$36\%$) and smaller $\eta$ values converge too slowly. Furthermore, FO per-sample clipping ($88.2\%$) and FO batch clipping ($88.6\%$) give almost the same peak test accuracy and only a small gain over unclipped FO-SGD ($87.8\%$), whereas in ZO-SGD, batch clipping fails to improve test accuracy (capped at $81.9\%$, Figure~\ref{fig:other_methods} middle) while per-sample clipping improves SNLI test accuracy from $80.7\%$ to $83.2\%$ ($+2.5\%$). Finally, along the stability boundary where clipping prevents divergence, FO per-sample clipping peaks near $\eta c \approx 5\times 10^{-2}$ (e.g., $\eta=5\times 10^{-3}, c=10$), which is $100\times$ larger than $\eta c = 5\times 10^{-4}$ in ZO-SGD. Because ZO-SGD multiplies the clipped scalar finite difference by a random Gaussian vector $u_t \sim \mathcal{N}(0, I_d)$ of norm $\approx \sqrt{d}$, per-sample clipping in ZO both reweights individual gradients by $2\Phi(c/\|\nabla f_i\|)-1$ (Proposition~\ref{propo:clippedgradient}) and caps the variance from the random direction $u_t$, allowing much larger learning rates than unclipped ZO-SGD. In particular, at $c=1$ where FO per-sample clipping reduces to normalized per-example FO gradients, test accuracy reaches at most $87.4\%$ (compared to $87.0\%$ unclipped at $\eta=10^{-5}$), confirming that per-example normalization alone does not explain the large accuracy gain in ZO-SGD.}

\section{Empirical analysis of clipped ZO-SGD} 
\vskip -1mm
\label{sec:analysis} 

To investigate the cause of accuracy gain and loss degradation of clipping (in both train and test data), we analyze the reliability diagram for calibration (Section~\ref{sec:calibrate}), inject artificial label noise (Section~\ref{sec:noise}), and retrain after pruning the frequently clipped examples (Section~\ref{sec:prune}).

\subsection{Clipping leads to miscalibration} 
\label{sec:calibrate}

\begin{figure}[h]
\begin{center}
\includegraphics[width=0.48\columnwidth]{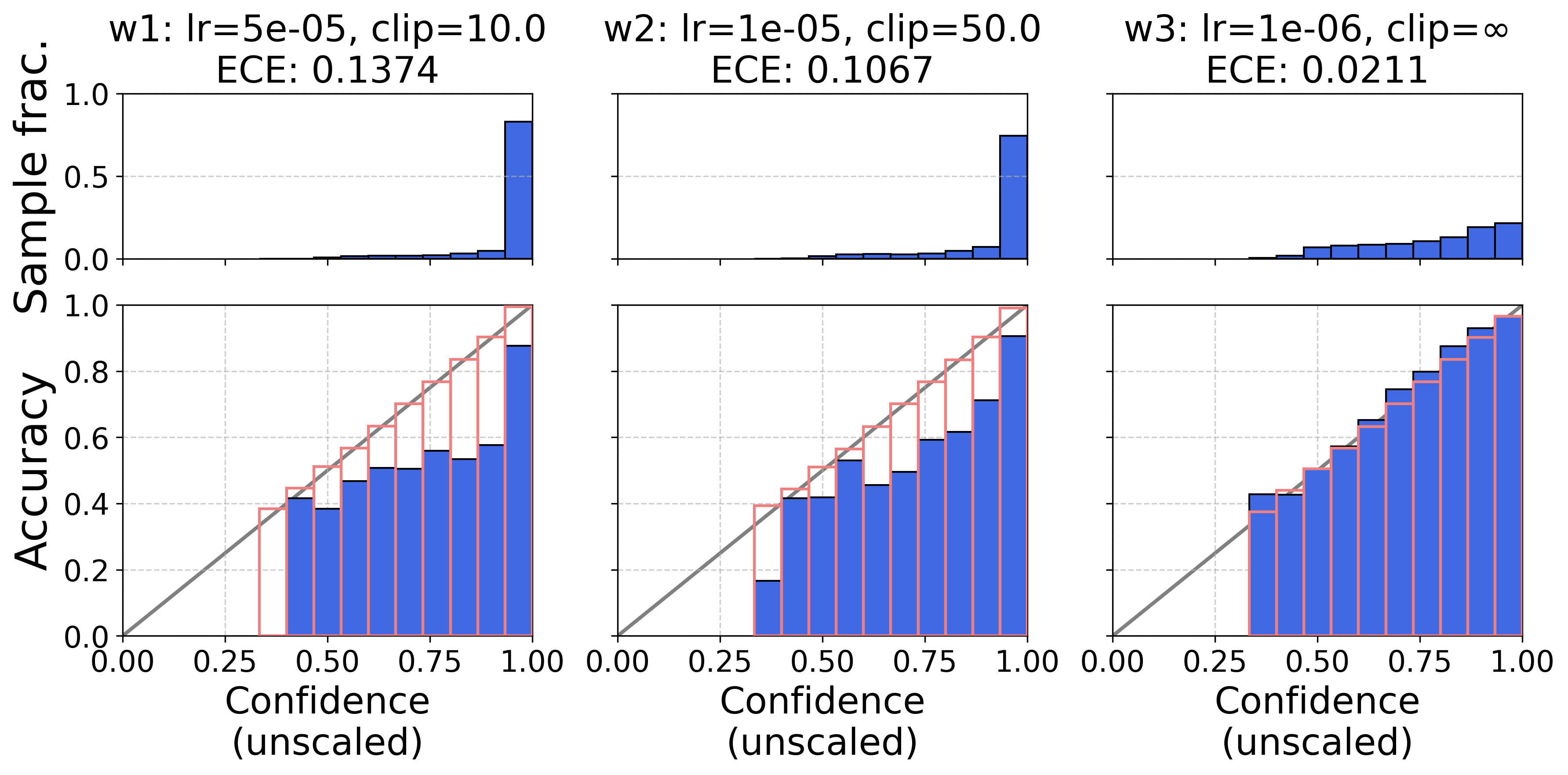}
\put(-135,100){fine-tuned on SNLI}
    \hspace{0.2cm}
\includegraphics[width=0.48\columnwidth]{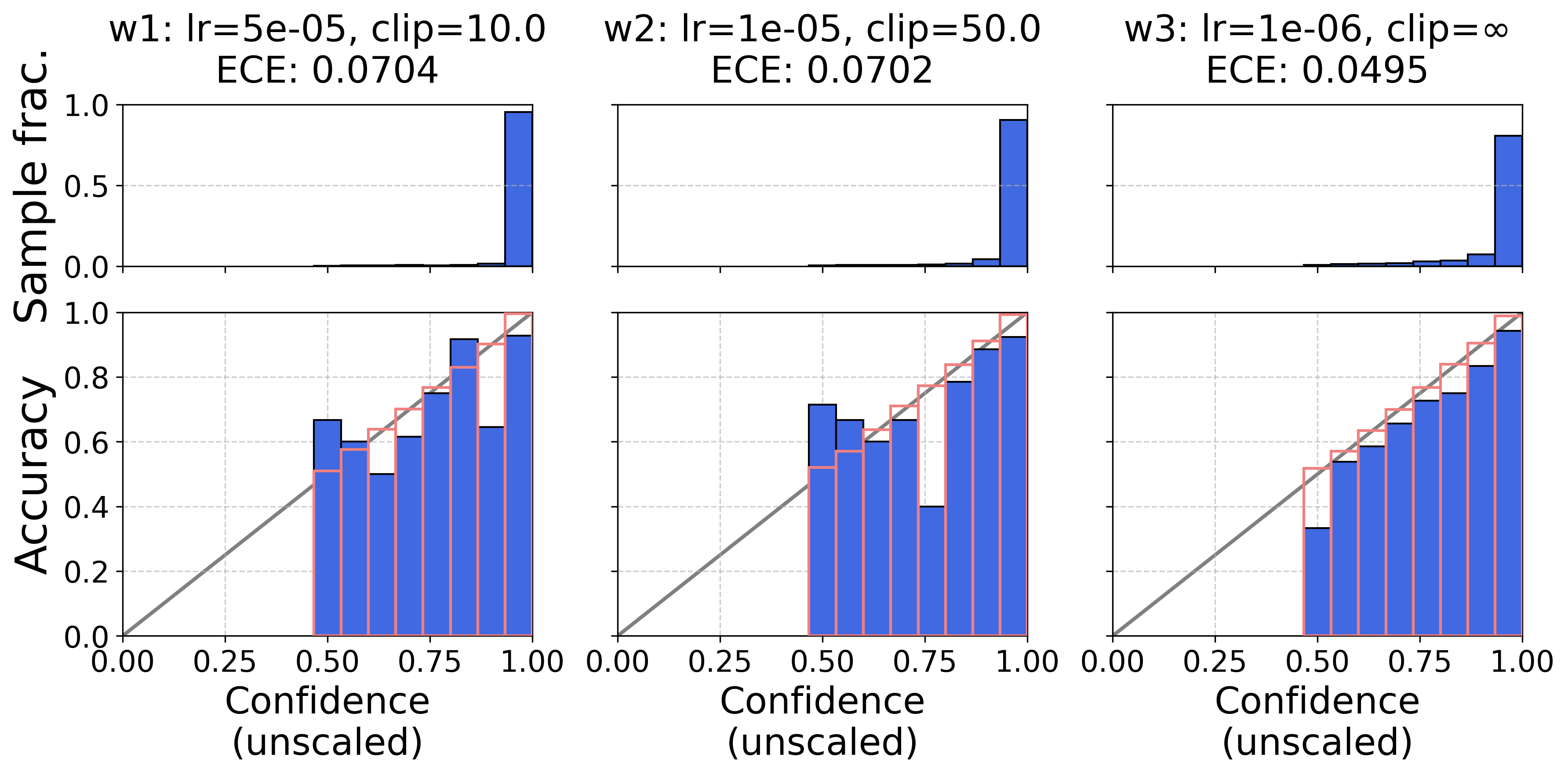} 
\put(-130,100){fine-tuned on SST-2}
\end{center}
\vspace{-3mm}
\caption{Model calibration after fine-tuning on SNLI (left) and SST-2 (right). Reliability diagrams \cite{degroot1983comparison} comparing clipped models ($w_1, w_2$) to an unclipped baseline ($w_3$). The clipped models consistently display miscalibration (blue bars away from the red box) and higher Expected Calibration Error (ECE).}
\label{fig:calibration_main}
\end{figure}

To better understand the structural properties of these solutions, we analyze the models' reliability diagrams on SNLI and SST-2 in Figure~\ref{fig:calibration_main}. Following standard evaluation practices \cite{minderer2021revisiting}, we define model confidence for an input $x$ as the maximum predicted probability among all classes, $p^*(x) = \max_y p_w(y|x)$ where $p_w(y|x)$ is the probability assigned to label $y$ on input $x$ for model $w$. A model achieves perfect top-label calibration when this confidence score accurately reflects the true likelihood of the prediction being correct, meaning $\mathbb{P}_{(x,y)\sim {\rm TestSet}}(y \in \arg\max p_w(\cdot|x) \mid |p^*(x) - p| \leq 1/(2m) ) = p$. To construct these reliability diagrams and calculate the Expected Calibration Error (ECE), the dataset of $n$ samples is partitioned into $m=15$ bins based on their predicted confidence scores. Within each bin $B_i$, we compute both the average confidence and the empirical accuracy (the fraction of correct predictions). The overall ECE is then quantified as the expected absolute difference between these two measures across all bins, weighted by the proportion of samples in each \cite{minderer2021revisiting}: 
\begin{equation*}
    \text{ECE} \; =\;   \sum_{i=1}^m \frac{|B_i|}{n} \big| \text{accuracy}(B_i) - \text{confidence}(B_i) \big|\;. 
\end{equation*}

The results in Figure~\ref{fig:calibration_main} suggest that clipping drives the model to a minima that is miscalibrated with higher loss but better accuracy. This is similar to other highly accurate or strongly regularized modern networks that tend toward systematic overconfidence \cite{minderer2021revisiting}. When comparing configurations along the $\eta c$ boundary, the predictive behaviors diverge sharply. The unclipped baseline ($w_3$) remains well-calibrated (SNLI ECE: 0.0211), distributing its confidence mass more evenly across intermediate bins, accurately reflecting the model's true performance. Conversely, models trained with strict clipping ($w_1, w_2$) exhibit much higher ECE scores (0.1374 and 0.1067). Tight clipping concentrates the vast majority of predictions into the top confidence bin ($>0.9$). While standard ZO-SGD finds minima with calibrated confidence, clipping drives the optimization into a fundamentally different solution. 

Figure~\ref{fig:calibration_main} shows that clipping results in many more confident predictions. We hypothesize that without clipping, there is a significant penalty in being confident on wrong predictions, optimization weighs those examples heavily, and the model learns to be more conservative. With clipping, overconfident wrong predictions do not get fixed as fast and the model can be more aggressive in increasing the confidence of easy examples.  This results in the overconfidence of both correct predictions on easy examples and wrong predictions on hard examples. To test this idea, we intentionally plant hard-to-fit examples that have wrong labels in the next section.

\subsection{Injected label noise} 
\label{sec:noise}
\vspace{-2mm}

\begin{figure}[h]
\begin{center}
\includegraphics[width=0.48\columnwidth]{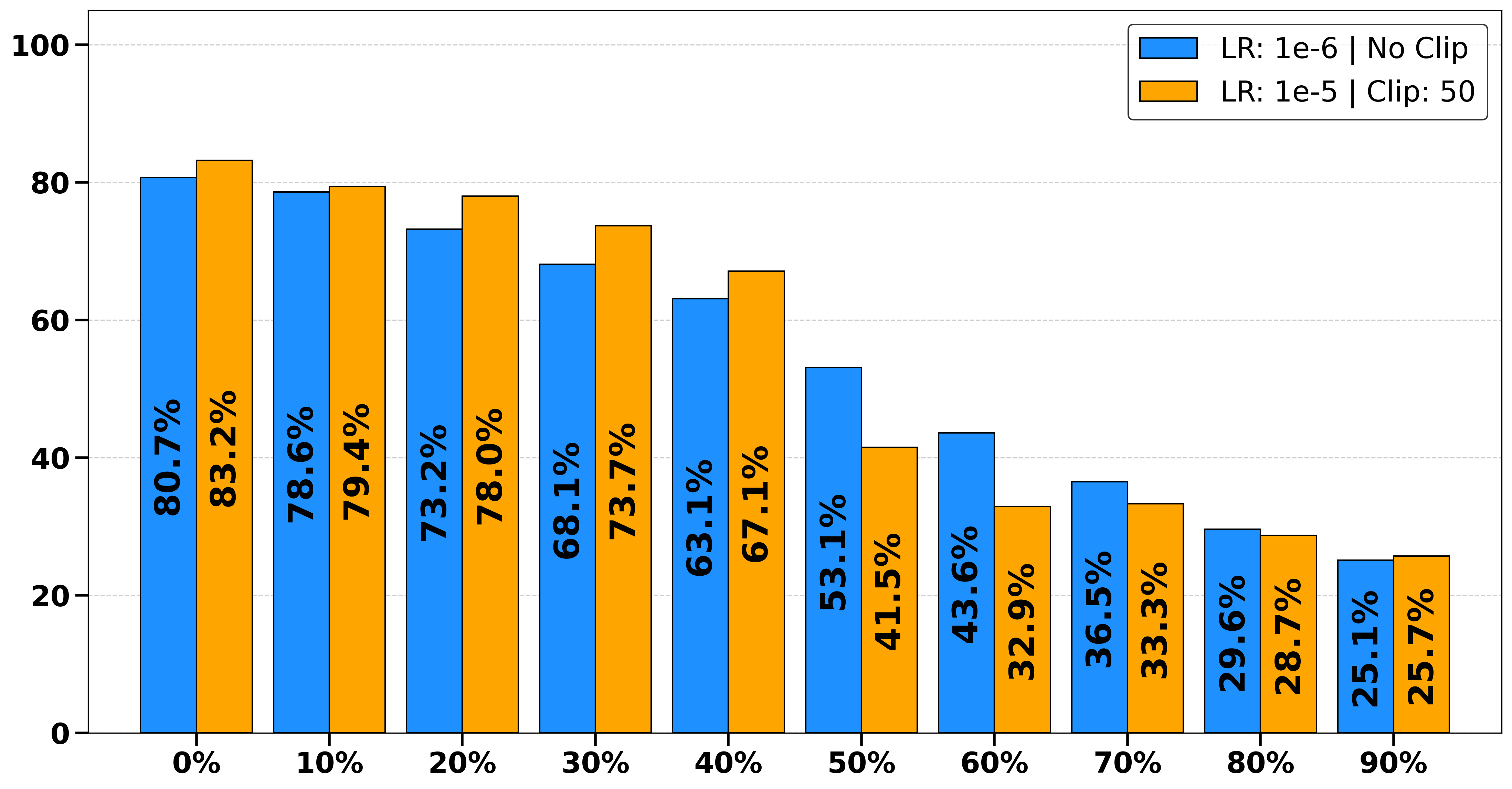} 
    \put(-110,-7){injected noise}
    \put(-198,50){\rotatebox[origin=c]{90}{test accuracy}}
   
    \hspace{0.2cm}
    \includegraphics[width=0.48\columnwidth]{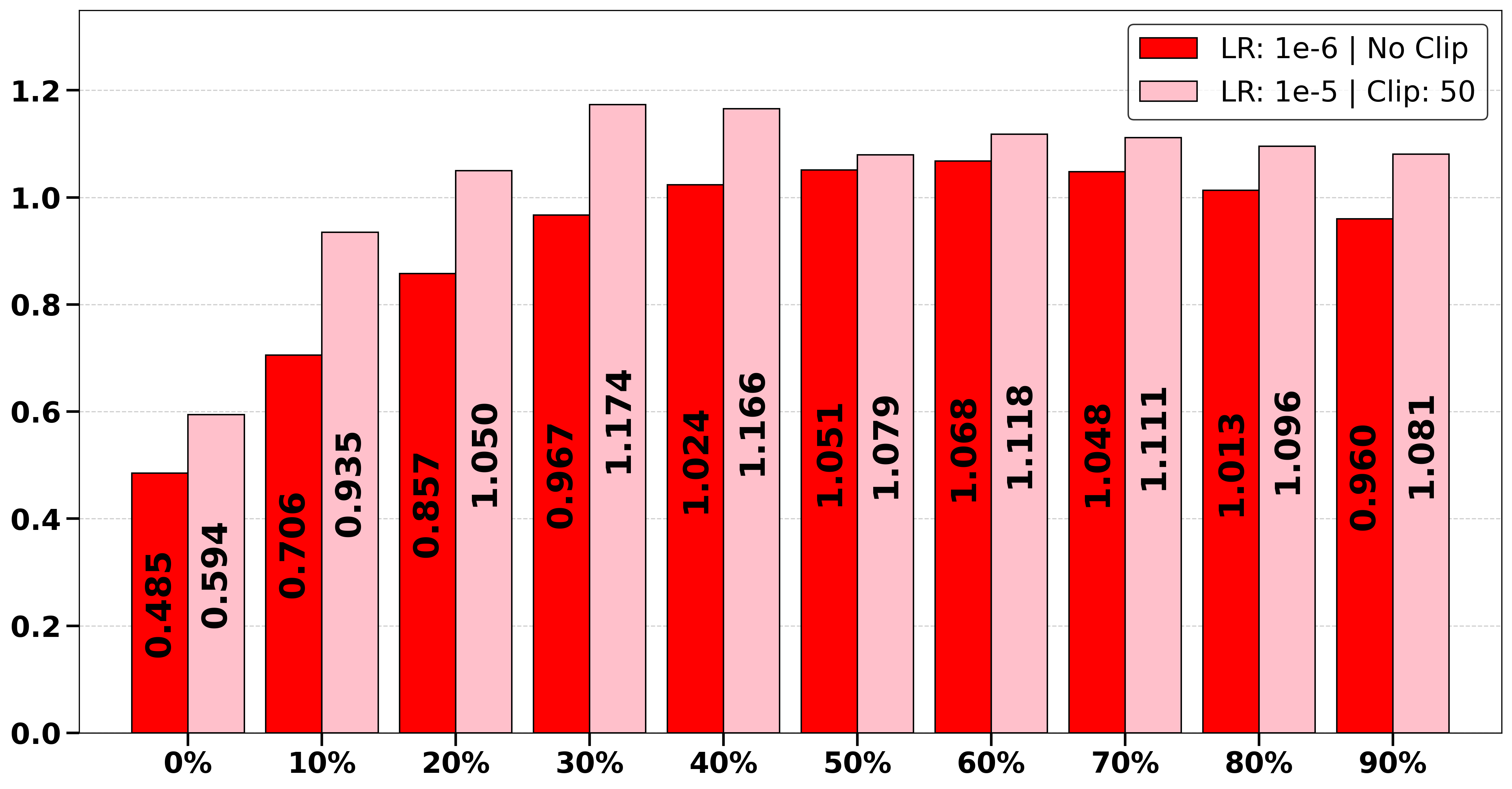} 
    \put(-110,-7){injected noise}
    \put(-198,55){\rotatebox[origin=c]{90}{train loss}}
   
\end{center}
\vspace{-3mm}
\caption{Test accuracy and training loss achieved by models trained with clipped ZO-SGD and unclipped ZO-SGD for various fractions of training data with injected noise. Clipping makes the training more robust to label noise until the injected noise corrupts $50\%$ of the training data.}
\label{fig:noise_diff}
\end{figure}

\begin{figure}[h]
\begin{center}
\includegraphics[width=0.31\columnwidth]{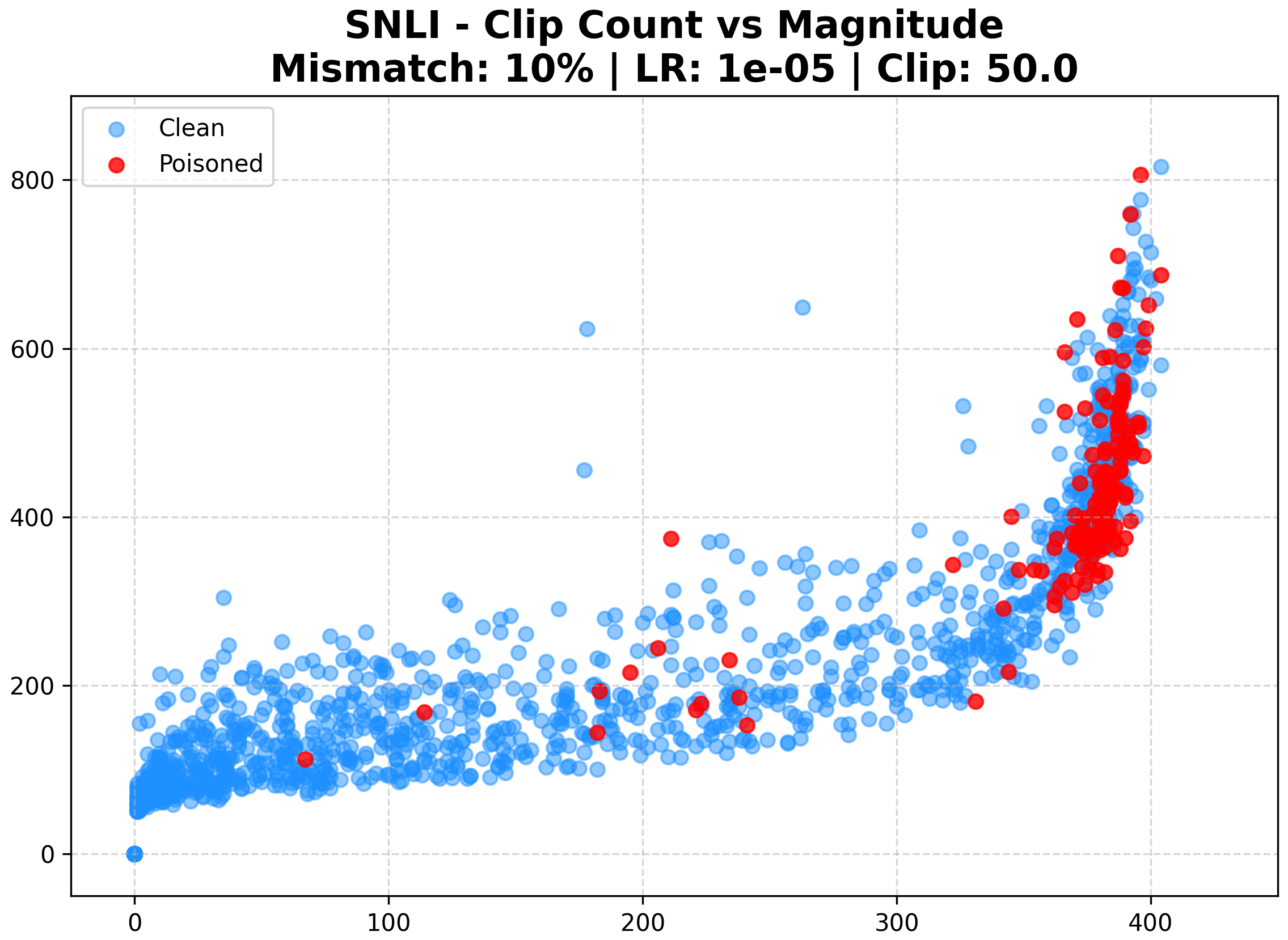}
    \put(-145,40){\rotatebox[origin=c]{90}{average gradient}}
    \put(-135,40){\rotatebox[origin=c]{90}{estimate magnitude}}
\includegraphics[width=0.31\columnwidth]{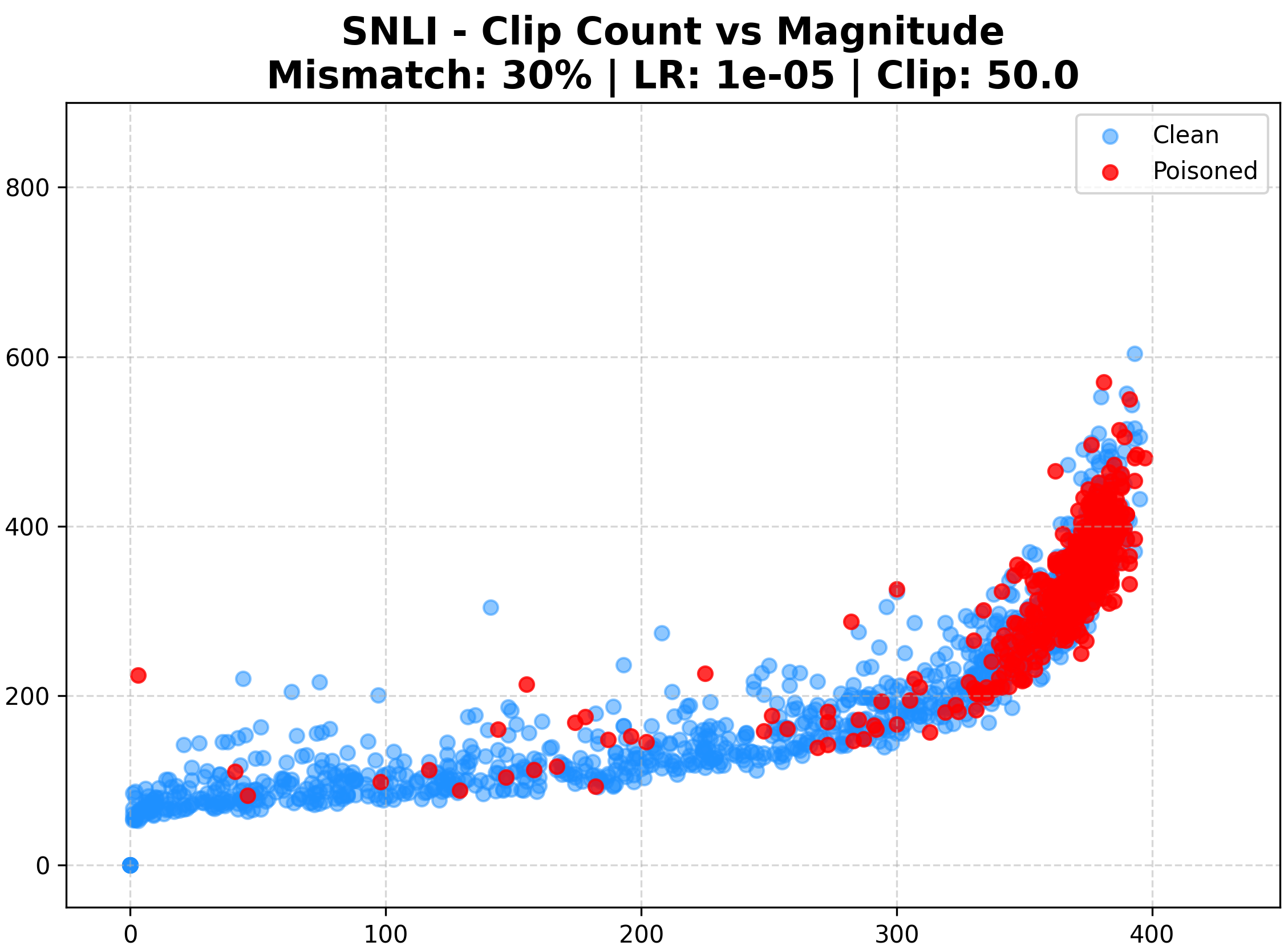}
    \put(-100,-10){total times clipped}
\includegraphics[width=0.31\columnwidth]{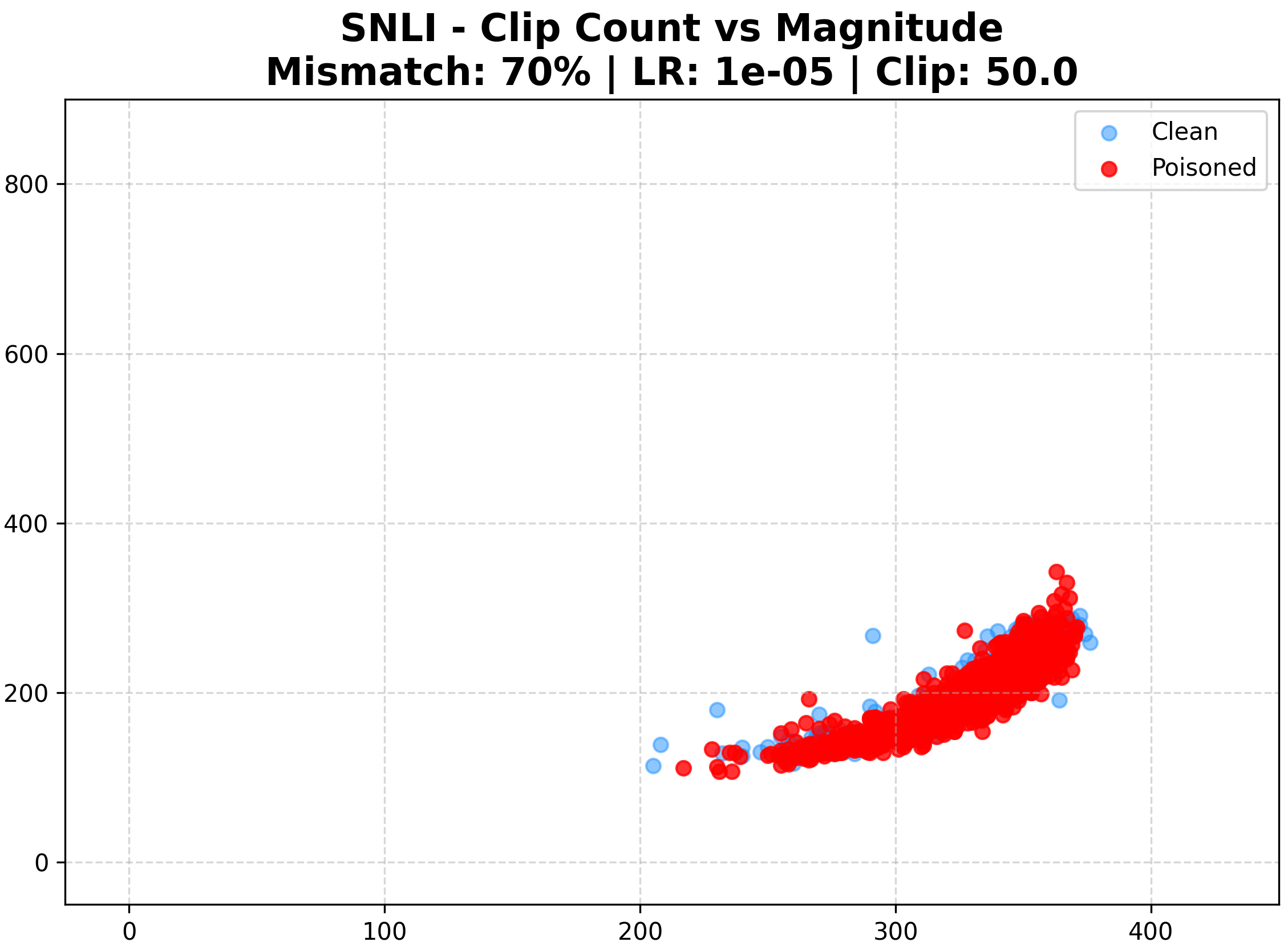}
\end{center}
    \caption{For each of SNLI training example, the average magnitude of the gradient estimates is positively correlated with the number of times the gradient estimate is clipped under clipped ZO-SGD with learning rate $\eta=1e-5$ and clipping threshold $c=50$.  Examples with injected label noise are clearly clipped more frequently due to larger gradient estimates, when noise injection rate is $10\%$ and $30\%$ (left and middle). When $70\%$ of training data is noisy, it becomes ambiguous what the groundtruth labels are and clean and noisy examples behave similarly (right). }
    \label{fig:noise_mag_count}
\end{figure}

To further investigate which samples get clipped in clipped ZO-SGD, we introduce artificial label noise by randomly mislabeling subsets of the training data at varying corruption levels, ranging from 0\% to 90\% as shown in Figure~\ref{fig:noise_diff}. When label noise is injected, the clipped model ($\eta=1\times10^{-5}, c=50.0$) is more robust, maintaining  higher test accuracy than the unclipped baseline. For example, at a 30\% corruption rate, the clipped model achieves a 73.7\% test accuracy compared to the unclipped model's 68.1\%. Generally, this robustness gap between clipped and unclipped ZO-SGD increases as we inject more noisy labels until it reaches $50\%$. Beyond $50\%$, clipped ZO-SGD loses its robustness due to the uncorrupted examples being clipped (Figure~\ref{fig:noise_mag_count}, right). On the other hand, training loss is consistently higher for clipped training, due to the systematic bias that pushes away from minimizing the training loss.

Figure~\ref{fig:noise_diff} shows that, in the regime where injected noise fraction is between $0\%$ and $40\%$, clipped ZO-SGD achieves higher test accuracy. We hypothesize that this is due to robustness of clipping; it down weights hard-to-fit samples, which includes examples with artificially injected label noise, examples with noisy human labels, and examples that are inherently ambiguous what the correct label is (see Figure~\ref{fig:examples} for examples). 
Since we can track the samples with injected noise, we can quantify how many times those examples are clipped as shown in Figure~\ref{fig:noise_mag_count}. This confirms that noise injected examples are clipped significantly more frequently, due to the gradient estimate being larger. This confirms our hypothesis on the origin of robustness when clipping and the resulting accuracy gain. When the injected noise reaches $50\%$ of the training data, the ground-truth labels become ambiguous and both noise-injected and clean examples behave similarly. The accuracy gain from robustness is also eliminated at this point as we see in Figure~\ref{fig:noise_diff}.  

\begin{figure}[h]
\begin{tcolorbox}[title={Mislabeled example, ID: 1093}]
{\small \textbf{Premise:} A blue excavator digging a large hold in cement.\\
\textbf{Hypothesis:} a child was there\\
\textbf{Original SNLI label:} Entailment, 
\textbf{Model predicted:} Contradiction\\
\textbf{Clipping stats:} 389 clips, {\bf Avg magnitude of gradient estimate:} 671.54 }
\end{tcolorbox}
\begin{tcolorbox}[title={Ambiguous example, ID: 1312}]
{\small \textbf{Premise:} Asian city scene of people in street with bright lights and glass buildings behind.\\
\textbf{Hypothesis:} There are people in front of buildings that are brightly lit.\\
\textbf{Original SNLI label:} Entailment, 
\textbf{Model predicted:} Contradiction\\
\textbf{Clipping stats:} 386 clips, {\bf Avg magnitude of gradient estimate:} 622.05}
\end{tcolorbox}
\caption{SNLI training examples that are frequently clipped are mislabeled or ambiguous. Label is for the relation between the premise and the hypothesis. The clipping and magnitude stats are from the $10\%$ corrupted scenario of Figure~\ref{fig:noise_mag_count}, left panel. 
More examples in Appendix~\ref{app:mismatch_examples}.}
\label{fig:examples}
\end{figure}

\subsection{Clipping as implicit data pruning}
\label{sec:prune} 

\begin{figure}[h]

\begin{center}
\includegraphics[width=0.45\columnwidth]{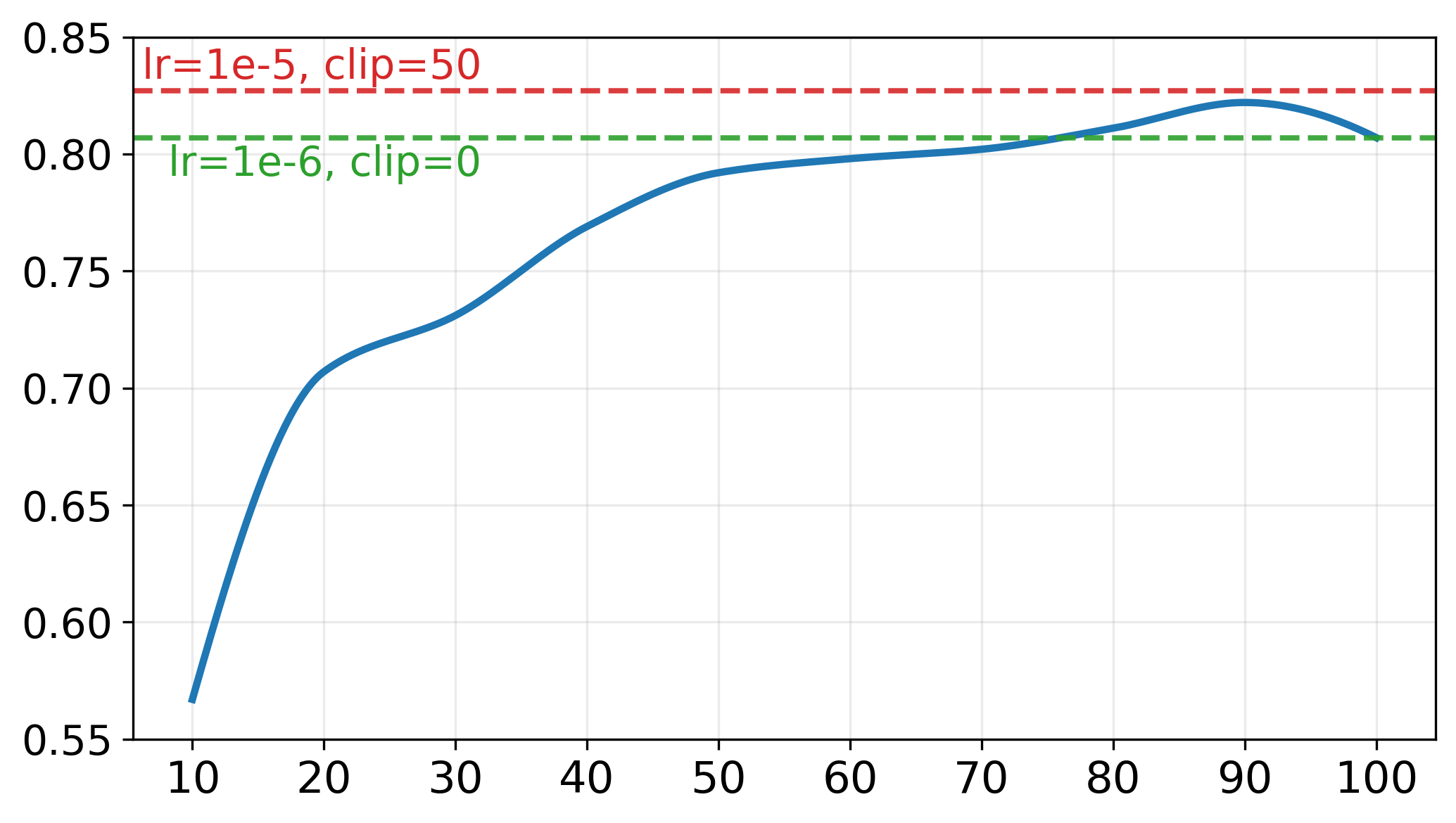}
    \put(-150,-7){percentage of training data used}
    \put(-187,55){\rotatebox[origin=c]{90}{test accuracy}}
    \hspace{0.1in}
    \;\;\;\;
\includegraphics[width=0.45\columnwidth]{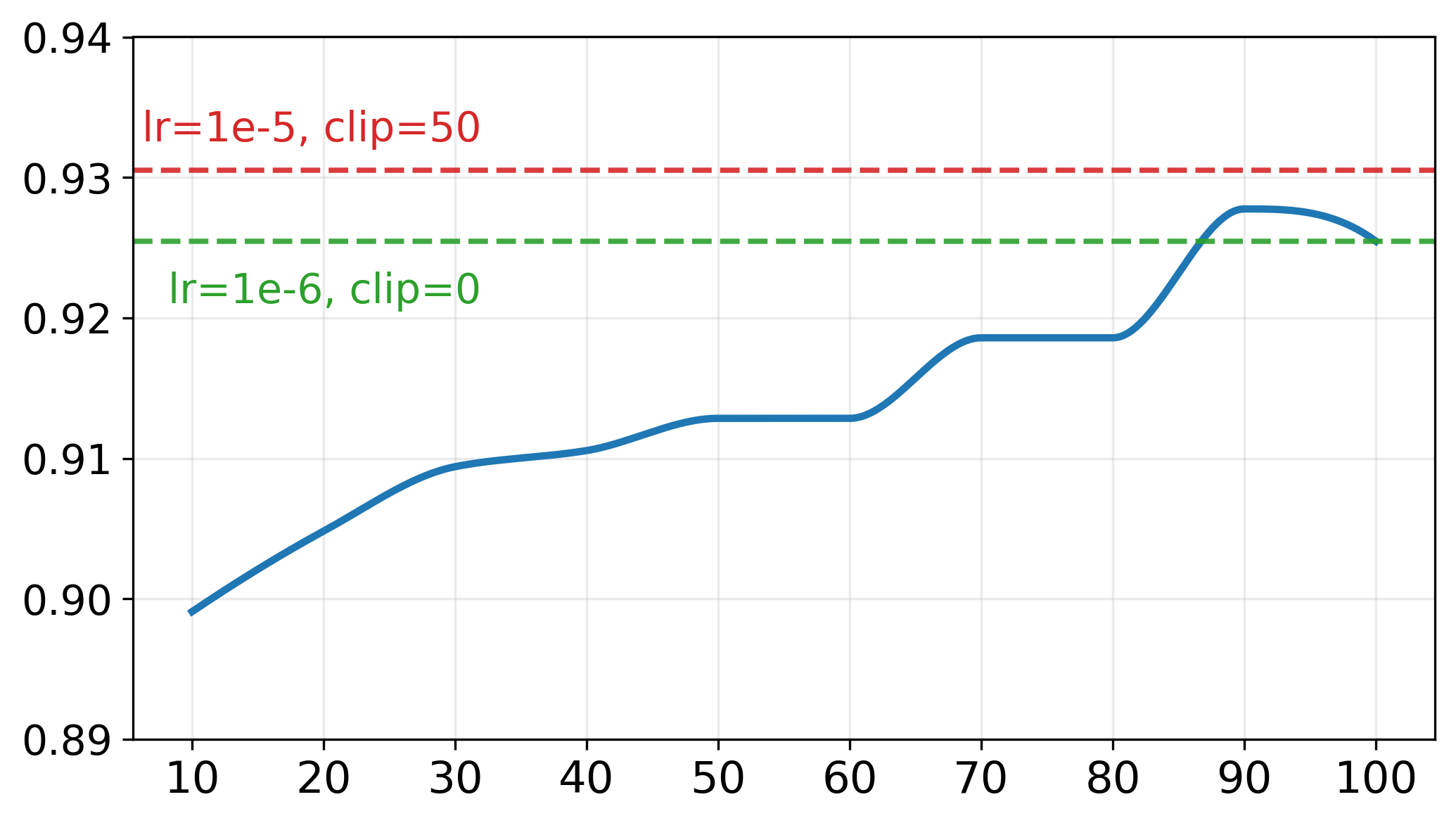}
    \put(-150,-7){percentage of training data used}
    \put(-187,55){\rotatebox[origin=c]{90}{test accuracy}}
\end{center}

\caption{{(left)} SNLI and {(right)} SST-2 test accuracy across subsets of the $a\%$ lowest-loss training examples. The test accuracy of unclipped ZO-SGD peaks at $90\%$ in both cases, corresponding to removing the $10\%$ highest-loss examples. This improves over training on $100\%$ of data with ZO-SGD (green dashed line). Clipped ZO-SGD on $100\%$ of the training data (red dashed line) performs slightly better, but more robustly over the hyperparameter choices and with only one training. 
}
\label{fig:train_data_pct}
\vspace{-4pt}
\end{figure}

The injected label noise experiments in Section~\ref{sec:noise} lead to a hypothesis that the accuracy gain with clipping in Figure~\ref{fig:grid_avg_roberta} (left) originates from down-weighting the noisy examples in the {\em original} training datasets. To test this idea, we design the following data-pruning experiment on the SNLI and SST-2 datasets. In the first phase, we train a model with unclipped ZO-SGD with the entire training dataset, and rank the training examples according to their loss. In the second phase, we again train another model from random initialization using unclipped ZO-SGD, but with a subset of $a\%$ of the training data containing the lowest-loss examples for $10\leq a \leq 100$. As shown in Figure~\ref{fig:train_data_pct}, as we decrease $a$ from $100\%$, perhaps surprisingly, the accuracy increases initially. It reaches its peak when trained with  $90\%$ of data, and then keeps degrading afterwards.   
We believe that there is a certain fraction of ambiguous or mislabeled examples in the original data, which achieve relatively higher loss in the first phase of training. Discarding such examples produces a better performing model.  
We can interpret clipped ZO-SGD as adaptively detecting and down-weighting such examples, without the need for tuning the hyperparameter $a$  or the need for training twice. This is supported by the fact that in Figure~\ref{fig:train_data_pct}, the clipped accuracy in dashed red improves upon the peak performance with pruning.

\sg{Taken together, Sections~\ref{sec:calibrate}--\ref{sec:prune} explain how down-weighting mislabeled examples versus ambiguous boundary examples contributes to the test accuracy improvement and calibration shift. As shown by the uncorrupted high-clip examples in Figure~\ref{fig:noise_mag_count} (left, blue points) and Figure~\ref{fig:examples}, frequently clipped samples consist of both mislabeled outliers (e.g., ID 1093) and genuinely ambiguous boundary examples (e.g., ID 1312). Down-weighting \emph{mislabeled} examples prevents the model from memorizing wrong targets, driving the robustness and test accuracy gains in Figure~\ref{fig:noise_diff}. Simultaneously, down-weighting \emph{ambiguous} boundary examples, or explicitly pruning the $10\%$ highest-loss examples in Figure~\ref{fig:train_data_pct}, prevents hard boundary cases from skewing the classifier and improves the test accuracy on the original dataset. However, because boundary examples normally penalize overconfident predictions, capping their gradients allows the model to push the remaining clean examples into the $>0.9$ confidence bin, producing the overconfidence observed in Figure~\ref{fig:calibration_main}. We explicitly verify this separation in Appendix~\ref{sec:additional_analysis} and Appendix~\ref{sec:abc_disentangle} by partitioning the training data into mislabeled ($A$), high-clip ambiguous ($B$), and low-clip clean ($C$) subsets (Table~\ref{tab:abc_disentangle} and Figures~\ref{fig:calibration10}--\ref{fig:abc_calibration}), confirming that removing $A$ and $B$ each independently improves test accuracy while removing $B$ causes even unclipped ZO-SGD to become overconfident.}

\section{Conclusion} 
\label{Sec:conclusion} 
\vspace{-3mm}
Zeroth-order optimization is emerging as an important resource-efficient tool for fine-tuning LLMs, due to the fact that they do not require backpropagation and hence require less memory. Given its increasing relevance, we systematically investigate a recently observed phenomenon where clipping each gradient estimate in ZO-SGD provides a better accuracy in both train and test data. We first show that this phenomenon is robustly observed in four fine-tuning tasks and two families of pretrained LLMs. We demonstrate that, as predicted by the asymptotic model update in the \sg{\em{clipping-dominated} regime}, the optimal choice of step size $\eta$ is inversely proportional to the clipping threshold $c$, and as long as we keep $\eta c$ fixed, the clipping gain persists for a wide range of these hyperparameters. In this regime, the model converges to a solution different from the unclipped ZO-SGD, and achieves higher accuracy. 

We further show that this phenomenon is unique to per-sample clipped ZO-SGD, by comparing it to other robust gradient aggregation methods applied to ZO-SGD. Our experiments with artificially injected label noise and retraining with pruned training data suggest that the gain is coming from the inherent down weighting of hard-to-fit examples in clipped ZO-SGD. \sg{An interesting avenue for future work is to further investigate the optimization dynamics of zeroth-order fine-tuning through the lens of Edge of Stability (EoS) phenomena and sharpness of the found model weight \cite{song2026zeroth, zhang2025zerothorderoptimizationfindsflat}. As a preliminary exploration, we examine the Hessian trace dynamics in Appendix \ref{app:eos}. Because EoS typically dictates the optimization dynamics of {\em randomly initialized} networks, our evaluations suggest that pretrained language models already reside in relatively flat basins and do not reach the typical EoS threshold during fine-tuning. Analyzing this distinct behavior under zeroth-order optimization remains an open direction for future research.}


\section*{Acknowledgment}
This work is supported by NSF awards 2112471, 2229876, 2505865, and 2502281. 

\section*{AI Usage}
In this work, we used generative AI tools to correct grammar and spelling errors, improve readability, and assist in developing the experimental codebase (specifically for debugging and adapting to resource limitations), and identify relevant literature. We have not used generative AI tools to generate synthetic datasets, develop theoretical models or conceptual frameworks, 
propose hypotheses, design experiments, clean datasets, or interpret results, and the remaining required disclosure tasks (such as translation and qualitative data analysis) are not applicable to this work.

We have carefully reviewed all AI-assisted work. Any code written or modified by LLMs was manually verified,  tested, and executed by the authors to guarantee correctness. We also independently read and verified the accuracy and relevance of all AI-suggested literature before citing it in this paper. We take responsibility for the final content of this work, including text, claims, or artifacts produced with the aid of generative AI.

\section*{Reproducibility statement}

To ensure the full reproducibility of our findings, we have detailed our theoretical and experimental frameworks throughout the paper and its appendices. Additionally we will make the complete codebase publicly available at the release of the camera ready version. The complete mathematical proof for the asymptotic behavior of the clipped gradient estimator (Proposition \ref{propo:clippedgradient}) is provided in Appendix \ref{app:proof} . Full details of our experimental setup, including dataset processing for SNLI, MNLI, SST-2, and TREC, model specifications for RoBERTa and OPT, and exact hyperparameter sweep ranges for the learning rate and clipping threshold, are explicitly documented in Appendix \ref{sec:experimental_setup}. The step-by-step formulation for the Winsorized mean baseline is detailed in Appendix \ref{app:winsorized_estimator}. Exhaustive learning curves, loss landscapes, and hyperparameter configurations for all specific tasks are available in Appendices \ref{app:add_roberta}, \ref{app:opt}, and \ref{app:sec3}. Finally, the source code required to execute all Zeroth-Order fine-tuning and label noise experiments will be provided in the supplementary materials. Our implementation builds directly upon the original DPZero and ZO codebases, from \cite{zhang2024dpzero} and \cite{song2026zeroth}, and includes all custom add-ons to that existing framework necessary to reproduce the specific experiments presented in this paper.

\bibliography{iclr2027_conference_rebuttal}

@article{nesterov2017random,
  title={Random gradient-free minimization of convex functions},
  author={Nesterov, Yurii and Spokoiny, Vladimir},
  journal={Foundations of Computational Mathematics},
  volume={17},
  number={2},
  pages={527--566},
  year={2017},
  publisher={Springer}
}

@inproceedings{abadi2016deep,
  title={Deep learning with differential privacy},
  author={Abadi, Martin and Chu, Andy and Goodfellow, Ian and McMahan, H Brendan and Mironov, Ilya and Talwar, Kunal and Zhang, Li},
  booktitle={Proceedings of the 2016 ACM SIGSAC conference on computer and communications security},
  pages={308--318},
  year={2016}
}

@article{liu2019roberta,
  title={RoBERTa: A robustly optimized BERT pretraining approach},
  author={Liu, Yinhan and Ott, Myle and Goyal, Naman and Du, Jingfei and Joshi, Mandar and Chen, Danqi and Levy, Omer and Lewis, Mike and Zettlemoyer, Luke and Stoyanov, Veselin},
  journal={arXiv preprint arXiv:1907.11692},
  year={2019}
}

@article{zhang2022opt,
  title={OPT: Open pre-trained Transformer language models},
  author={Zhang, Susan and Roller, Stephen and Goyal, Naman and Artetxe, Mikel and Chen, Moya and Chen, Shuohui and Dewan, Christopher and Diab, Mona and Li, Xian and Lin, Xi Victoria and others},
  journal={arXiv preprint arXiv:2205.01068},
  year={2022}
}

@inproceedings{cohen2021gradient,
  title={Gradient Descent on Neural Networks Typically Occurs at the Edge of Stability},
  author={Cohen, Jeremy M and Kaur, Simran and Li, Yuanzhi and Kolter, J Zico and Talwalkar, Ameet},
  booktitle={International Conference on Learning Representations},
  year={2021}
}

@inproceedings{ahn2022understanding,
  title={Understanding the unstable convergence of gradient descent},
  author={Ahn, Kwangjun and Zhang, Jingzhao and Sra, Suvrit},
  booktitle={International Conference on Machine Learning},
  year={2022}
}

@article{ghadimi2013stochastic,
  title={Stochastic first-and zeroth-order methods for nonconvex stochastic programming},
  author={Ghadimi, Saeed and Lan, Guanghui},
  journal={SIAM Journal on Optimization},
  volume={23},
  number={4},
  pages={2341--2368},
  year={2013},
  publisher={SIAM}
}

@article{liu2020primer,
  title={A primer on zeroth-order optimization in signal processing and machine learning},
  author={Liu, Sijia and Chen, Pin-Yu and Kailkhura, Bhavya and Zhang, Gaoyuan and Hero III, Alfred O and Varshney, Pramod K},
  journal={IEEE Signal Processing Magazine},
  volume={37},
  number={5},
  pages={43--54},
  year={2020},
  publisher={IEEE}
}

@article{song2026zeroth,
  title={Zeroth-Order Optimization at the Edge of Stability},
  author={Song, Minhak and Zhang, Liang and Li, Bingcong and He, Niao and Muehlebach, Michael and Oh, Sewoong},
  journal={arXiv preprint arXiv:2604.14669},
  year={2026}
}

@inproceedings{zhang2024dpzero,
  title={{DPZero}: Private Fine-Tuning of Language Models without Backpropagation},
  author={Zhang, Liang and Li, Bingcong and Thekumparampil, Kiran Koshy and Oh, Sewoong and He, Niao},
  booktitle={International Conference on Machine Learning (ICML)},
  year={2024}
}

@article{shamir2017optimal,
  title={An optimal algorithm for bandit and zero-order convex optimization with two-point feedback},
  author={Shamir, Ohad},
  journal={Journal of Machine Learning Research},
  volume={18},
  number={1},
  pages={1703--1713},
  year={2017}
}

@article{lin2022gradient,
  title={Gradient-free methods for deterministic and stochastic nonsmooth nonconvex optimization},
  author={Lin, Tianyi and Zheng, Zeyu and Jordan, Michael},
  journal={Advances in Neural Information Processing Systems},
  volume={35},
  pages={26160--26175},
  year={2022}
}

@inproceedings{malladi2023fine,
  title={Fine-tuning language models with just forward passes},
  author={Malladi, Sadhika and Gao, Tianyu and Nichani, Eshaan and Damian, Alex and Lee, Jason D and Chen, Danqi and Arora, Sanjeev},
  booktitle={Advances in Neural Information Processing Systems},
  volume={36},
  pages={53038--53075},
  year={2023}
}

@inproceedings{zhang2024b,
  title={Revisiting zeroth-order optimization for memory-efficient {LLM} fine-tuning: A benchmark},
  author={Zhang, Y and Li, P and Hong, J and Li, J and Zhang, Y and Zheng, W and Chen, P-Y and Lee, JD and Yin, W and Hong, M and others},
  booktitle={International Conference on Machine Learning},
  year={2024}
}

@inproceedings{arora2022understanding,
  title={Understanding gradient descent on the edge of stability in deep learning},
  author={Arora, S and Li, Z and Panigrahi, A},
  booktitle={International Conference on Machine Learning},
  year={2022}
}

@inproceedings{damian2023self,
  title={Self-stabilization: The implicit bias of gradient descent at the edge of stability},
  author={Damian, A and Nichani, E and Lee, JD},
  booktitle={International Conference on Learning Representations},
  year={2023}
}

@inproceedings{chen2020understanding,
  title={Understanding gradient clipping in private {SGD}: A geometric perspective},
  author={Chen, Xiangyi and Wu, Steven Z and Hong, Mingyi},
  booktitle={Advances in Neural Information Processing Systems},
  volume={33},
  pages={13773--13782},
  year={2020}
}

@inproceedings{fang2023improved,
  title={Improved convergence of differential private {SGD} with gradient clipping},
  author={Fang, Huang and Li, Xiaoyun and Fan, Chenglin and Li, Ping},
  booktitle={International Conference on Learning Representations},
  year={2023}
}

@inproceedings{koloskova2023revisiting,
  title={Revisiting gradient clipping: Stochastic bias and tight convergence guarantees},
  author={Koloskova, Anastasia and Hendrikx, Hadrien and Stich, Sebastian U},
  booktitle={International Conference on Machine Learning},
  year={2023}
}

@inproceedings{bowman2015large,
  title={A large annotated corpus for learning natural language inference},
  author={Bowman, Samuel R and Angeli, Gabor and Potts, Christopher and Manning, Christopher D},
  booktitle={Proceedings of the 2015 Conference on Empirical Methods in Natural Language Processing},
  pages={632--642},
  year={2015}
}

@inproceedings{williams2018broad,
  title={A broad-coverage challenge corpus for sentence understanding through inference},
  author={Williams, Adina and Nangia, Nikita and Bowman, Samuel},
  booktitle={Proceedings of the 2018 Conference of the North American Chapter of the Association for Computational Linguistics: Human Language Technologies, Volume 1 (Long Papers)},
  pages={1112--1122},
  year={2018}
}

@inproceedings{socher2013recursive,
  title={Recursive deep models for semantic compositionality over a sentiment treebank},
  author={Socher, Richard and Perelygina, Alex and Wu, Jean and Chuang, Jason and Manning, Christopher D and Ng, Andrew Y and Potts, Christopher},
  booktitle={Proceedings of the 2013 conference on empirical methods in natural language processing},
  pages={1631--1642},
  year={2013}
}

@inproceedings{li2002learning,
  title={Learning question classifiers},
  author={Li, Xin and Roth, Dan},
  booktitle={COLING 2002: The 19th International Conference on Computational Linguistics},
  year={2002}
}

@inproceedings{ainsworthgit,
  title={Git Re-Basin: Merging Models modulo Permutation Symmetries},
  author={Ainsworth, Samuel and Hayase, Jonathan and Srinivasa, Siddhartha},
  booktitle={The Eleventh International Conference on Learning Representations}, year={2023}
}

@inproceedings{wangeliminating,
  title={Eliminating Sharp Minima from SGD with Truncated Heavy-tailed Noise},
  author={Wang, Xingyu and Oh, Sewoong and Rhee, Chang-Han},
  booktitle={International Conference on Learning Representations}, year={2022}
}

@inproceedings{nasery2025pleas,
  title={PLeaS—Merging Models with Permutations and Least Squares},
  author={Nasery, Anshul and Hayase, Jonathan and Koh, Pang Wei and Oh, Sewoong},
  booktitle={2025 IEEE/CVF Conference on Computer Vision and Pattern Recognition (CVPR)},
  pages={30493--30502},
  year={2025},
  organization={IEEE}
}

@inproceedings{zhanggradient,
  title={Why Gradient Clipping Accelerates Training: A Theoretical Justification for Adaptivity},
  author={Zhang, Jingzhao and He, Tianxing and Sra, Suvrit and Jadbabaie, Ali},
  booktitle={International Conference on Learning Representations}, year={2020}
}

@inproceedings{wortsman2022model,
  title={Model soups: averaging weights of multiple fine-tuned models improves accuracy without increasing inference time},
  author={Wortsman, Mitchell and Ilharco, Gabriel and Gadre, Samir Ya and Roelofs, Rebecca and Gontijo-Lopes, Raphael and Morcos, Ari S and Namkoong, Hongseok and Farhadi, Ali and Carmon, Yair and Kornblith, Simon and   Schmidt, Ludwig},
  booktitle={International conference on machine learning},
  pages={23965--23998},
  year={2022},
  organization={PMLR}
}

@inproceedings{chen2017zoo,
  title={Zoo: Zeroth order optimization based black-box attacks to deep neural networks without training substitute models},
  author={Chen, Pin-Yu and Zhang, Huan and Sharma, Yash and Yi, Jinfeng and Hsieh, Cho-Jui},
  booktitle={Proceedings of the 10th ACM workshop on artificial intelligence and security},
  pages={15--26},
  year={2017}
}

@inproceedings{tu2019autozoom,
  title={Autozoom: Autoencoder-based zeroth order optimization method for attacking black-box neural networks},
  author={Tu, Chun-Chen and Ting, Paishun and Chen, Pin-Yu and Liu, Sijia and Zhang, Huan and Yi, Jinfeng and Hsieh, Cho-Jui and Cheng, Shin-Ming},
  booktitle={Proceedings of the AAAI conference on artificial intelligence},
  volume={33},
  number={01},
  pages={742--749},
  year={2019}
}

@article{duchi2015optimal,
  title={Optimal rates for zero-order convex optimization: The power of two function evaluations},
  author={Duchi, John C and Jordan, Michael I and Wainwright, Martin J and Wibisono, Andre},
  journal={IEEE Transactions on Information Theory},
  volume={61},
  number={5},
  pages={2788--2806},
  year={2015},
  publisher={IEEE}
}

@inproceedings{szegedy2016rethinking,
  title={Rethinking the inception architecture for computer vision},
  author={Szegedy, Christian and Vanhoucke, Vincent and Ioffe, Sergey and Shlens, Jon and Wojna, Zbigniew},
  booktitle={Proceedings of the IEEE conference on computer vision and pattern recognition},
  pages={2818--2826},
  year={2016}
}

@article{ye2018hessian,
  title={Hessian-aware zeroth-order optimization for black-box adversarial attack},
  author={Ye, Haishan and Huang, Zhichao and Fang, Cong and Li, Chris Junchi and Zhang, Tong},
  journal={arXiv preprint arXiv:1812.11377},
  year={2018}
}

@inproceedings{minderer2021revisiting,
  title={Revisiting the Calibration of Modern Neural Networks},
  author={Minderer, Matthias and Djolonga, Josip and Romijnders, Rob and Hubis, Frances Ann and Zhai, Xiaohua and Houlsby, Neil and Tran, Dustin and Lu{\v{c}}i{\'c}, Mario},
  booktitle={Advances in Neural Information Processing Systems},
  volume={34},
  pages={15682--15694},
  year={2021},
  publisher={Curran Associates, Inc.}
}

@misc{zhang2025zerothorderoptimizationfindsflat,
      title={Zeroth-Order Optimization Finds Flat Minima}, 
      author={Liang Zhang and Bingcong Li and Kiran Koshy Thekumparampil and Sewoong Oh and Michael Muehlebach and Niao He},
      year={2025},
      eprint={2506.05454},
      archivePrefix={arXiv},
      primaryClass={cs.LG},
      url={https://arxiv.org/abs/2506.05454}, 
}

@inproceedings{garipov2018loss,
  title={Loss surfaces, mode connectivity, and fast ensembling of DNNs},
  author={Garipov, Timur and Izmailov, Pavel and Podoprikhin, Dmitrii and Vetrov, Dmitry P and Wilson, Andrew G},
  booktitle={Advances in neural information processing systems},
  volume={31},
  year={2018}
}

@inproceedings{li2018visualizing,
  title={Visualizing the loss landscape of neural nets},
  author={Li, Hao and Xu, Zheng and Taylor, Gavin and Studer, Christoph and Goldstein, Tom},
  booktitle={Advances in neural information processing systems},
  volume={31},
  year={2018}
}

@article{liu2018zeroth,
  title={Zeroth-order stochastic variance reduction for nonconvex optimization},
  author={Liu, Sijia and Kailkhura, Bhavya and Chen, Pin-Yu and Ting, Paishun and Chang, Shiyu and Amini, Lisa},
  journal={Advances in neural information processing systems},
  volume={31},
  year={2018}
}

@article{chen2019zo,
  title={Zo-adamm: Zeroth-order adaptive momentum method for black-box optimization},
  author={Chen, Xiangyi and Liu, Sijia and Xu, Kaidi and Li, Xingguo and Lin, Xue and Hong, Mingyi and Cox, David},
  journal={Advances in neural information processing systems},
  volume={32},
  year={2019}
}

@article{liu2026position,
  title={Position: Zeroth-Order Optimization in Deep Learning Is Underexplored, Not Underpowered},
  author={Liu, Sijia and Lang, Yicheng and Pal, Soumyadeep and Wang, Changsheng and Huang, Yancheng and Fan, Chongyu and Diffenderfer, James and Kailkhura, Bhavya and Zhang, Yihua},
  journal={arXiv preprint arXiv:2605.15622},
  year={2026}
}

@article{lang2026powering,
  title={Powering up zeroth-order training via subspace gradient orthogonalization},
  author={Lang, Yicheng and Wang, Changsheng and Zhang, Yihua and Hong, Mingyi and Zhang, Zheng and Yin, Wotao and Liu, Sijia},
  journal={arXiv preprint arXiv:2602.17155},
  year={2026}
}

@article{chen2026accelerating,
  title={Accelerating Zeroth-Order Spectral Optimization with Partial Orthogonalization from Power Iteration},
  author={Chen, Jiahe and Ma, Ziye},
  journal={arXiv preprint arXiv:2605.09034},
  year={2026}
}

@article{tan2026grzo,
  title={GRZO: Group-Relative Zeroth-Order Optimization for Large Language Model Fine-Tuning},
  author={Tan, Liyan and Zhao, Yequan and Yang, Yifan and Zhang, Ruijie and Yu, Xinling and Zhang, Zheng},
  journal={arXiv preprint arXiv:2606.02857},
  year={2026}
}

@article{zhang2022robustify,
  title={How to robustify black-box ml models? a zeroth-order optimization perspective},
  author={Zhang, Yimeng and Yao, Yuguang and Jia, Jinghan and Yi, Jinfeng and Hong, Mingyi and Chang, Shiyu and Liu, Sijia},
  journal={arXiv preprint arXiv:2203.14195},
  year={2022}
}

@inproceedings{chen2024deepzero,
  title={Deepzero: Scaling up zeroth-order optimization for deep model training},
  author={Chen, Aochuan and Zhang, Yimeng and Jia, Jinghan and Diffenderfer, James and Parasyris, Konstantinos and Liu, Jiancheng and Zhang, Yihua and Zhang, Zheng and Kailkhura, Bhavya and Liu, Sijia},
  booktitle={International Conference on Learning Representations},
  volume={2024},
  pages={50185--50206},
  year={2024}
}

@inproceedings{ilyas2018black,
  title={Black-box adversarial attacks with limited queries and information},
  author={Ilyas, Andrew and Engstrom, Logan and Athalye, Anish and Lin, Jessy},
  booktitle={International conference on machine learning},
  pages={2137--2146},
  year={2018},
  organization={PMLR}
}

@inproceedings{sun2022black,
  title={Black-box tuning for language-model-as-a-service},
  author={Sun, Tianxiang and Shao, Yunfan and Qian, Hong and Huang, Xuanjing and Qiu, Xipeng},
  booktitle={International Conference on Machine Learning},
  pages={20841--20855},
  year={2022},
  organization={PMLR}
}

@article{wang2025global,
  title={Global Dynamics of Heavy-Tailed SGDs in Nonconvex Loss Landscape: Characterization and Control},
  author={Wang, Xingyu and Rhee, Chang-Han},
  journal={arXiv preprint arXiv:2510.20905},
  year={2025}
}

@article{degroot1983comparison,
  title={The comparison and evaluation of forecasters},
  author={DeGroot, Morris H and Fienberg, Stephen E},
  journal={Journal of the Royal Statistical Society: Series D (The Statistician)},
  volume={32},
  number={1-2},
  pages={12--22},
  year={1983},
  publisher={Wiley Online Library}
}

@misc{neyshabur2021transferredtransferlearning,
      title={What is being transferred in transfer learning?}, 
      author={Behnam Neyshabur and Hanie Sedghi and Chiyuan Zhang},
      year={2021},
      eprint={2008.11687},
      archivePrefix={arXiv},
      primaryClass={cs.LG},
      url={https://arxiv.org/abs/2008.11687}, 
}

@article{rehn2026optimal,
  title={On Optimal Hyperparameters for Differentially Private Deep Transfer Learning},
  author={Rehn, Tilman and others},
  journal={arXiv preprint arXiv:2510.20616},
  year={2026}
}
\bibliographystyle{iclr2027_conference_rebuttal}
\newpage
\appendix

\section{Related Work}
\label{app:related}

\subsection{Zeroth-Order Optimization}
Zeroth-order (ZO) optimization has become increasingly relevant for scenarios where computing exact gradients is either computationally prohibitive or strictly impossible (see \cite{liu2026position} for a survey of latest advances). Classical applications heavily focused on black-box optimization \cite{nesterov2017random, ghadimi2013stochastic, shamir2017optimal, lin2022gradient}, where access to the objective function is limited to forward evaluations. 
Significant algorithmic advances have resulted from this productive line of research, including \cite{Zhang2022robustify, chen2024deepzero, ilyas2018black, sun2022black, tan2026grzo,chen2026accelerating,lang2026powering}. 
Recently, however, there has been a significant push to adapt ZO techniques for memory-efficient training of large neural networks. Because backpropagation requires caching intermediate activations, scaling standard first-order methods to billions of parameters often exceeds available hardware memory. Recent approaches like MeZO \cite{malladi2023fine, zhang2024b, zhang2024dpzero} address this by utilizing a standard two-point ZO estimator, demonstrating that models can be effectively fine-tuned using only forward passes. This dramatically reduces the memory footprint, making ZO an attractive alternative for aligning and adapting modern language models under strict hardware constraints.

\subsection{Gradient Clipping and Private Fine-Tuning}
Beyond standard memory constraints, ZO methods are actively being adapted to solve challenges in differentially private (DP) machine learning. A fundamental mechanism in deep DP learning is gradient clipping, which tightly bounds the sensitivity of model updates to ensure no single training example disproportionately influences the weights \cite{abadi2016deep, chen2020understanding, fang2023improved, koloskova2023revisiting}. 

In traditional DP-SGD, clipping requires the explicit computation and materialization of per-example gradients, which exacerbates the memory bottleneck of backpropagation. To circumvent this, recent work integrates clipping directly into the zeroth-order estimation process. By clipping the scalar loss differences or the forward-pass directional derivatives, algorithms like DPZero \cite{zhang2024dpzero} can bound the sensitivity of the update step without ever performing backpropagation. This use of clipping seamlessly bridges the memory benefits of ZO optimization with the rigorous privacy guarantees of DP, allowing for the private fine-tuning of massive autoregressive architectures that would otherwise be impossible to train with standard first-order clipping techniques.

\sg{Recently, \citet{rehn2026optimal} empirically studied hyperparameter tuning in first-order DP-SGD and observed that small clipping thresholds under Gaussian privacy noise suppress hard examples. Our work complements \citet{rehn2026optimal} by analyzing noise-free zeroth-order optimization ($\sigma=0$), where clipping acts on scalar random-direction finite differences rather than full first-order gradient norms, inducing the smooth reweighting of Proposition~\ref{propo:clippedgradient}, a strict $\eta c = \text{constant}$ trade-off that does not arise in noise-free first-order clipping (Section~\ref{sec:first_order}), and a distinct trade-off between accuracy, loss, and calibration.}

\subsection{The Edge of Stability}
The optimization dynamics of deep neural networks frequently exhibit a non-equilibrium behavior known as the Edge of Stability (EoS). Originally documented in first-order full-batch gradient descent \cite{cohen2021gradient, ahn2022understanding, arora2022understanding, damian2023self}, the EoS phenomenon occurs when a network's training trajectory naturally gravitates toward and oscillates around regions where the maximum eigenvalue of the Hessian (often referred to as the "sharpness") hovers just above $2/\eta$, where $\eta$ is the learning rate.

While first-order stability is dictated almost entirely by this maximum eigenvalue, the dynamics shift significantly under zeroth-order methods. Because ZO methods rely on stochastic gradient estimates, their mean-square linear stability is governed not just by the sharpness, but by the entire Hessian spectrum \cite{song2026zeroth}. Specifically, theoretical bounds show that ZO methods operate at a distinct edge of stability characterized by a dependence on both the maximum eigenvalue and the Hessian trace. When trained with large step sizes, ZO optimization induces an implicit regularization effect that suppresses the growth of the Hessian trace, forcing the model to converge along a modified stability boundary uniquely dictated by the variance of the zeroth-order estimator.

\section{Detailed Formulation of the Winsorized Mean Calculations}
\label{app:winsorized_estimator}

To isolate the specific dynamics induced by static gradient clipping, we benchmark our approach against alternative statistical robustification techniques, most notably the Winsorized mean. Unlike standard clipping, which applies a fixed, data-independent threshold $c$, Winsorization dynamically bounds the finite differences based on the empirical distribution of the current batch.

For a given batch $B_t$ at iteration $t$, we first compute the scalar finite difference for each sample $i \in B_t$ along the random perturbation direction $u_t$:
\begin{equation}
    v_i = \frac{f_i(x_t+\lambda u_t)-f_i(x_t - \lambda u_t)}{2\lambda} \;.
\end{equation}

To apply Winsorization, we sort these scalars to obtain the order statistics:
\begin{equation}
    v_{(1)} \leq v_{(2)} \leq \dots \leq v_{(|B_t|)} \;.
\end{equation}

For a chosen trimming parameter $k$ (where $0 < k < |B_t|/2$), we define the dynamic lower and upper bounds as the $(k+1)$-th and the $(|B_t|-k)$-th order statistics, respectively:
\begin{equation}
    L_t = v_{(k+1)} \quad \text{and} \quad U_t = v_{(|B_t|-k)} \;.
\end{equation}

The $k$-Winsorized method replaces the $k$ lowest values with $L_t$ and the $k$ highest values with $U_t$. The zeroth-order gradient estimator is then computed as the mean of these bounded values:
\begin{equation}
    g_\lambda(x_t) = \left( \frac{1}{|B_t|} \sum_{i\in B_t} {\rm Clip}_{[L_t, U_t]}(v_i) \right) u_t \;,
\end{equation}
where the bounding function is defined as:
\begin{equation}
    {\rm Clip}_{[L_t, U_t]}(a) = 
    \begin{cases} 
      L_t & \text{if } a < L_t \\
      a & \text{if } L_t \leq a \leq U_t \\
      U_t & \text{if } a > U_t 
    \end{cases} \;.
\end{equation}

This formulation ensures that the gradient estimate remains robust to extreme outlier perturbations without completely discarding the mass of those samples, offering a data-dependent alternative to standard threshold clipping.

\section{Proof of Proposition~\ref{propo:clippedgradient}}
\label{app:proof} 

We first prove that for any $g\in{\mathbb R}^d$ and $c>0$, 
\begin{align*}
 {\mathbb E}_{u\sim {\cal N}(0,{\bf I}_{d\times d})} \big[\, {\rm Clip}_c(\langle u,g\rangle) u\,\big] &\; = \; {\mathbb E}_{u\sim {\cal N}(0,{\bf I}_{d\times d})} \big[\, \nabla_u {\rm Clip}_c(\langle u,g\rangle) \,\big] \nonumber \\
 &\;=\; {\mathbb E}_{u\sim {\cal N}(0,{\bf I}_{d\times d})} \big[\, g\, {\bf 1}_{|\langle u,g\rangle|<c}  \,\big] \\
 &\;=\; \big(\,2\Phi (c/\|g\|)-1\,\big)g \;,
\end{align*}
where $\Phi(\cdot)$ is the standard normal CDF. The first line follows from Stein's identity  for a $1$-Lipschitz function ${\rm Clip}_c(\cdot)$. Together with the fact \cite{duchi2015optimal} that 
$$
 \lim_{\lambda\to 0} g_{\lambda,c}(w) \;=\; 
 \langle u, (1/|B_t|) \sum_{i\in B_t} \nabla f_i(w) \rangle u\;,
$$
this proves the desired claim by replacing $g$ with $\nabla f_i(w)$ for each term.

\section{Experimental Setup}
\label{sec:experimental_setup}

The primary goal of our evaluation is to understand how per-sample gradient clipping affects the convergence behavior of Zeroth-Order (ZO) optimization. Our study is motivated by an observation from the original DPZero framework by Zhang et al.~\cite{zhang2024dpzero}. Although their work focused on differential privacy, they noted that clipping also improved optimization in the non-private setting:

\begin{quote}
``...the non-private baseline MeZO also appears to benefit from clipping. For instance, without clipping, the original MeZO encounters non-convergence issues at a stepsize of $5 \times 10^{-6}$. Conversely, incorporating clipping permits the use of larger stepsizes and yields better results. A thorough investigation of this phenomenon is reserved for future research.''
\end{quote}

This paper investigates that phenomenon directly. We adapt the DPZero framework for large language model fine-tuning while removing all privacy-related components by disabling Gaussian noise injection ($\sigma = 0$). This isolates the clipping threshold ($c$) and learning rate ($\eta$) as the only variables under study.

The original DPZero work evaluated both RoBERTa and OPT models using a few-shot learning setup with at most 512 training samples per class. We follow the same setup and evaluate performance across tasks of varying complexity: sentiment analysis on SST-2~\cite{socher2013recursive}, natural language inference on SNLI~\cite{bowman2015large} and MNLI~\cite{williams2018broad}, and question classification on TREC~\cite{li2002learning}.

For each model--dataset pair, we perform a grid search over learning rates ($\eta \in [5\times10^{-7}, 1\times10^{-4}]$) and clipping thresholds ($c \in [1.0, 500.0]$), including an unclipped baseline ($c=\infty$).

We then evaluate several aspects of clipped ZO optimization. First, we study the trade-off between clipping and learning rate by measuring final validation accuracy and convergence speed, allowing us to characterize the observed ``diagonal trend'' between $\eta$ and $c$. Second, we identify failure modes by tracking training loss to determine when models either diverge due to high variance or fail to learn because clipping is too restrictive. Finally, we analyze how clipping shapes the loss landscape by measuring distance from initialization ($\|w_t - w_0\|$) and evaluating linear mode connectivity between checkpoints. Together, these experiments map how different $(\eta, c)$ settings influence convergence behavior and solution geometry.

\section{Additional experimental results for clipped zeroth-order optimization: RoBERTa } 
\label{app:add_roberta}

We provide the train accuracy and loss when sweeping through learning rate $\eta$ and clipping threshold $c$ in the figure below. \\

\label{app:additional}

\begin{figure}[h]
\vspace{1mm}
\centering

\begin{tabular}{ccc}

\begin{overpic}[width=0.31\columnwidth]{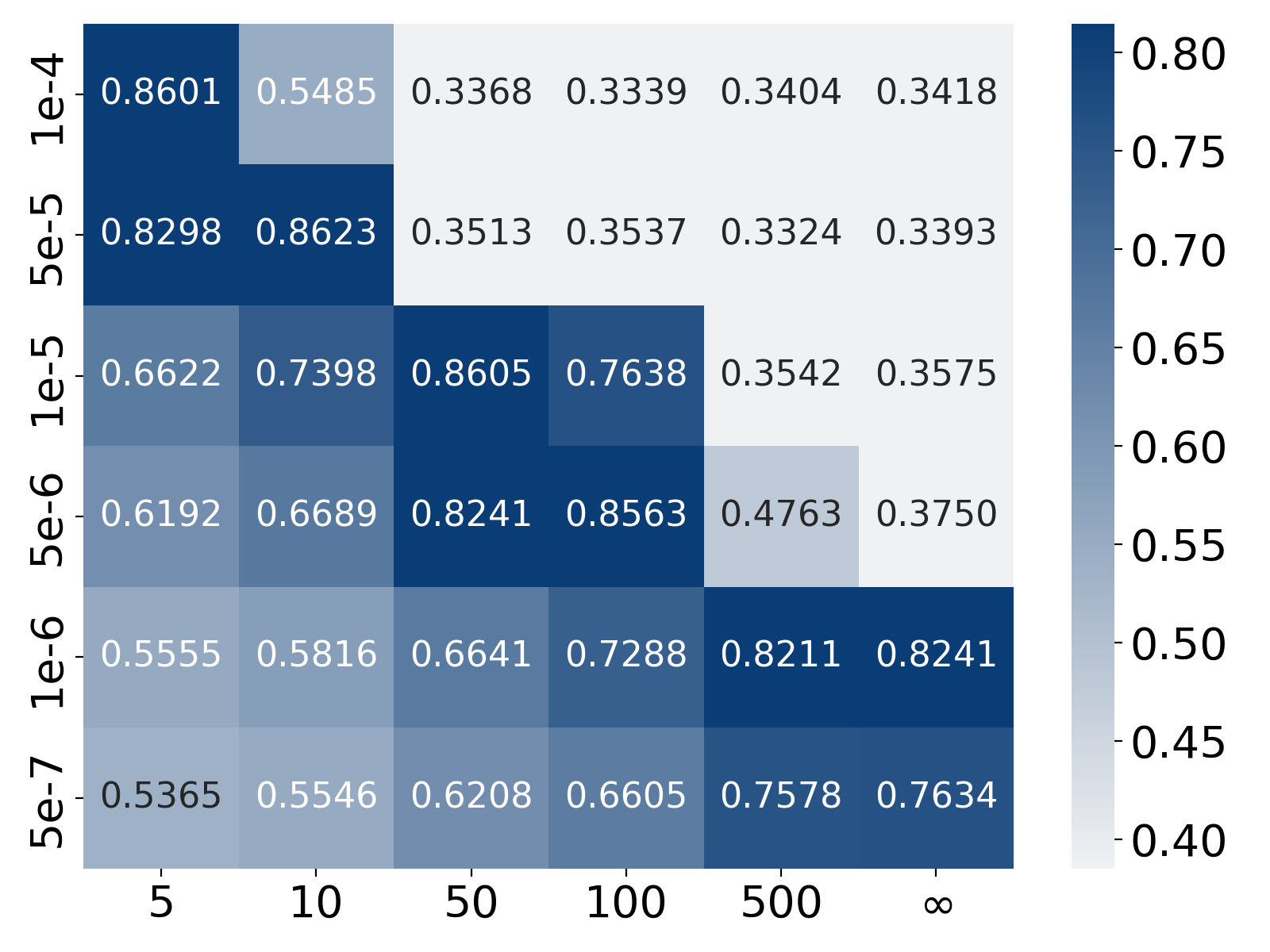}
    \put(13,-5){Clipping threshold $c$}
    \put(17,77){Train accuracy}
    \put(-8,40){\rotatebox[origin=c]{90}{Learning rate $\eta$}}
\end{overpic}
&
\begin{overpic}[width=0.31\columnwidth]{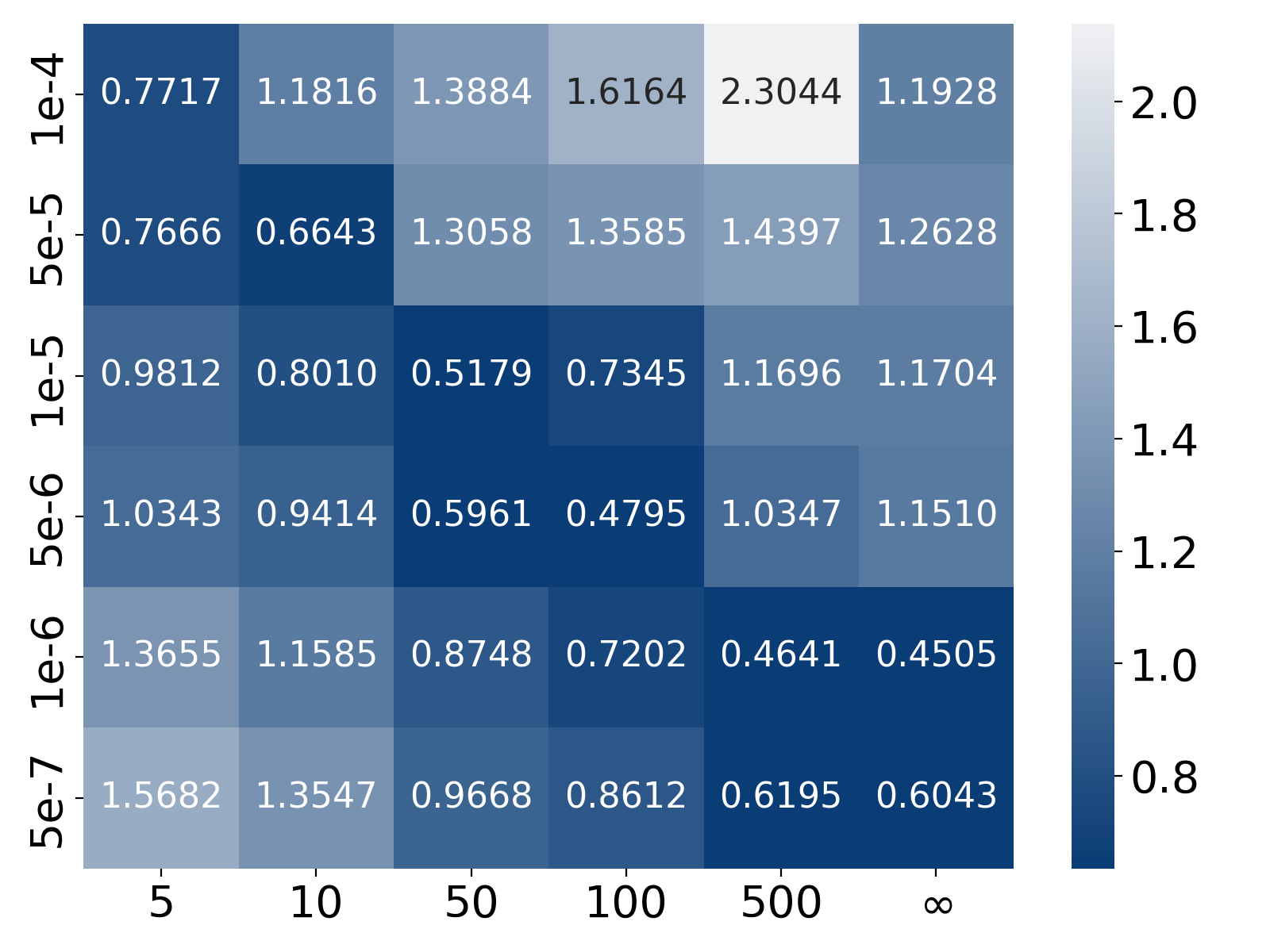}
    \put(13,-5){Clipping threshold $c$}
    \put(25,77){Train loss}
    \put(-8,40){\rotatebox[origin=c]{90}{Learning rate $\eta$}}
\end{overpic}
&
\begin{overpic}[width=0.31\columnwidth]{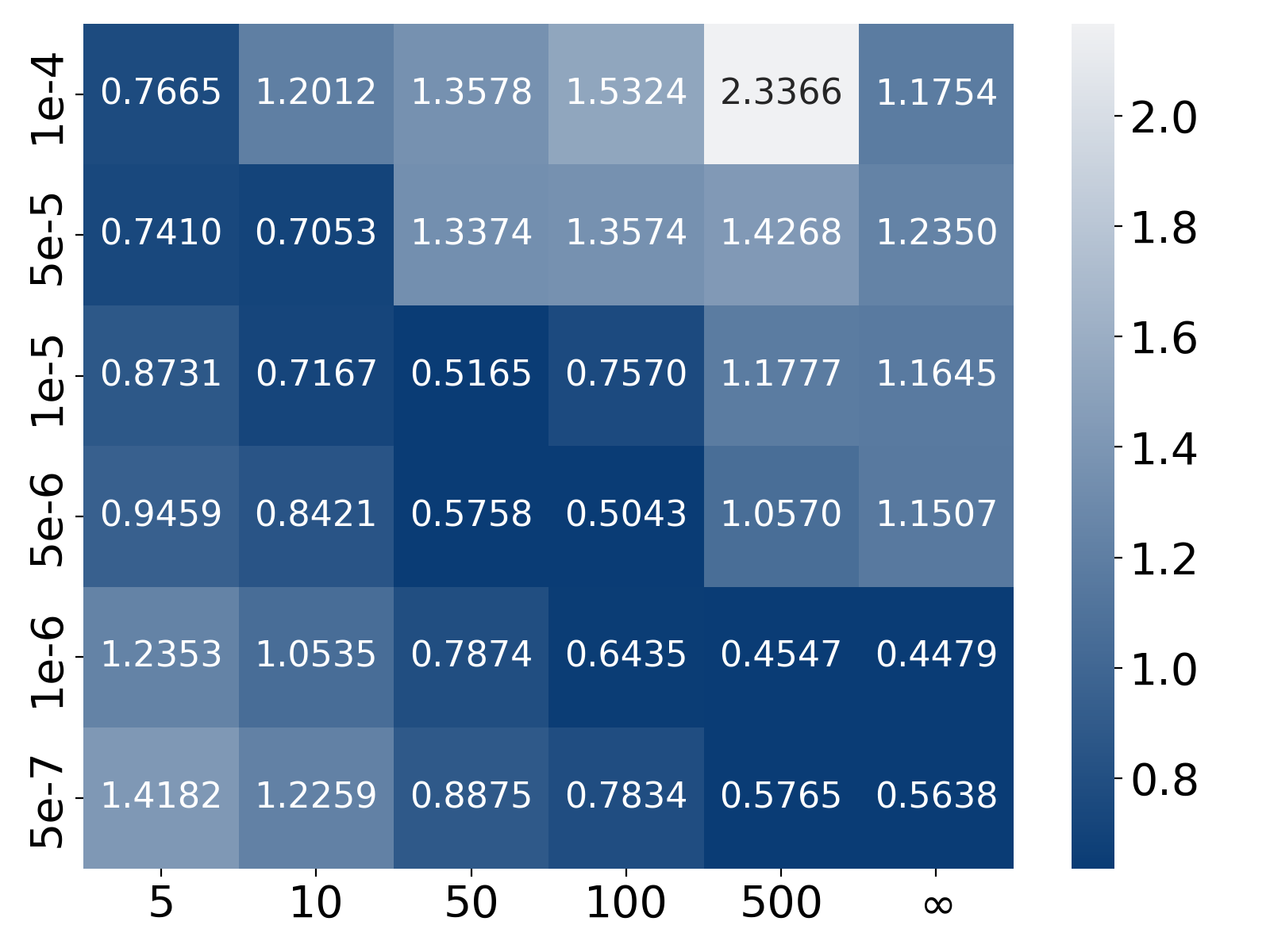}
    \put(13,-5){Clipping threshold $c$}
    \put(25,77){Test loss}
    \put(-8,40){\rotatebox[origin=c]{90}{Learning rate $\eta$}}
\end{overpic}

\end{tabular}

\vspace{-1mm}
\caption{Each cell shows the train accuracy (left), train loss (middle), and test loss (right) averaged over four models fine-tuned from RoBERTa \cite{liu2019roberta} on SNLI \cite{bowman2015large}, MNLI \cite{williams2018broad}, SST-2 \cite{socher2013recursive} and Trec \cite{li2002learning}, respectively. $c=\infty$ is the no clipping baseline.}
\label{fig:trainAccuracyLoss}
\end{figure}

We provide extended empirical results for the fine-tuning of RoBERTa (355M) using the clipped zeroth-order (ZO) estimator. We evaluate the performance across four standard downstream classification tasks: SST-2, MNLI, SNLI, and TREC. Across all datasets, we observe a consistent, structural interplay between the learning rate ($\eta$) and the clipping threshold ($c$).

\begin{figure}[h!]
\begin{center}
    \includegraphics[width=0.31\columnwidth]{images/heatmaps/SNLI_test_acc_heatmap.png}
    \put(-110,-7){Clipping threshold $c$}
    \put(-130,50){\rotatebox[origin=c]{90}{Learning rate $\eta$}}
    \hspace{0.2cm}
    \includegraphics[width=0.30\columnwidth]{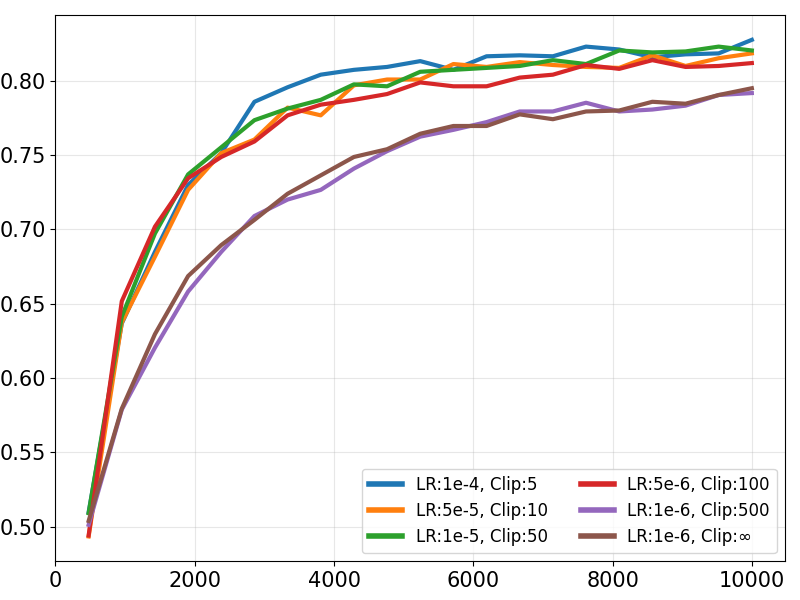}
    \put(-70,-7){Steps}
    \put(-130,50){\rotatebox[origin=c]{90}{Accuracy}}
    \hspace{0.2cm}
    \includegraphics[width=0.30\columnwidth]{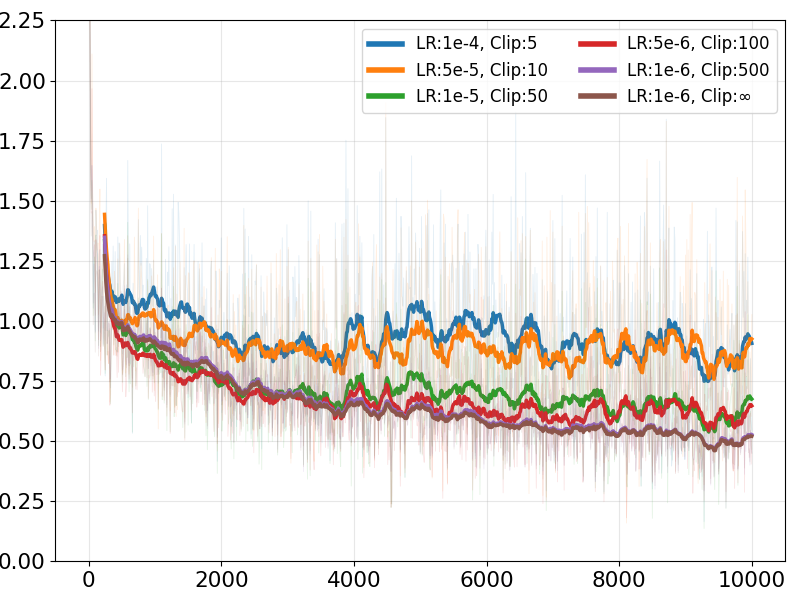}
    \put(-70,-7){Steps}
    \put(-128,45){\rotatebox[origin=c]{90}{Loss}}
\caption{\textbf{Left:} Test accuracy grid for SNLI on RoBERTA after sweeping through the learning rate $\eta$ and the clipping threshold $c$, we observe that the optimal learning rate for each clipping threshold lies on the curve $\eta\times c = 5\times 10^{-4}$. \textbf{Middle: } Learning curves for $\eta$ and $c$ pairs along the $\eta c$ line depicting the various rates of convergence. \textbf{Right: } SNLI train loss of the configurations along the line $\eta\,c=\text{constant}$ line. The higher loss and higher accuracy achieved along the diagonal ($\eta c=5\times10^{-4}$) with clipping ($c\leq 100$) compared to the no-clipping best loss implies that the gain of clipping is from better generalization and not better optimization.}
\label{fig:roberta_snli_grid}
\end{center}
\end{figure}

\begin{figure}[h!]
\begin{center}
    \includegraphics[width=0.41\columnwidth]{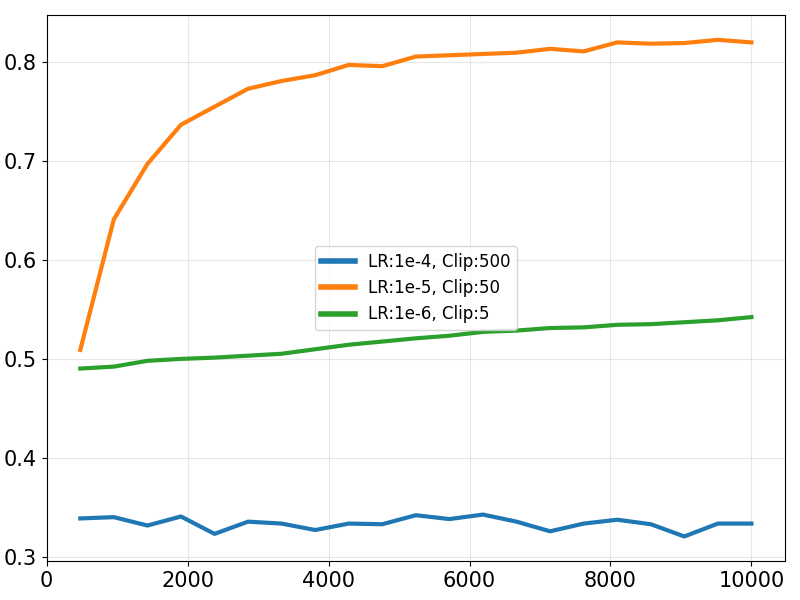}
    \put(-120,-7){Clipping threshold $c$}
    \put(-175,60){\rotatebox[origin=c]{90}{Accuracy}}
    \hspace{0.2cm}
    \includegraphics[width=0.42\columnwidth]{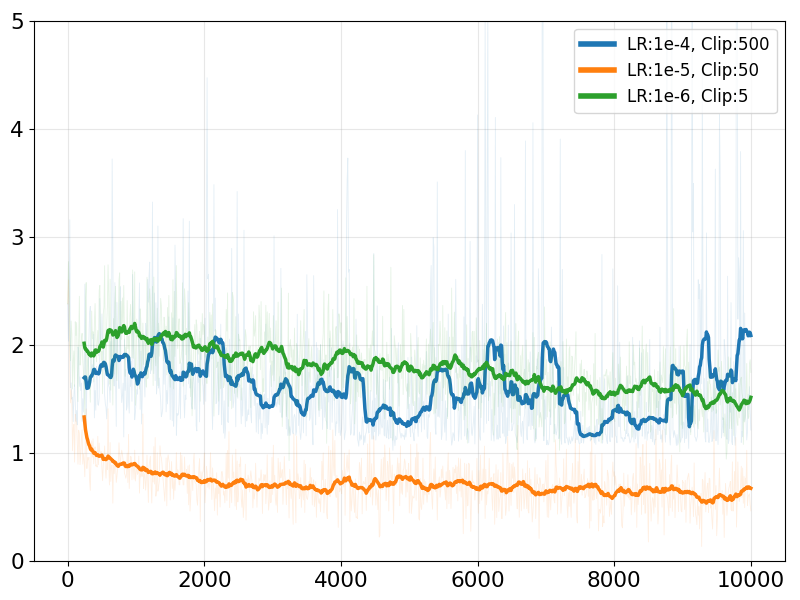}
    \put(-100,-7){Steps}
    \put(-175,60){\rotatebox[origin=c]{90}{Loss}}
    \hspace{0.2cm}
\caption{SNLI Learning curves (\textbf{left}) and loss curves (\textbf{right}) comparing configurations that deviate from the $\eta c$ diagonal ($\eta = 10^{-4}, c = 500$ and $\eta = 10^{-6}, c = 5$) against the configuration $\eta = 10^{-5}, c = 50$. Off-diagonal configurations show two failure modes: high $\eta$ with large $c$ leads to unstable training and random-guess accuracy (about 33\%), while low $\eta$ with small $c$ yields stable but overly slow convergence, plateauing at 49–57\% accuracy within 10,000 steps.}
\label{fig:roberta_snli_grid_off}
\end{center}
\end{figure}

\subsection{Observed Trends: SNLI}
\label{app:snli_trends}
The Stanford Natural Language Inference (SNLI) dataset \cite{bowman2015large} is a widely adopted benchmark formatted as a 3-class problem to predict entailment, contradiction, or neutral relationships between sentence pairs.  
Training with the clipped zeroth-order estimator reveals a strict, inversely proportional trend between the learning rate ($\eta$) and the clipping threshold ($c$). The test accuracy heatmaps demonstrate that as the learning rate increases, the clipping threshold must be correspondingly decreased to maintain stability. The peak performance zone, yielding the highest test accuracies ($\sim$83\%), lie along this optimal diagonal band (e.g., $\eta=5\times 10^{-5}$ paired with $c=10$, and $\eta=1\times 10^{-5}$ paired with $c=50$) satisfying the {\em clipping trade-off} at $\eta\,c=\text{constant}$.

Configurations deviating from this diagonal exhibit two distinct failure modes. In the upper-right region of the hyperparameter grid (high $\eta$, large $c$), the models fail to learn entirely. The test loss trajectories exhibit extreme, volatile spikes, and accuracy flatlines at the random guessing baseline of approximately 33\%. Conversely, configurations in the lower-left region (low $\eta$, small $c$) produce stable and smooth loss curves. However, their convergence is excessively slow; these models fail to reach peak accuracy within the allocated 10,000-step limit, stalling between 49\% and 57\% accuracy  (Figure~\ref{fig:roberta_snli_grid_off}).

Further examination of the continuous learning trajectories reveals that hyperparameter configurations split into distinct behavioral groups. Aggressive pairings—characterized by larger learning rates and smaller clipping thresholds (e.g., $\eta=10^{-4}, c=5$ and $\eta=5 \times 10^{-5}, c=10$)—exhibit rapid initial convergence, quickly reaching near-peak accuracy within the first 2,000 to 4,000 iterations. However, this speed is accompanied by a highly variant and elevated training loss profile. Instead, conservative configurations with lower learning rates and larger clipping (e.g., $\eta=10^{-6}, c=500$ or $c=\infty$) display a much shallower ascent in accuracy, but descend cleanly to lower, smoother training loss plateaus. 
The magnitude of the clipping threshold dictates the optimization trajectory, steering the models to converge into distinctly different solutions. 

\begin{figure}[]
\begin{center}
    \includegraphics[width=0.31\columnwidth]{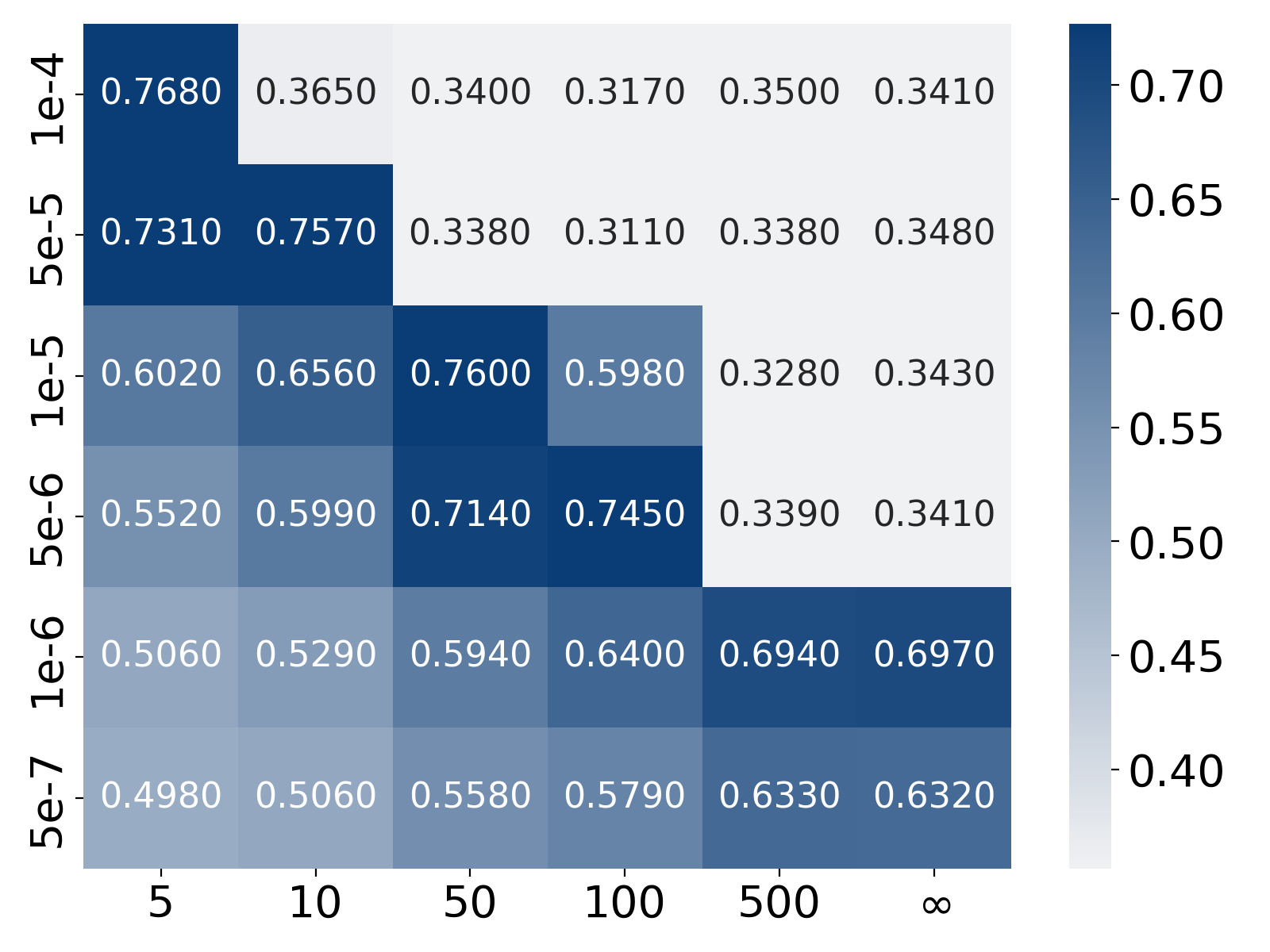}
    \put(-110,-7){Clipping threshold $c$}
    \put(-130,50){\rotatebox[origin=c]{90}{Learning rate $\eta$}}
    \hspace{0.2cm}
    \includegraphics[width=0.30\columnwidth]{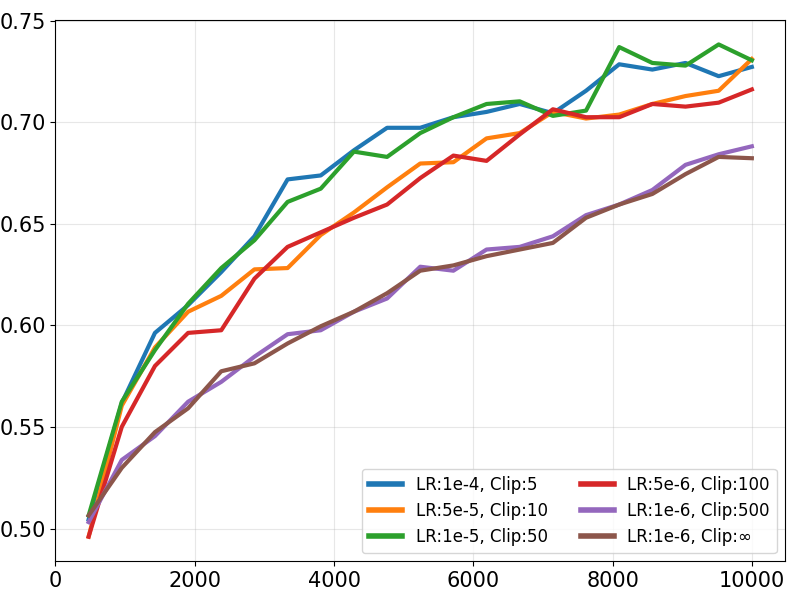}
    \put(-70,-7){Steps}
    \put(-130,50){\rotatebox[origin=c]{90}{Accuracy}}
    \hspace{0.2cm}
    \includegraphics[width=0.30\columnwidth]{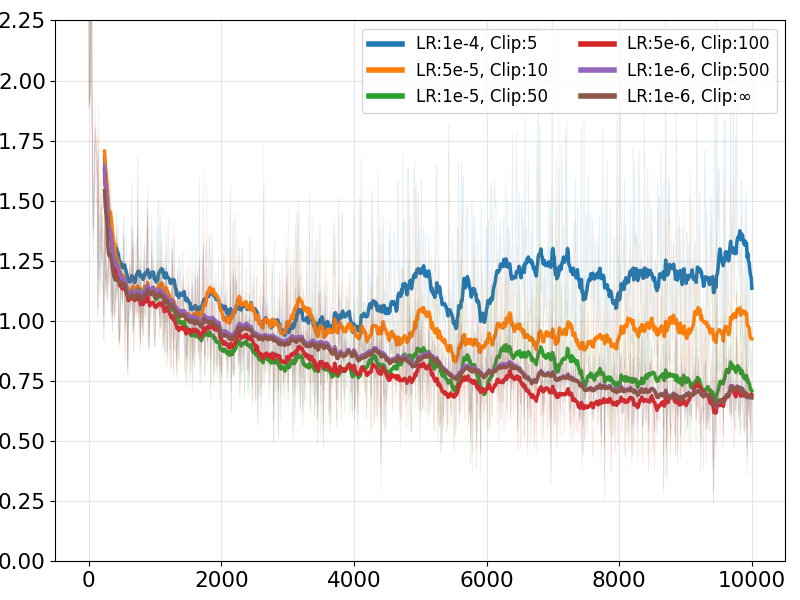}
    \put(-70,-7){Steps}
    \put(-128,45){\rotatebox[origin=c]{90}{Loss}}
    \hspace{0.2cm}
\caption{\textbf{Left:} MNLI test accuracy for RoBERTa across learning rates $\eta$ and clipping thresholds $c$. The best learning rate at each threshold follows $\eta c = 5\times10^{-4}$. \textbf{Middle:} Learning curves for $(\eta, c)$ pairs along this line, showing different convergence rates. \textbf{Right:} MNLI training loss for configurations satisfying $\eta c=\text{constant}$. As with SNLI (Appendix~\ref{app:snli_trends}), clipped runs ($c\leq100$) achieve higher loss but better accuracy than the best unclipped run, suggesting clipping improves generalization rather than optimization.}
\label{fig:roberta_mnli_grid}
\end{center}
\end{figure}

\begin{figure}[]
\begin{center}
    \includegraphics[width=0.41\columnwidth]{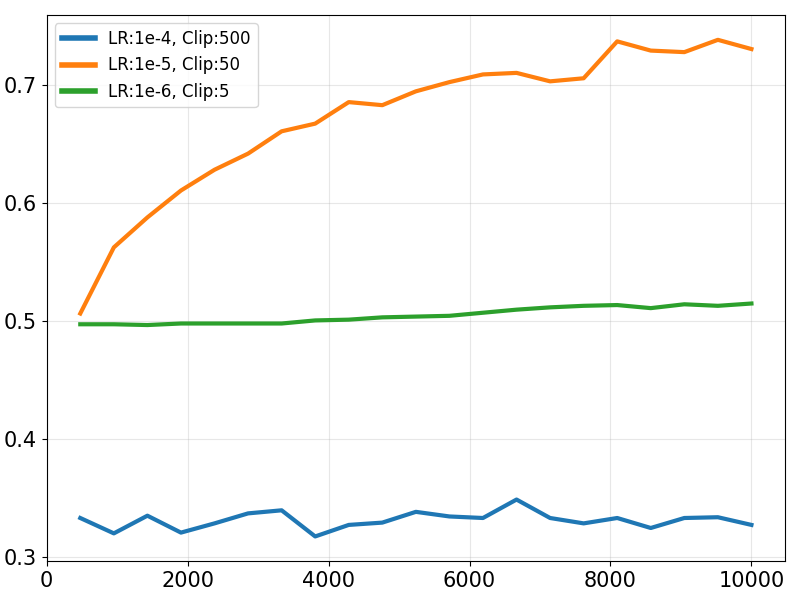}
    \put(-120,-7){Clipping threshold $c$}
    \put(-175,60){\rotatebox[origin=c]{90}{Accuracy}}
    \hspace{0.2cm}
    \includegraphics[width=0.42\columnwidth]{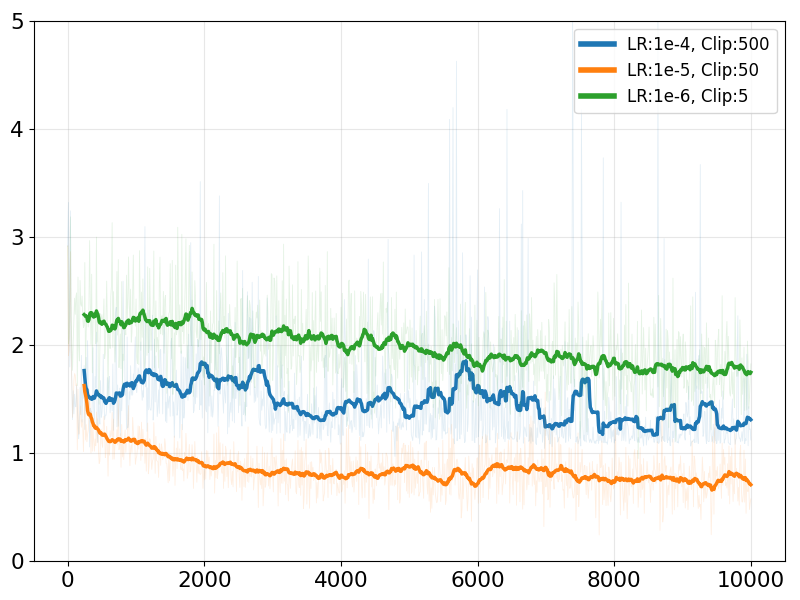}
    \put(-100,-7){Steps}
    \put(-175,60){\rotatebox[origin=c]{90}{Loss}}
    \hspace{0.2cm}
\caption{MNLI learning curves (\textbf{left}) and loss curves (\textbf{right}) comparing configurations that deviate from the $\eta c$ diagonal ($\eta = 10^{-4}, c = 500$ and $\eta = 10^{-6}, c = 5$) against the configuration $\eta = 10^{-5}, c = 50$. Similar to SNLI, the off-diagonal settings perform worse: high $\eta$ with loose $c$ causes unstable training and random-guess accuracy, while low $\eta$ with strict $c$ converges slowly and plateaus below the baseline.}
\label{fig:roberta_mnli_grid_off}
\end{center}
\end{figure}

\subsection{Observed Trends: MNLI}
\label{app:mnli_trends}

Like SNLI, the Multi-Genre Natural Language Inference (MNLI) corpus \cite{williams2018broad} is a widespread 3-class sentence understanding benchmark requiring models to predict entailment, contradiction, or neutral relationships across diverse text genres.

Empirical results on the MNLI dataset perfectly mirror the structural dynamics observed in SNLI, confirming that the clipping trade-off is a robust and consistent phenomenon across different datasets. The test accuracy heatmap reveals that the highest performance (approximately 76\%) also lies strictly along the established diagonal cells, maintaining an inversely proportional relationship between the learning rate ($\eta$) and the clipping threshold ($c$).

Deviating from this optimal band yields two distinct failure modes. In the high $\eta$ and large $c$ regime, failing to sufficiently clip the large learning rates results in the accuracy remaining trapped at the 33\% random-guessing baseline. On the opposite end of the spectrum, employing a low $\eta$ with a small $c$ produces smooth, linear accuracy climbs. However, the step sizes are too small, preventing the model from traversing the loss landscape quickly enough to reach a similar performance of the configurations along the diagonal band within the 10,000-step computational limit.

\begin{figure}[]
\begin{center}
    \includegraphics[width=0.31\columnwidth]{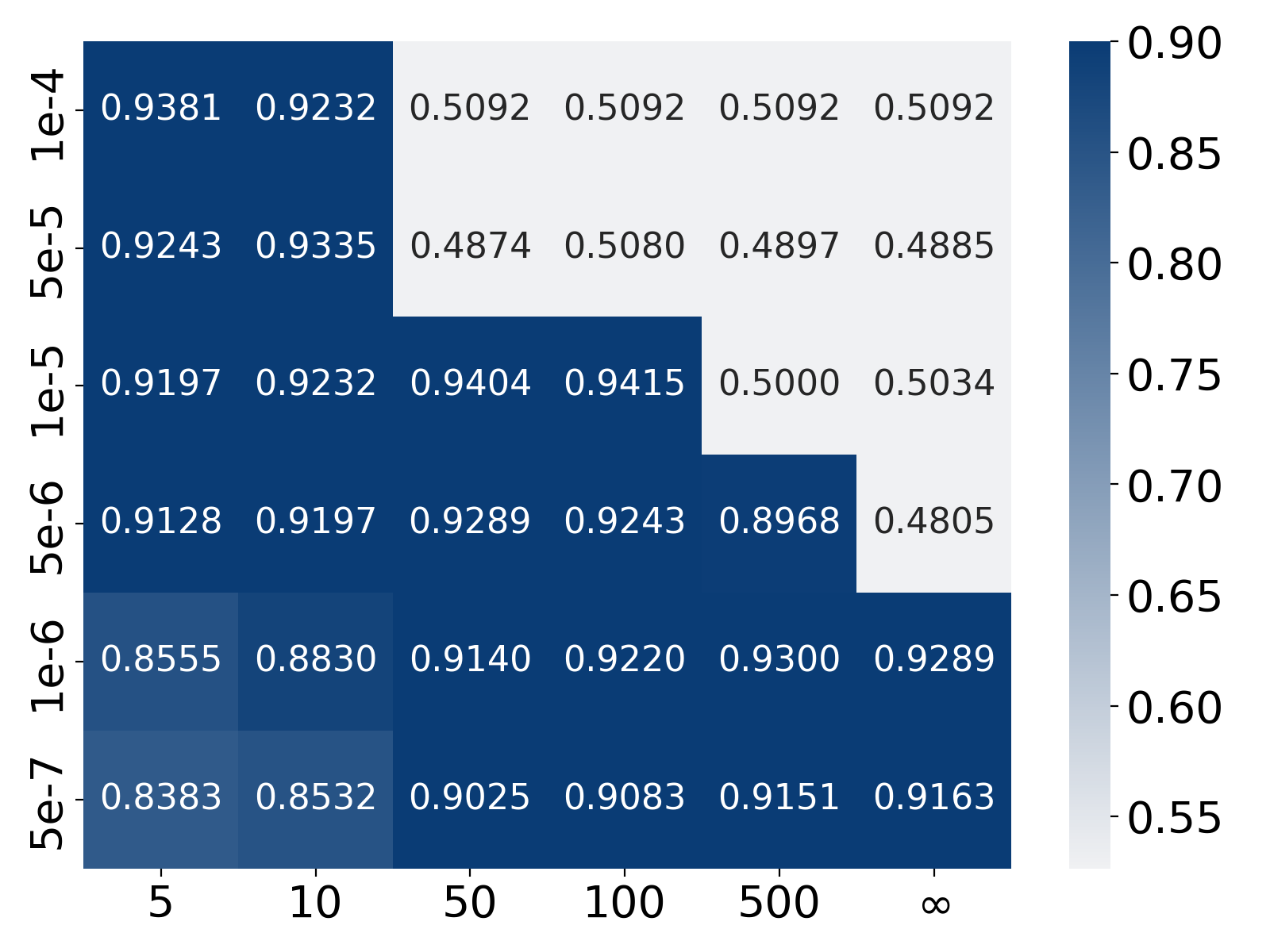}
    \put(-110,-7){Clipping threshold $c$}
    \put(-130,50){\rotatebox[origin=c]{90}{Learning rate $\eta$}}
    \hspace{0.2cm}
    \includegraphics[width=0.30\columnwidth]{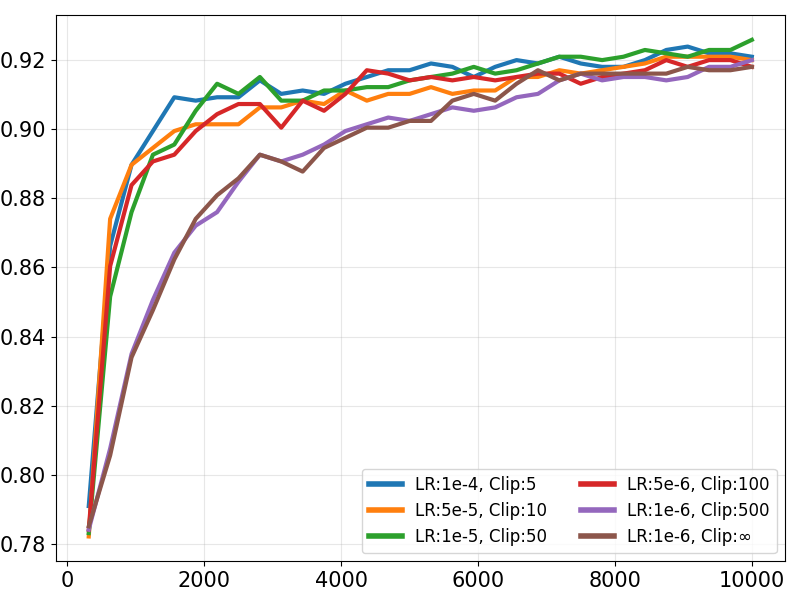}
     \put(-70,-7){Steps}
    \put(-130,50){\rotatebox[origin=c]{90}{Accuracy}}
    \hspace{0.2cm}
    \includegraphics[width=0.30\columnwidth]{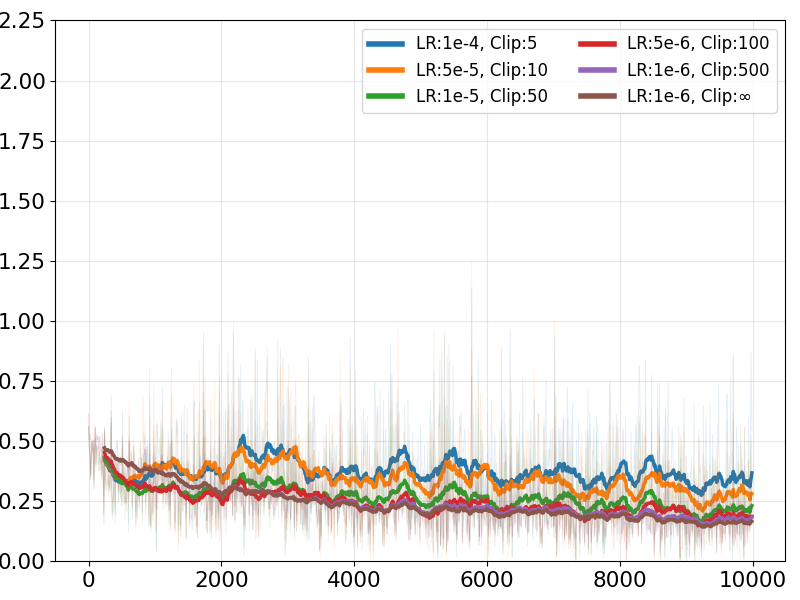}
    \put(-70,-7){Steps}
    \put(-128,45){\rotatebox[origin=c]{90}{Loss}}
    \hspace{0.2cm}
\caption{The clipping trade-off trend is less prominent for Roberta fine-tuned on SST-2.}
\label{fig:roberta_sst2_grid}
\end{center}
\end{figure}

\begin{figure}[]
\begin{center}
    \includegraphics[width=0.41\columnwidth]{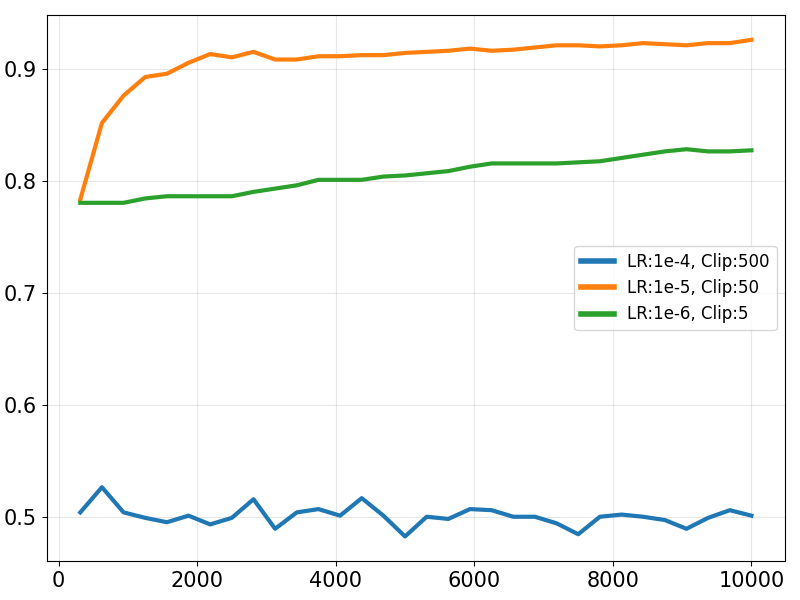}
    \put(-120,-7){Clipping threshold $c$}
    \put(-175,60){\rotatebox[origin=c]{90}{Accuracy}}
    \hspace{0.2cm}
    \includegraphics[width=0.42\columnwidth]{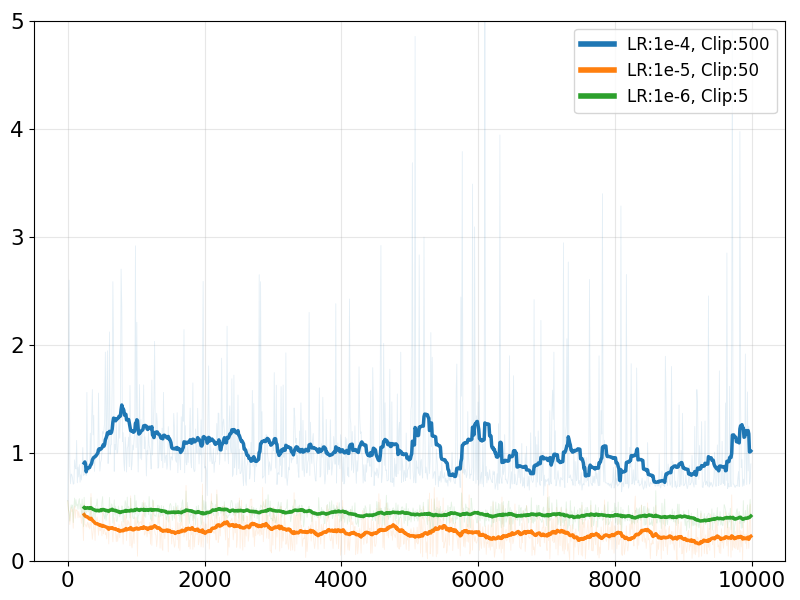}
    \put(-100,-7){Steps}
    \put(-175,60){\rotatebox[origin=c]{90}{Loss}}
    \hspace{0.2cm}
\caption{SST-2 learning (\textbf{left}) and loss (\textbf{right}) curves comparing off-diagonal configurations ($\eta = 10^{-4}, c = 500$ and $\eta = 10^{-6}, c = 5$) with $\eta = 10^{-5}, c = 50$. Similar to SNLI and MNLI, the off-diagonal settings perform worse: high $\eta$ with loose $c$ leads to unstable training and chance-level accuracy, while low $\eta$ with strict $c$ converges slowly and plateaus below the baseline.}
\label{fig:roberta_sst2_grid_off}
\end{center}
\vskip -0.2in
\end{figure}

\subsection{Observed Trends: SST-2}
\label{app:sst2_trends}

The Stanford Sentiment Treebank (SST-2) \cite{socher2013recursive} is a standard binary classification task designed to predict the sentiment of movie reviews. Training with the clipped zeroth-order estimator also shows the trade-off. The optimal performance is concentrated along the diagonal band, maintaining the inversely proportional relationship between the learning rate ($\eta$) and the clipping threshold ($c$). Configurations parameterized along this optimal band (e.g., $\eta=1\times 10^{-5}$ paired with $c=50$) achieve the highest convergence, consistently exceeding 93\% test accuracy. 
 Outside this optimal diagonal exposes severe optimization bottlenecks similar to SNLI and MNLI. 
\begin{figure}[]
\centering
    \includegraphics[width=0.31\columnwidth]{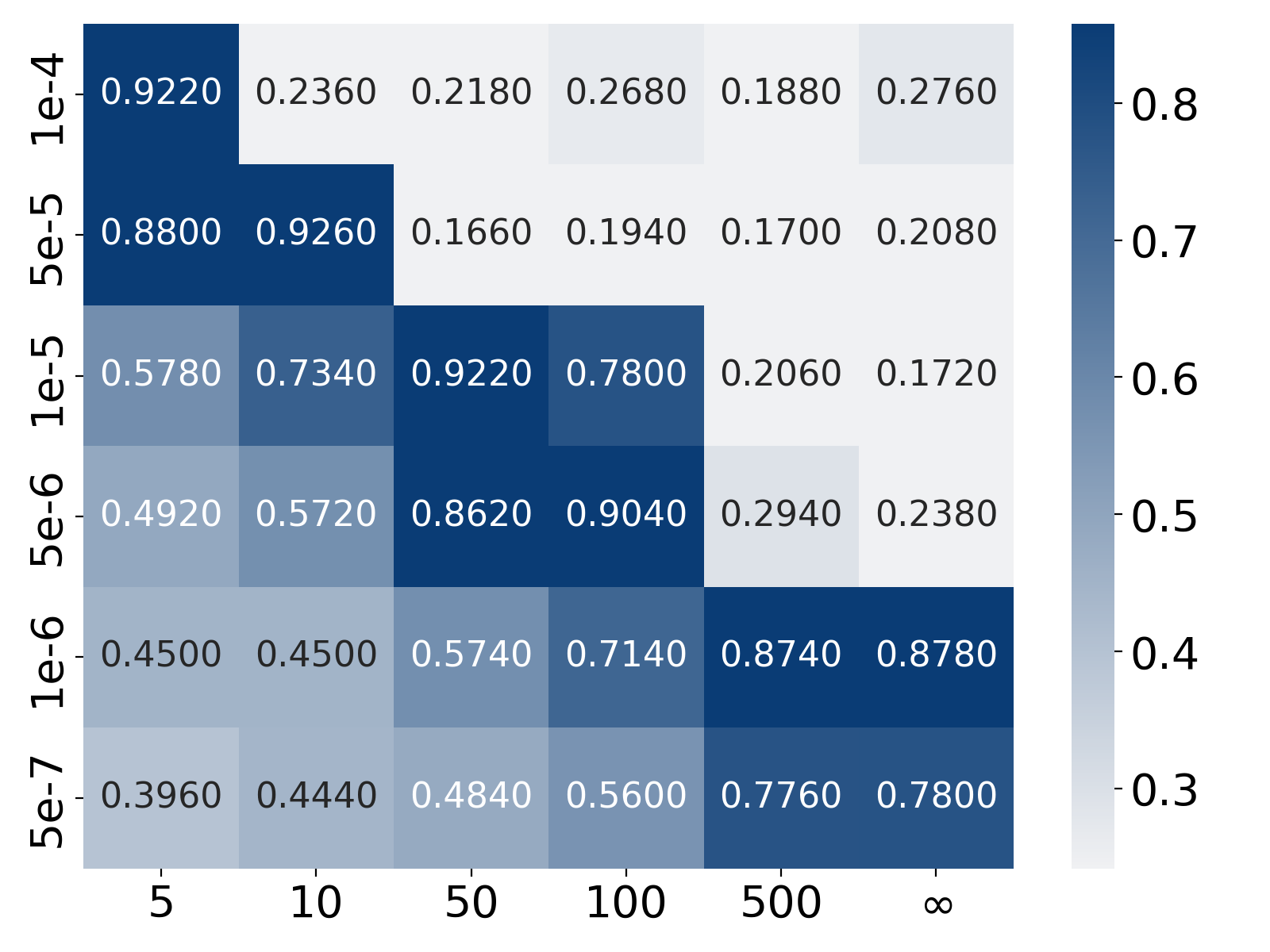}
    \put(-110,-7){Clipping threshold $c$}
    \put(-130,50){\rotatebox[origin=c]{90}{Learning rate $\eta$}}
    \hspace{0.2cm}
    \includegraphics[width=0.30\columnwidth]{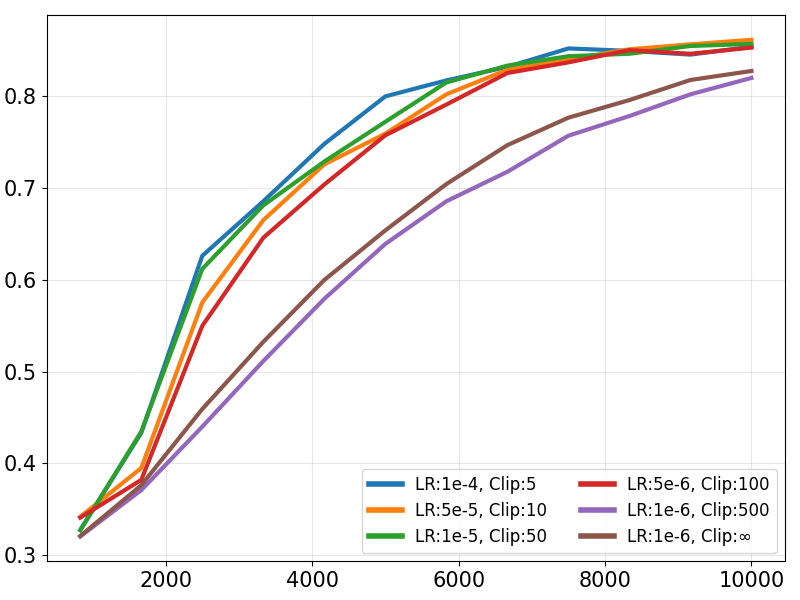}
    \put(-70,-7){Steps}
    \put(-130,50){\rotatebox[origin=c]{90}{Accuracy}}
    \hspace{0.2cm}
    \includegraphics[width=0.30\columnwidth]{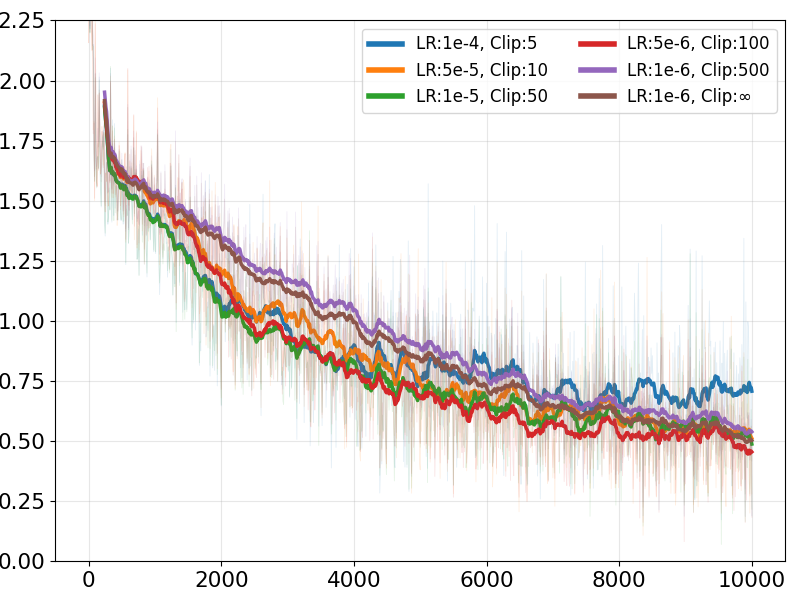}
    \put(-70,-7){Steps}
    \put(-128,45){\rotatebox[origin=c]{90}{Loss}}
    \hspace{0.2cm}
\caption{Clipping trade-off trends for RoBERTa fine-tuned on TREC, showing the same {\em clipping trade-off} behavior observed in \ref{app:snli_trends}, \ref{app:mnli_trends}, and \ref{app:sst2_trends}.}
\label{fig:roberta_trec_grid}
\end{figure}

\begin{figure}[]
\centering
    \includegraphics[width=0.41\columnwidth]{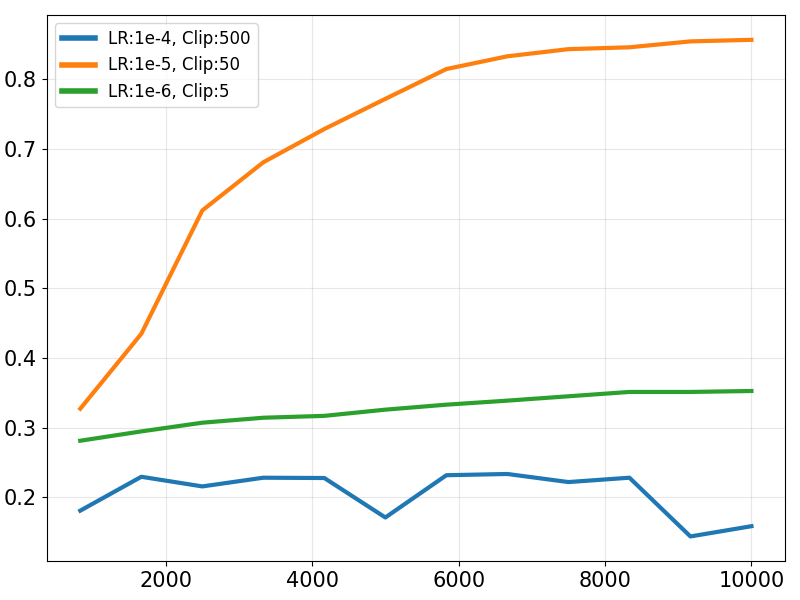}
    \put(-120,-7){Clipping threshold $c$}
    \put(-175,60){\rotatebox[origin=c]{90}{Accuracy}}
    \hspace{0.2cm}
    \includegraphics[width=0.42\columnwidth]{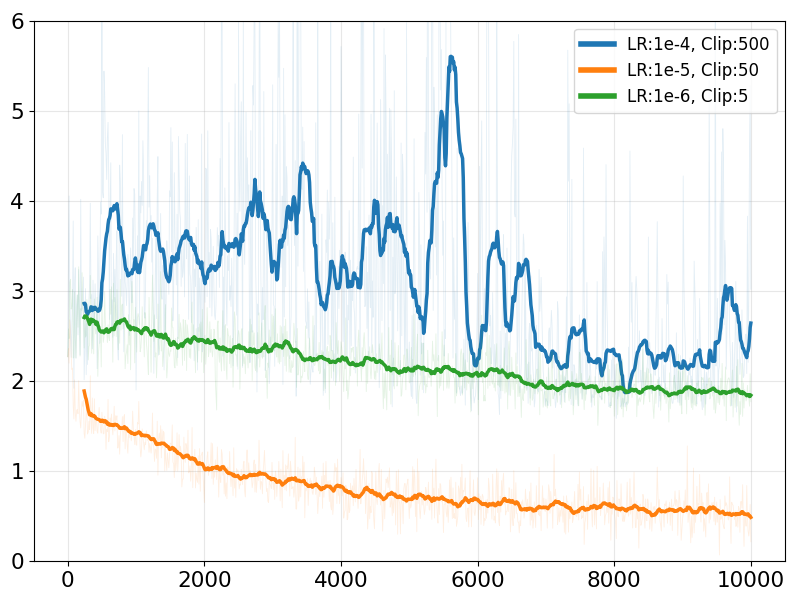}
    \put(-100,-7){Steps}
    \put(-175,60){\rotatebox[origin=c]{90}{Loss}}
    \hspace{0.2cm}

\caption{TREC learning curves (\textbf{left}) and loss curves (\textbf{right}) comparing configurations that deviate from the $\eta c$ diagonal ($\eta = 10^{-4}, c = 500$ and $\eta = 10^{-6}, c = 5$) against the configuration $\eta = 10^{-5}, c = 50$. Similar to SNLI, SST-2, and MNLI, the off-diagonal settings perform worse: high $\eta$ with loose $c$ causes unstable training and random-guess accuracy, while low $\eta$ with strict $c$ converges slowly and plateaus below the baseline.}
\label{fig:roberta_trec_grid_off}
\end{figure}

\subsection{Observed Trends: TREC}
\label{app:trec_trends}

The TREC dataset \cite{li2002learning} is a question classification benchmark requiring models to categorize open-domain, fact-based questions into one of six distinct semantic categories. Training with the clipped zeroth-order estimator on this 6-class problem confirms the  clipping trade-off. 
Configurations parameterized away from this stability boundary exhibit severe performance degradation \ref{fig:roberta_trec_grid_off}. 

\section{Additional experiments: OPT} 
\label{app:opt}

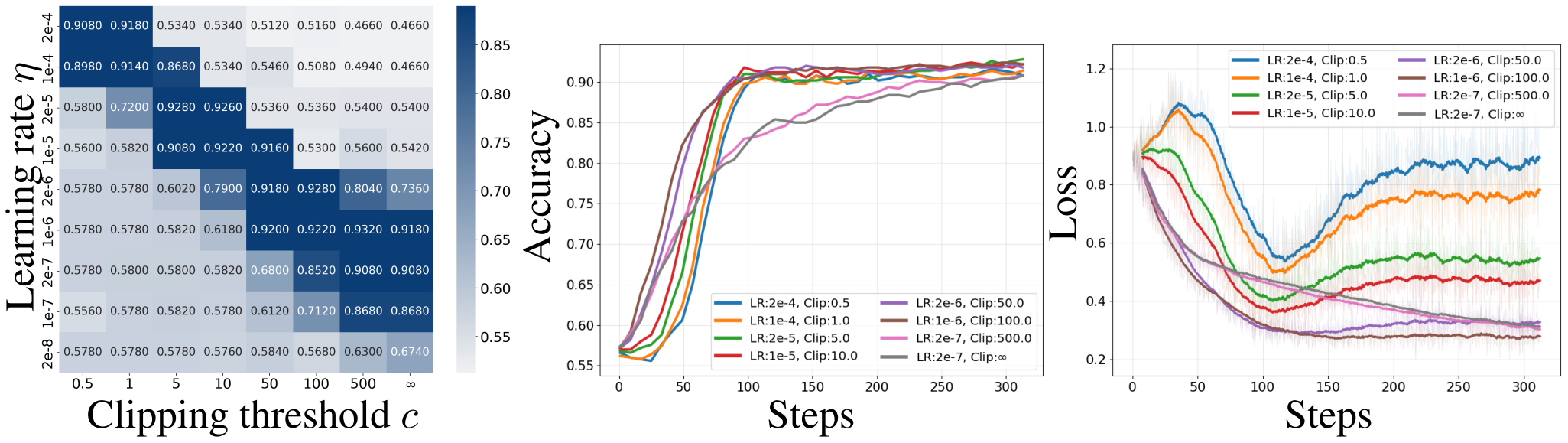
\begin{figure}[h]
\begin{center}
    \includegraphics[width=0.31\columnwidth]{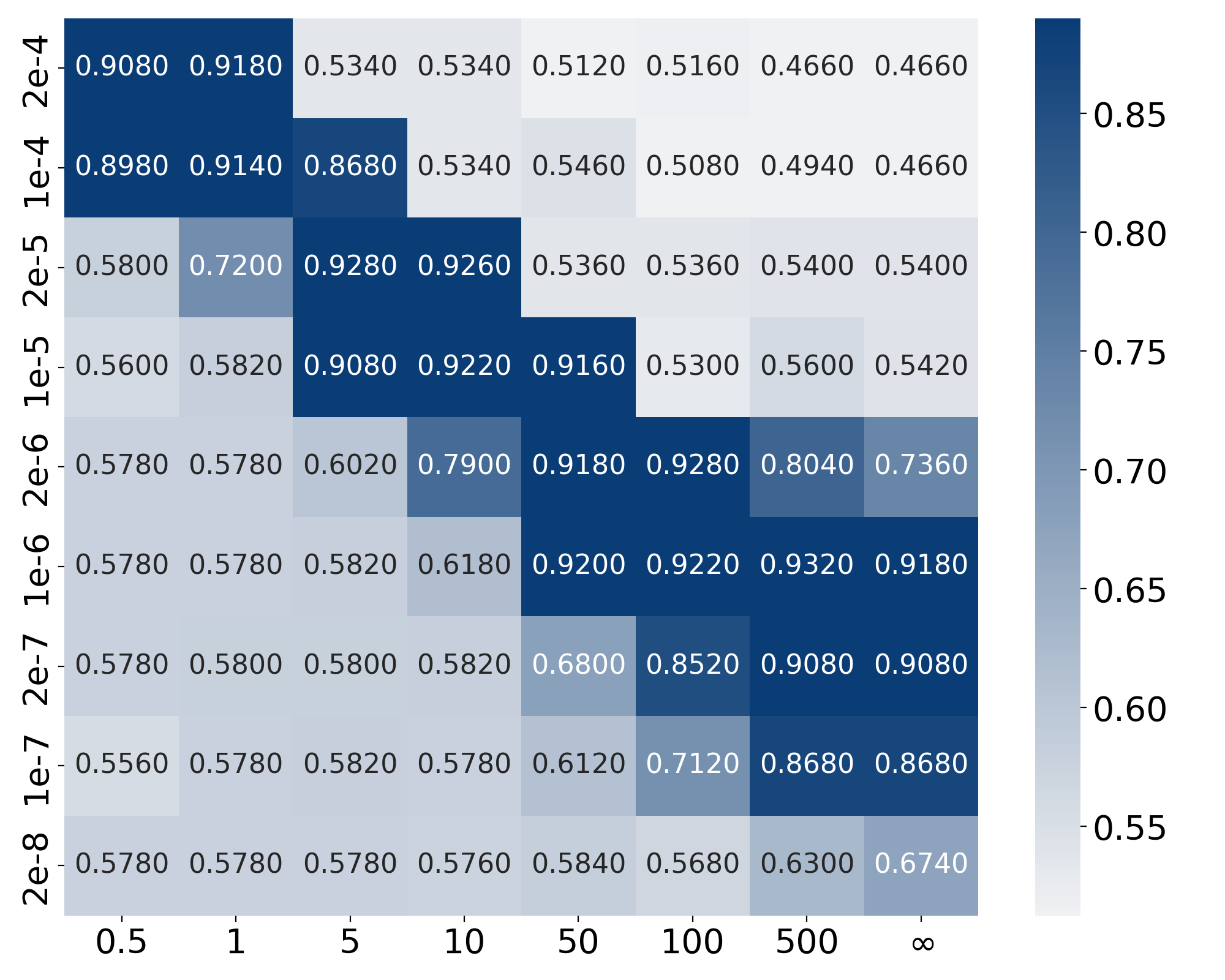}
    \put(-110,-7){Clipping threshold $c$}
    \put(-130,50){\rotatebox[origin=c]{90}{Learning rate $\eta$}}
    \hspace{0.2cm}
    \includegraphics[width=0.30\columnwidth]{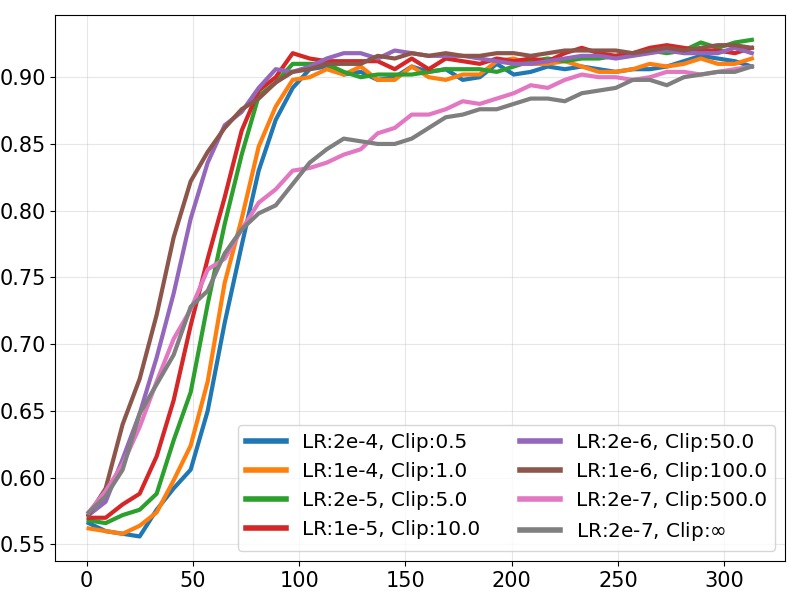}
    \put(-70,-7){Steps}
    \put(-130,50){\rotatebox[origin=c]{90}{Accuracy}}
    \hspace{0.2cm}
    \includegraphics[width=0.30\columnwidth]{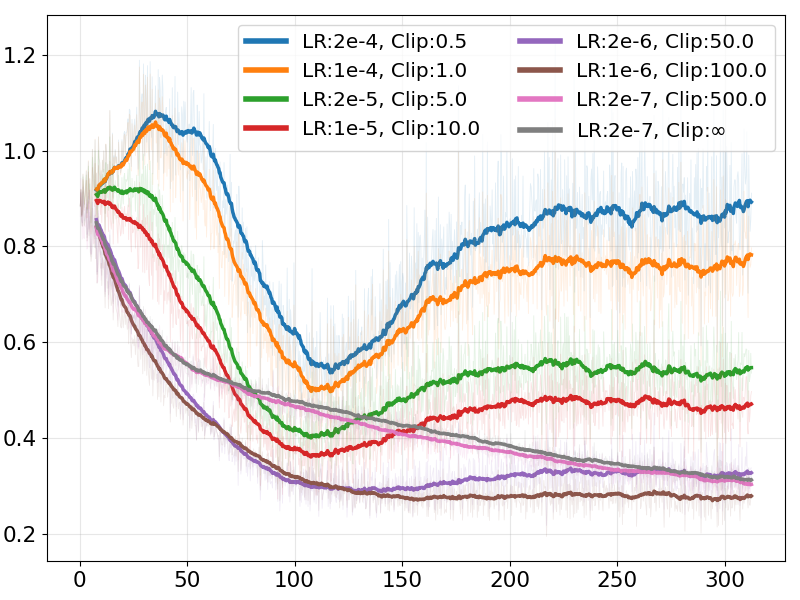}
    \put(-70,-7){Steps}
    \put(-128,45){\rotatebox[origin=c]{90}{Loss}}
    \hspace{0.2cm}

\end{center}
\caption{Experiments on private fine-tuning OPT (1.3B) for SST-2 with \ref{eq:clip}. \textbf{Left:} Test accuracy \% when varying the stepsize and clipping threshold together. \textbf{Middle:} Accuracy plots depicting the convergence of the diagonal band configurations. \textbf{Right:} (Smoothed) training
loss curves of the diagonal band configurations. Consistent with previous experiments, we observe a similar behavior with clipping and zeroth order methods converging with different behaviors when changing models from Roberta to OPT.}
\label{fig:sst_opt}
\end{figure}

\subsection{Observed Trends: SST-2}
\label{app:sst2_trends_opt}

To verify that the clipping trade-off generalizes to beyond Roberta for more advanced architectures, we extend our evaluation to the OPT \cite{zhang2022opt} model fine-tuned on the SST-2 binary classification benchmark \cite{socher2013recursive}. 
The test accuracy heatmap demonstrates that the OPT architecture strongly preserves the diagonal scaling trend previously observed in encoder-only models. Peak performance forms a distinct structural band where higher learning rates ($\eta$) strictly require tighter clipping thresholds ($c$). Accuracies exceeding 90\%, and reaching up to 92.8\%, are clustered entirely along this optimal diagonal (e.g., $\eta=2\times 10^{-5}$ paired with $c=5.0$, and $\eta=1\times 10^{-6}$ paired with $c=100.0$). Deviating from this boundary yields familiar failure modes. In the upper-right regime (e.g., $\eta=1\times 10^{-4}$ with $c \geq 10.0$), the unconstrained variance causes complete model failure, with accuracy catastrophically collapsing to the binary random-guessing baseline of approximately 49\% to 54\%. In the overly constrained lower-left region (e.g., $\eta \leq 1\times 10^{-6}$ with $c \leq 1.0$), the model avoids this collapse but stagnates, plateauing at a sub-optimal 57\% to 58\% accuracy within the given training budget.

An analysis of the continuous learning trajectories reveals unified convergence for configurations situated on the optimal diagonal. Despite utilizing drastically different hyperparameter pairings, all sampled configurations successfully navigate to a similar peak accuracy of 91\% to 93\%. However, the speed of this traversal varies significantly. Configurations parameterized with a higher base learning rate (e.g., $\eta=2\times 10^{-5}, c=5.0$) rapidly optimize the objective, achieving 90\% accuracy in fewer than 50 epochs. Conversely, conservative setups utilizing lower learning rates and looser clipping thresholds (e.g., $\eta=1\times 10^{-6}, c=100.0$) exhibit a much shallower ascent, requiring 150 to 200 epochs to cross the same performance threshold.

Examining the corresponding training loss profiles exposes a direct correlation between the learning rate and the zeroth-order estimator's variance. High-$\eta$ configurations inherently display highly volatile, thick loss bands throughout the training run, reflecting the larger stochastic perturbations. In contrast, configurations with low learning rates (e.g., $\eta=2\times 10^{-7}, c=500.0$) produce tightly clustered, smooth loss curves. 

\subsection{Expanded Clipping Window: RoBERTa-SNLI} 
\label{app:roberta-expand}

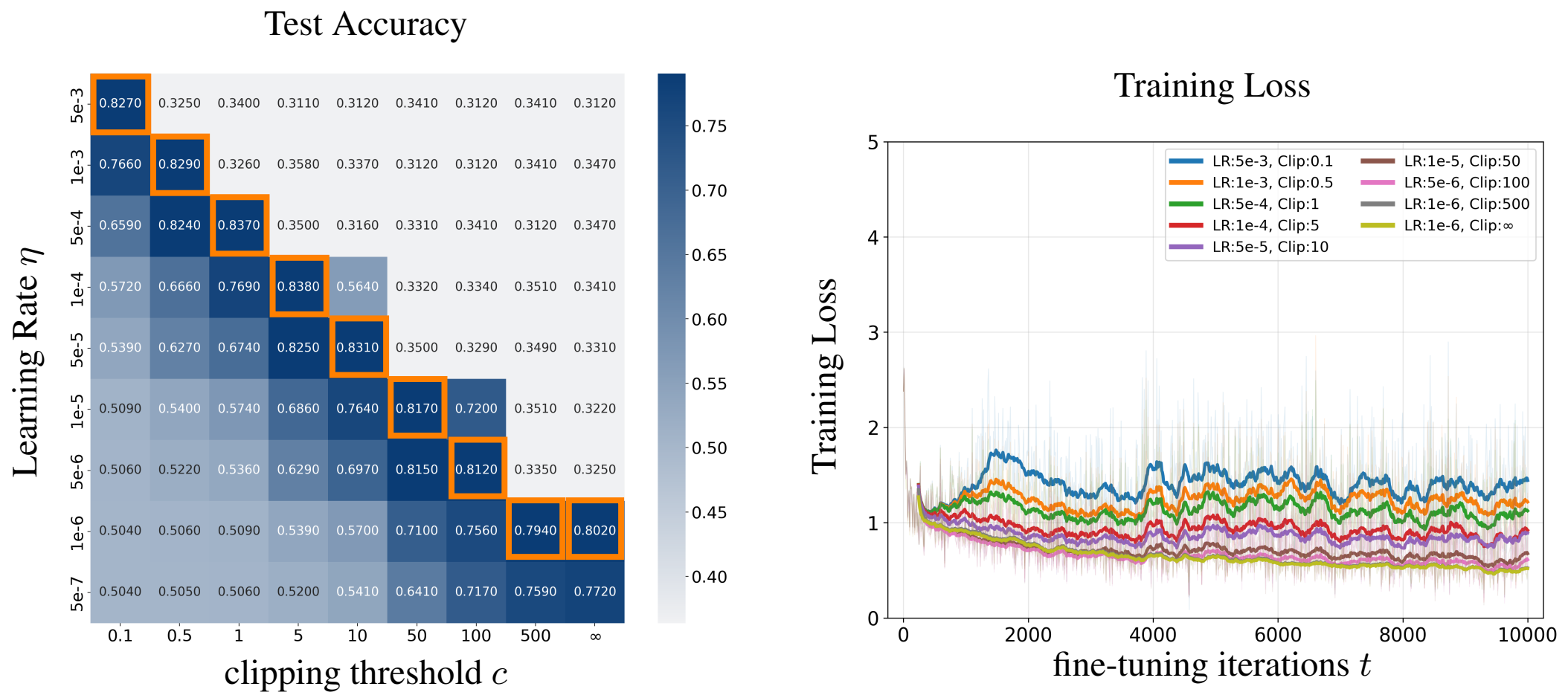
\begin{figure}[h!]
\centering

\makebox[\columnwidth][c]{%
    \begin{tikzpicture}[baseline]
        \node[anchor=south west, inner sep=0, outer sep=0]
            (image) at (0,0)
        {\includegraphics[width=0.45\columnwidth]
        {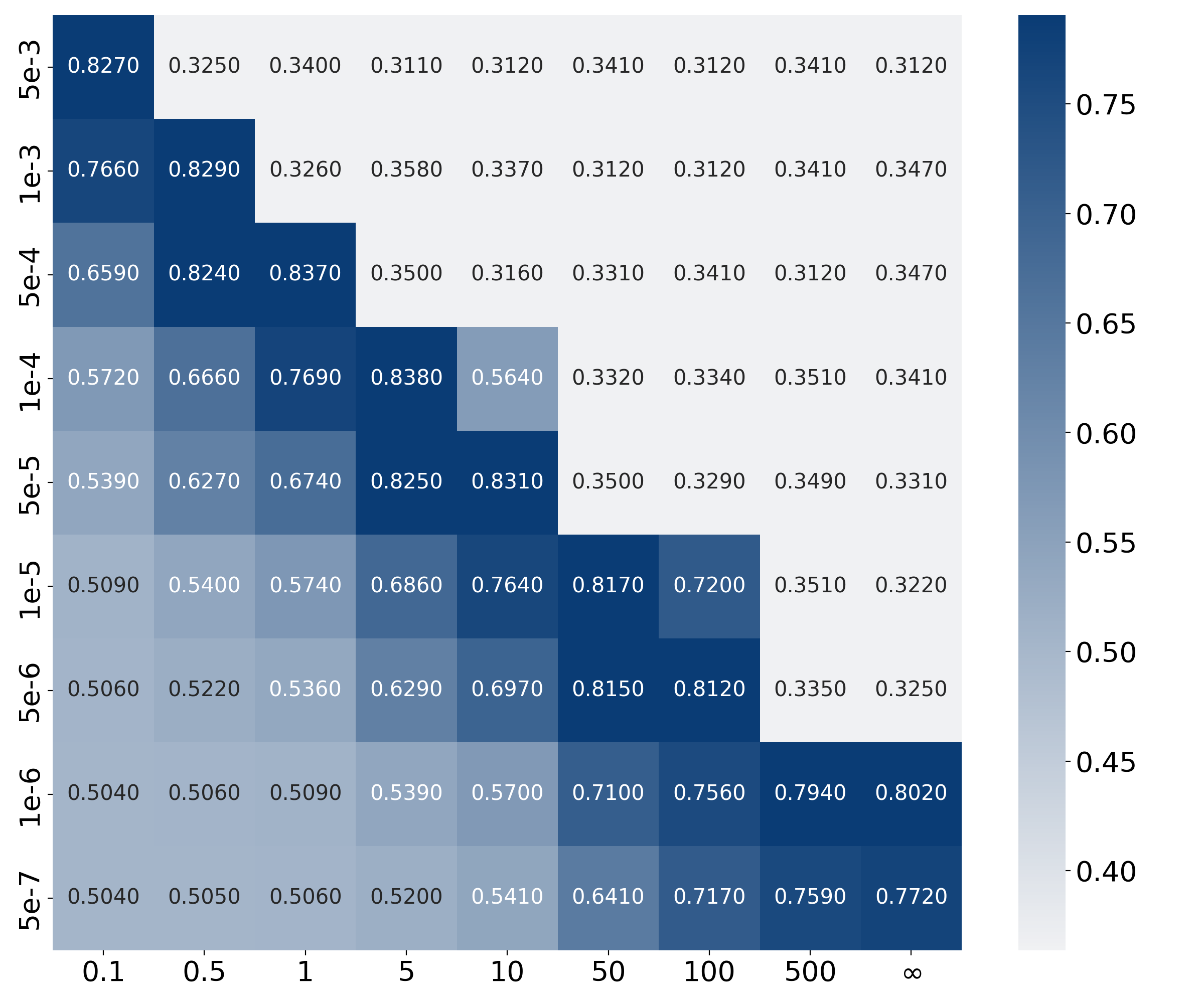}};

        \begin{scope}[x={(image.south east)}, y={(image.north west)}]

            \draw[orange, line width=1.5pt]
                (0.0475,0.885) rectangle (0.125,0.98);

            \draw[orange, line width=1.5pt]
                (0.132,0.782) rectangle (0.208,0.878);

            \draw[orange, line width=1.5pt]
                (0.2165,0.679) rectangle (0.2925,0.775);

            \draw[orange, line width=1.5pt]
                (0.301,0.576) rectangle (0.377,0.672);

            \draw[orange, line width=1.5pt]
                (0.3855,0.473) rectangle (0.4615,0.569);

            \draw[orange, line width=1.5pt]
                (0.468,0.370) rectangle (0.544,0.466);

            \draw[orange, line width=1.5pt]
                (0.554,0.267) rectangle (0.629,0.363);

            \draw[orange, line width=1.5pt]
                (0.635,0.164) rectangle (0.711,0.260);

            \draw[orange, line width=1.5pt]
                (0.720,0.164) rectangle (0.794,0.260);

        \end{scope}

        \node at ([xshift=-12pt,yshift=0pt]image.north) [above=2pt]
            {\small Test Accuracy};

        \node at ([xshift=-12pt,yshift=10pt]image.south)[below=8pt]
            {\small clipping threshold $c$};

        \node at ([xshift=-0pt,yshift=32pt]image.west)[left=8pt, rotate=90]
            {\small Learning Rate $\eta$};

    \end{tikzpicture}%
    \hspace{0.02\columnwidth}%
    \begin{tikzpicture}[baseline]
        \node[anchor=south west, inner sep=0, outer sep=0]
            (loss) at (0,0)
        {\includegraphics[width=0.45\columnwidth]
        {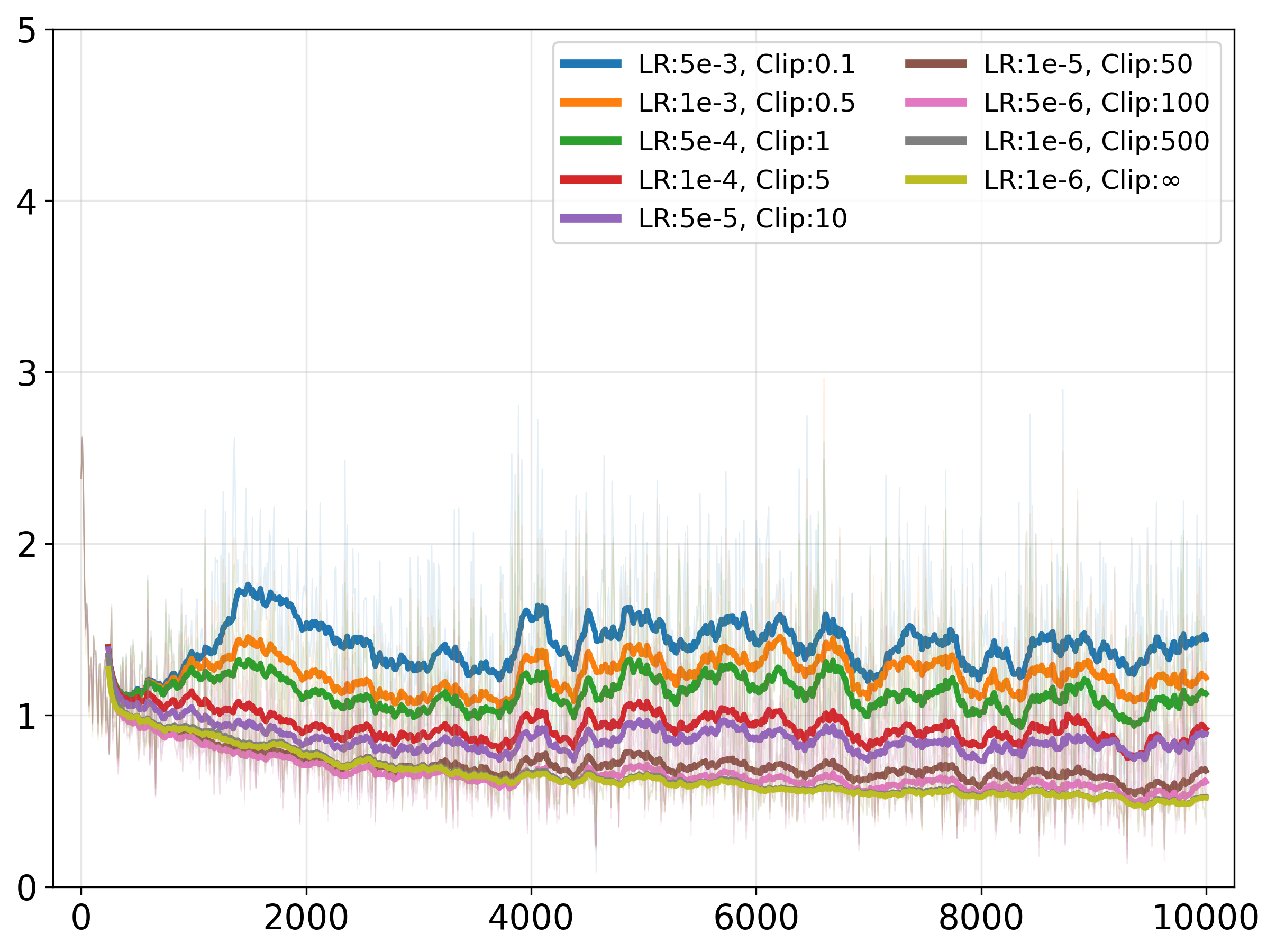}};

        \node at ([xshift=-0pt,yshift=0pt]loss.north) [above=2pt]
            {\small Training Loss};

        \node at ([xshift=-0pt,yshift=12pt]loss.south) [below=8pt]
            {\small fine-tuning iterations $t$};

        \node at ([xshift=-0pt,yshift=32pt]loss.west)[left=8pt, rotate=90]
            {\small Training Loss};

    \end{tikzpicture}%
}

\caption{\sg{Extended SNLI evaluation including smaller clipping thresholds $c \in \{0.1, 0.5, 1\}$ along the $\eta c = 5\times 10^{-4}$ trade-off line (highlighted in orange). (\textbf{left}) Test accuracy heatmap across learning rates $\eta \in [5\times 10^{-7}, 5\times 10^{-3}]$ and clipping thresholds $c \in [0.1, 500]\cup\{\infty\}$, showing that both the $\eta c = 5\times 10^{-4}$ scaling and high test accuracy ($82.7\%$--$83.8\%$ vs.\ $80.2\%$ at $c=\infty$) persist down to $c=0.1$. (\textbf{right}) Training loss trajectories over $10{,}000$ steps for the nine orange-highlighted diagonal configurations, showing that tighter clipping ($c \in \{0.1, 0.5, 1\}$) systematically plateaus at higher training loss while maintaining high test accuracy.}}
\label{fig:snli_small_c}

\end{figure}

\begin{figure}[h!]
\begin{center}
    \includegraphics[width=0.48\columnwidth]{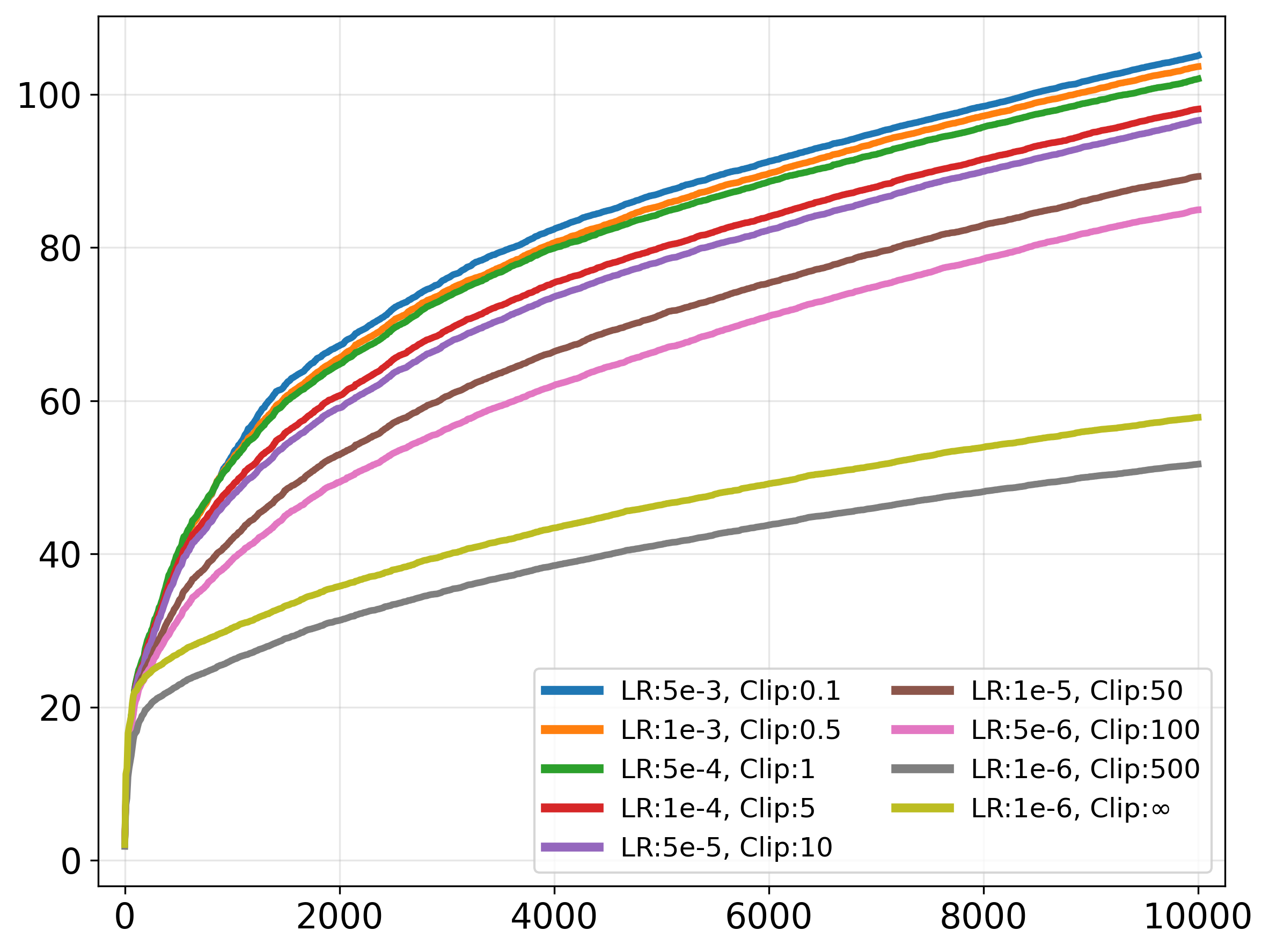}
    \put(-135,-8){\small fine-tuning iterations $t$}
    \put(-200,70){\rotatebox[origin=c]{90}{\small Distance $\|w_t-w_0\|$}}
    \hspace{0.2cm}
    \includegraphics[width=0.48\columnwidth]{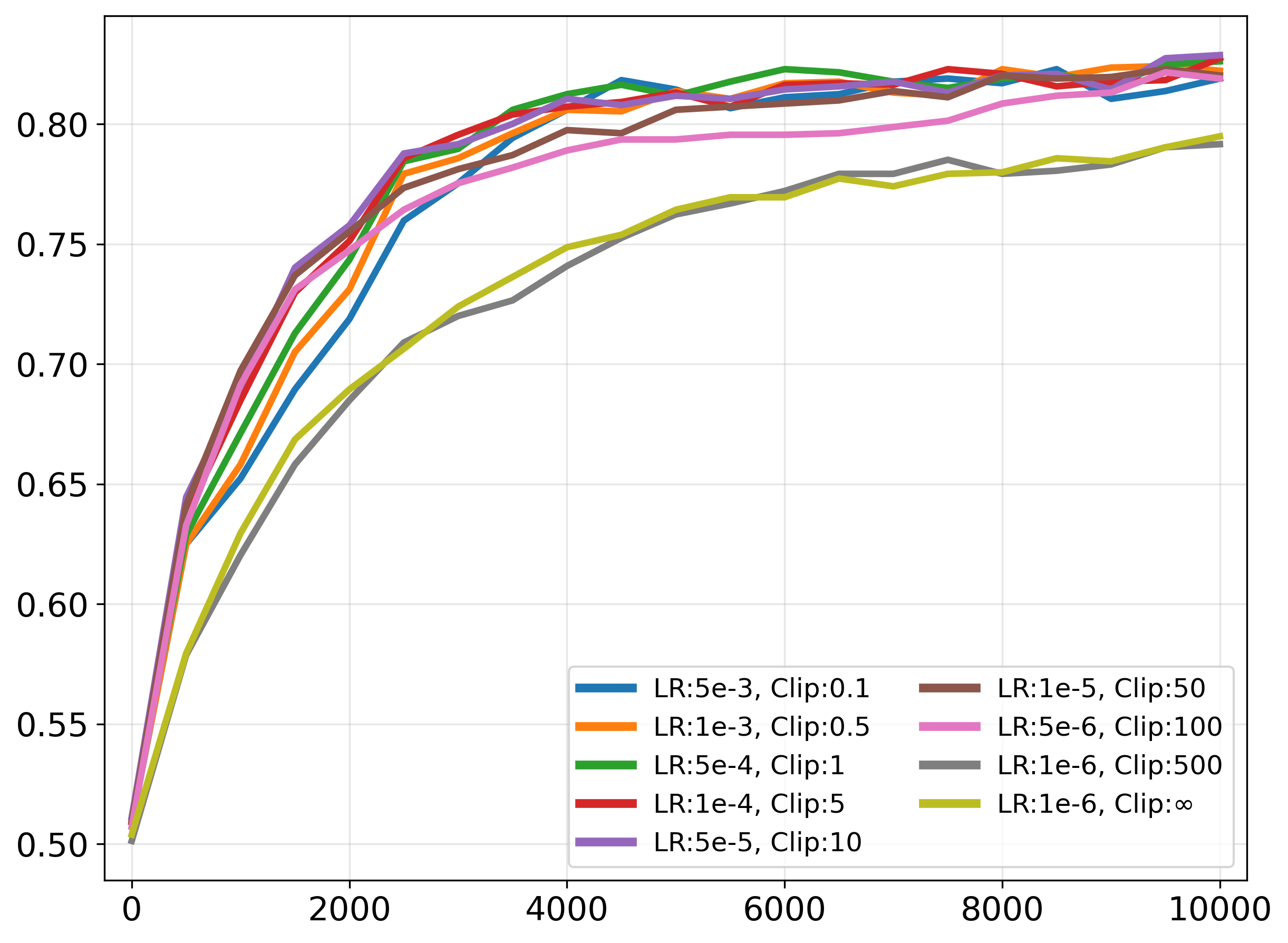}
    \put(-135,-8){fine-tuning iterations $t$}
    \put(-200,72){\rotatebox[origin=c]{90}{\small Accuracy}}
\caption{\sg{Optimization trajectories on SNLI for the extended $\eta c = 5\times 10^{-4}$ trade-off configurations ($c \in \{0.1, 0.5, 1, 5, 10, 50, 100, 500, \infty\}$). (\textbf{left}) Euclidean distance $\|w_t - w_0\|$ from the pretrained initialization over $10{,}000$ fine-tuning iterations, showing that all tightly clipped configurations down to $c=0.1$ follow the same high-distance trajectory ($\|w_t - w_0\| \in [85, 105]$), well separated from $c \ge 500$ ($\|w_t - w_0\| \in [52, 58]$). (\textbf{right}) Corresponding SNLI test accuracy trajectories over training, where all configurations with $c \in [0.1, 100]$ rapidly converge to the higher test accuracy plateau ($\sim 82\%$--$83\%$) compared to $c \ge 500$ ($\sim 79\%$).}}
\label{fig:snli_small_c_traj}
\end{center}
\end{figure}

\begin{figure}[h!]
\begin{center}
    \includegraphics[width=0.45\columnwidth]{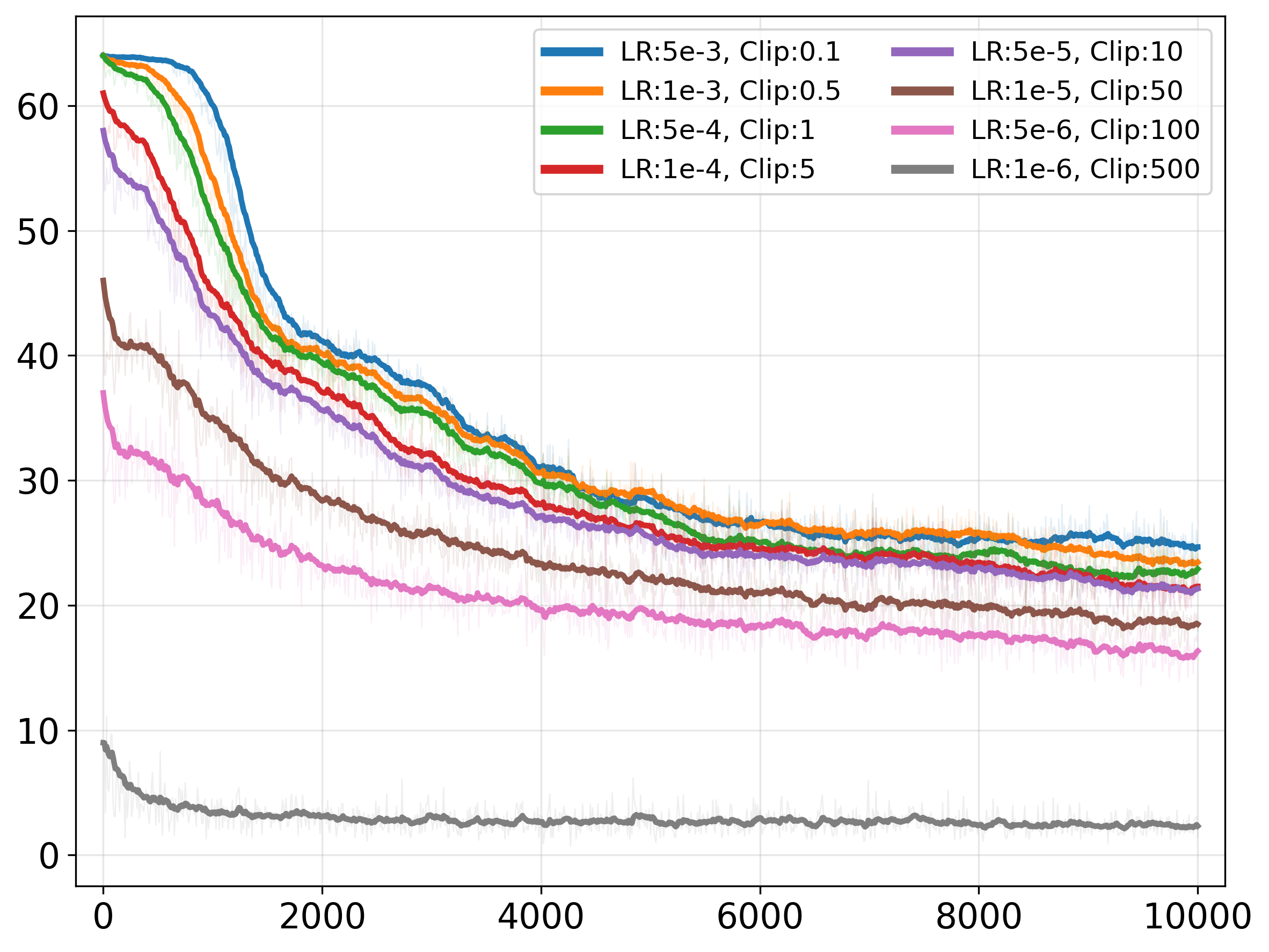}
    \put(-125,-8){\small fine-tuning iterations $t$}
    \put(-190,70){\rotatebox[origin=c]{90}{ \# Clipped Grads per Iterations }}
    \hspace{0.2cm}
    \includegraphics[width=0.52\columnwidth]{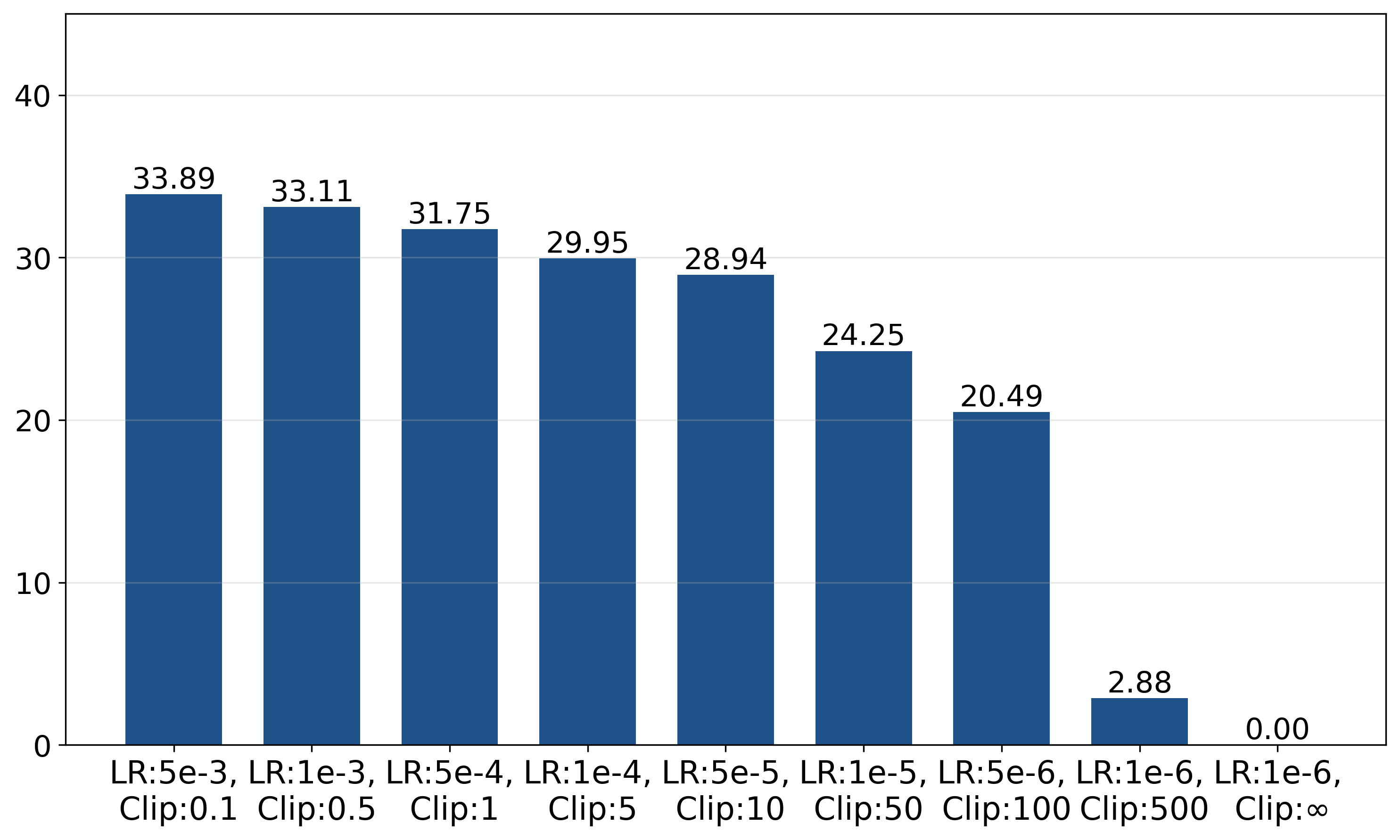}
    \put(-111,102){\small avg. \# of clipped gradients}
\caption{\sg{Per-sample gradient clipping counts on SNLI across the extended $\eta c = 5\times 10^{-4}$ configurations (batch size $= 64$). (\textbf{left}) Number of clipped per-sample gradients per iteration across the $10{,}000$ training steps, starting near the full batch size ($60$--$64$ out of $64$ for $c \le 5$) at initialization and decreasing over the first $\sim 4{,}000$ steps as the model adapts. (\textbf{right}) Average number of clipped gradients per iteration across all $10{,}000$ steps.}}
\label{fig:snli_small_c_clips}
\end{center}
\end{figure}

\sg{To examine optimization dynamics under smaller clipping thresholds ($c \le 1$), we expand the SNLI hyperparameter sweep to include $c \in \{0.1, 0.5, 1.0\}$ and correspondingly larger learning rates $\eta \in \{5\times 10^{-4}, 10^{-3}, 5\times 10^{-3}\}$ along the $\eta c = 5\times 10^{-4}$ diagonal (Figures~\ref{fig:snli_small_c}--\ref{fig:snli_small_c_clips}). Across all small-$c$ configurations on the diagonal, the $\eta c = 5\times 10^{-4}$ scaling relationship and test accuracy gains ($82.7\%$--$83.7\%$ vs.\ $80.2\%$ at $c=\infty$) continue to hold (Figure~\ref{fig:snli_small_c}, left), while smaller $c$ values plateau at higher training loss (Figure~\ref{fig:snli_small_c}, right). The Euclidean distance trajectories $\|w_t - w_0\|$ (Figure~\ref{fig:snli_small_c_traj}, left) and test accuracy curves (Figure~\ref{fig:snli_small_c_traj}, right) likewise follow the same behavior as $c \in [5, 100]$. Finally, tracking the number of clipped gradients per iteration over training (Figure~\ref{fig:snli_small_c_clips}, left) shows why the $10{,}000$-step average clip count is below $64$: for $c \le 5$, nearly all $64$ per-sample gradient estimates in each batch are clipped at the beginning of training ($60$--$64$ out of $64$), and as the model adapts over the first $\sim 4{,}000$ steps, fewer gradients exceed $c$, pulling the $10{,}000$-step average down to $31.75$--$33.89$ for $c \in [0.1, 1]$ (Figure~\ref{fig:snli_small_c_clips}, right).}

\section{Additional experiments from Section~\ref{sec:tradeoff}} 
\label{app:sec3}

\subsection{Loss Landscape}
\label{sec:loss_landscape_test/train}
\begin{figure}[h]
\vskip -0.1in
\begin{center}
\includegraphics[width=0.43\columnwidth]{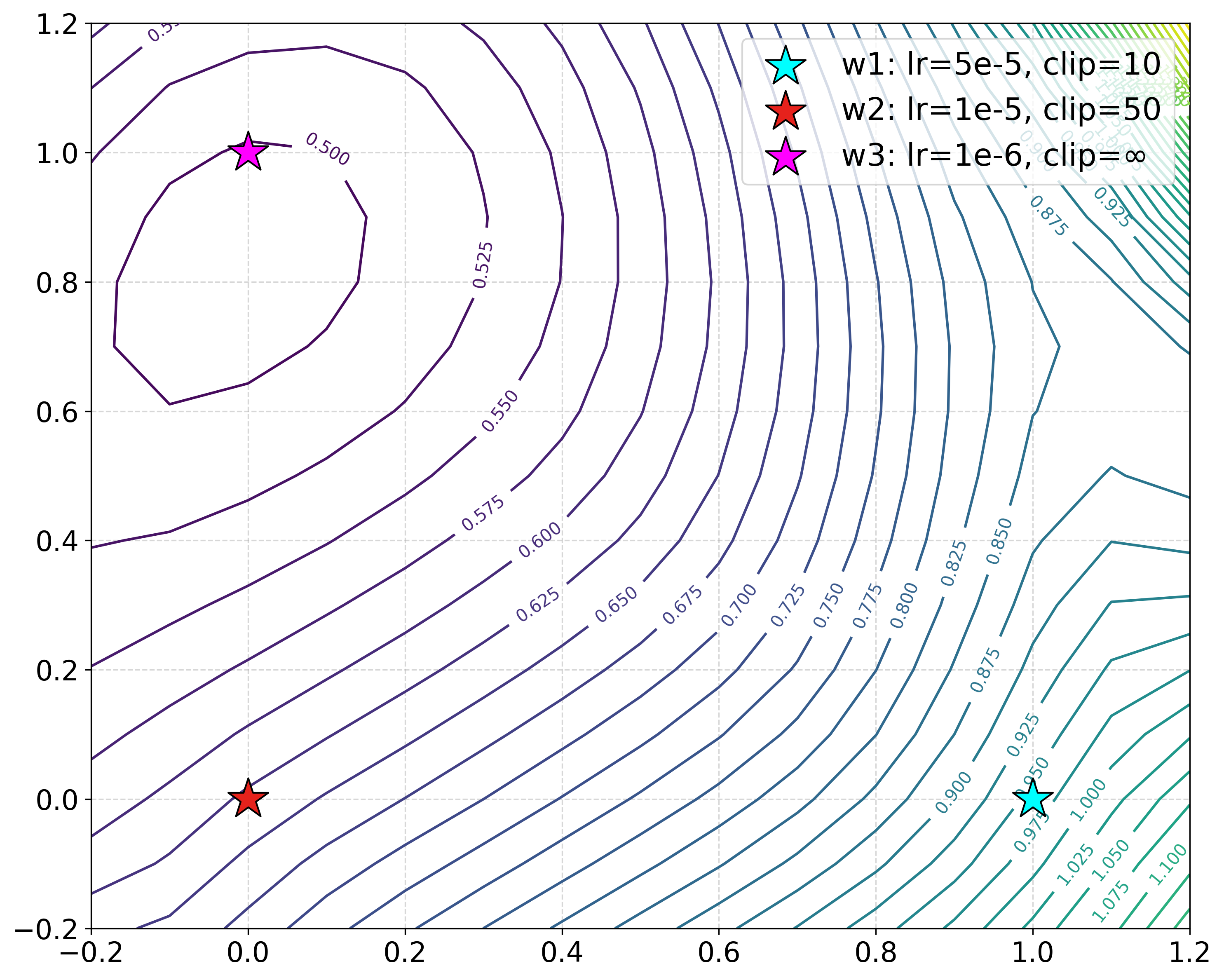}
    \put(-85,-5){$x$}
    \put(-180,65){$y$}
    \;\;\;\;\;\;
\includegraphics[width=0.43\columnwidth]{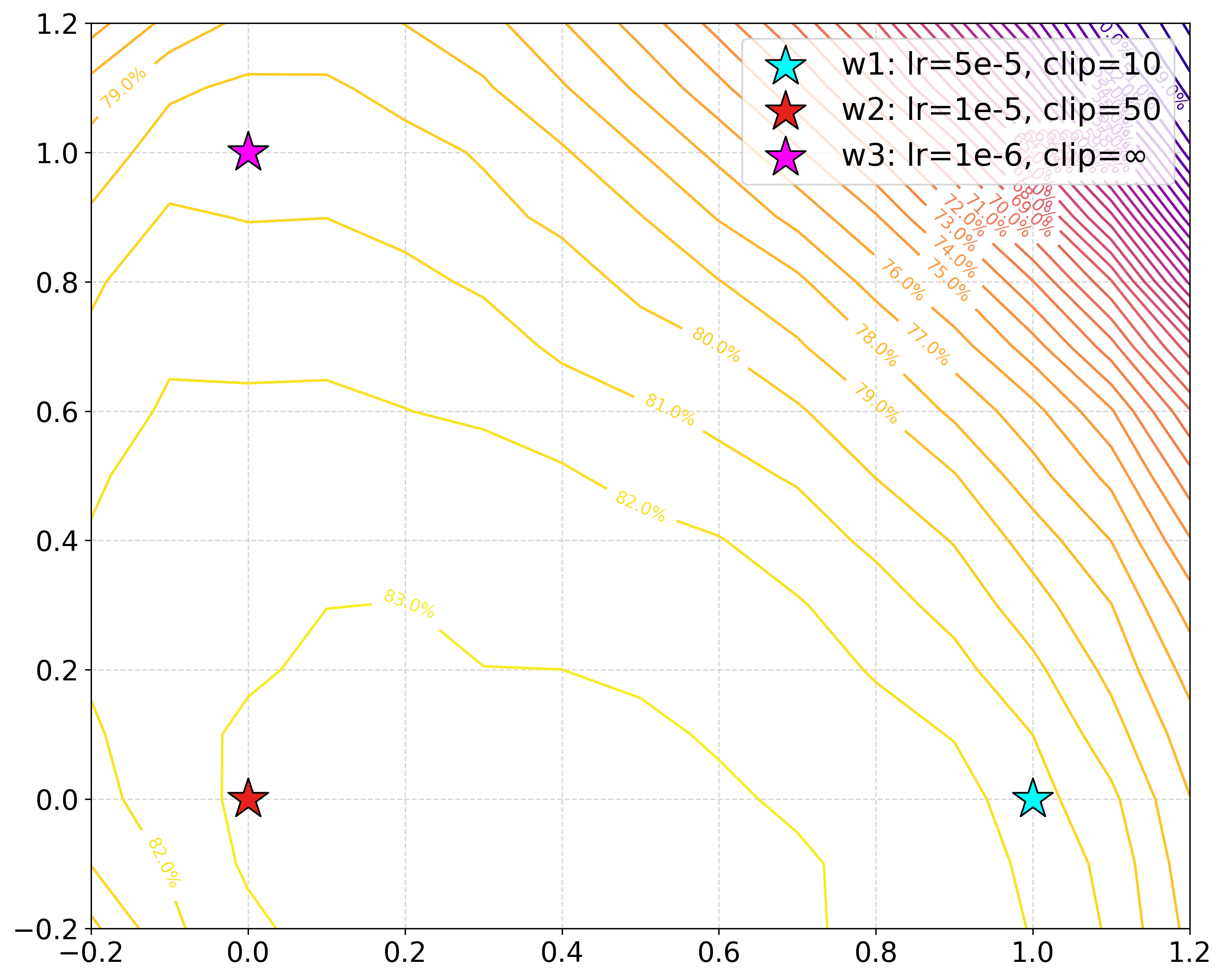}
    \put(-85,-5){$x$}
    \put(-180,65){$y$}    
\end{center}
\vskip -0.1in
\caption{Pretrained RoBERTa fine-tuned on SNLI. (left) Test loss landscape projected down to $(x,y)$-plane evaluated at $w_2+(w_1-w_2)x+(w_3-w_2)y$ for three  model checkpoints $w_1,w_2,w_3$. (right) Test Accuracy landscape projected down to $(x,y)$-plane evaluated at $w_2+(w_1-w_2)x+(w_3-w_2)y$ for three  model checkpoints $w_1,w_2,w_3$.}
\label{fig:mismatches_test}
\vskip -0.1in
\end{figure}

As discussed in Section \ref{sec:basin}, previous experiments suggest that hyperparameter choices along the \textit{clipping trade-off line} (i.e., $\eta c = 5\times10^{-4}$) converge to two different types of solutions depending on whether clipping is significant enough ($c\leq100$). To visualize the geometry of these solutions, similar to \citep{li2018visualizing,garipov2018loss}, we project the high-dimensional test loss and test accuracy landscapes onto the 2D plane spanned by three model checkpoints along this clipping trade-off line: $w_1$ ($\eta = 5e{-5}$, $c=10$), $w_2$ ($\eta = 1e{-5}$, $c=50$), and $w_3$ ($\eta = 1e{-6}$, $c=\infty$). Projecting these test metrics onto the exact same 2D plane defined by checkpoints \(w_1, w_2,\) and \(w_3\), we observe that the test landscapes perfectly mirror the training dynamics in \ref{fig:mismatches}. The test loss landscape (Figure~\ref{fig:mismatches_test}, left) shows that the unclipped baseline (\(w_3\)) resides in a local minima characterized by lower overall test loss. However, the test accuracy landscape (Figure~\ref{fig:mismatches_test}, right) demonstrates that the clipped models (\(w_1\) and \(w_2\)) navigate to a fundamentally distinct basin that achieves superior generalization accuracy, despite residing in a region of higher test loss. This confirms that the implicit bias introduced by per-sample clipping, trading higher loss for improved accuracy, is not an artifact of training set memorization, but a robust structural property that translates directly to test performance.

\subsection{Gradient Estimator Spread}
\label{sec:aestimator_distribution}

\begin{figure}[h]
\begin{center}
    \includegraphics[width=0.9\columnwidth]{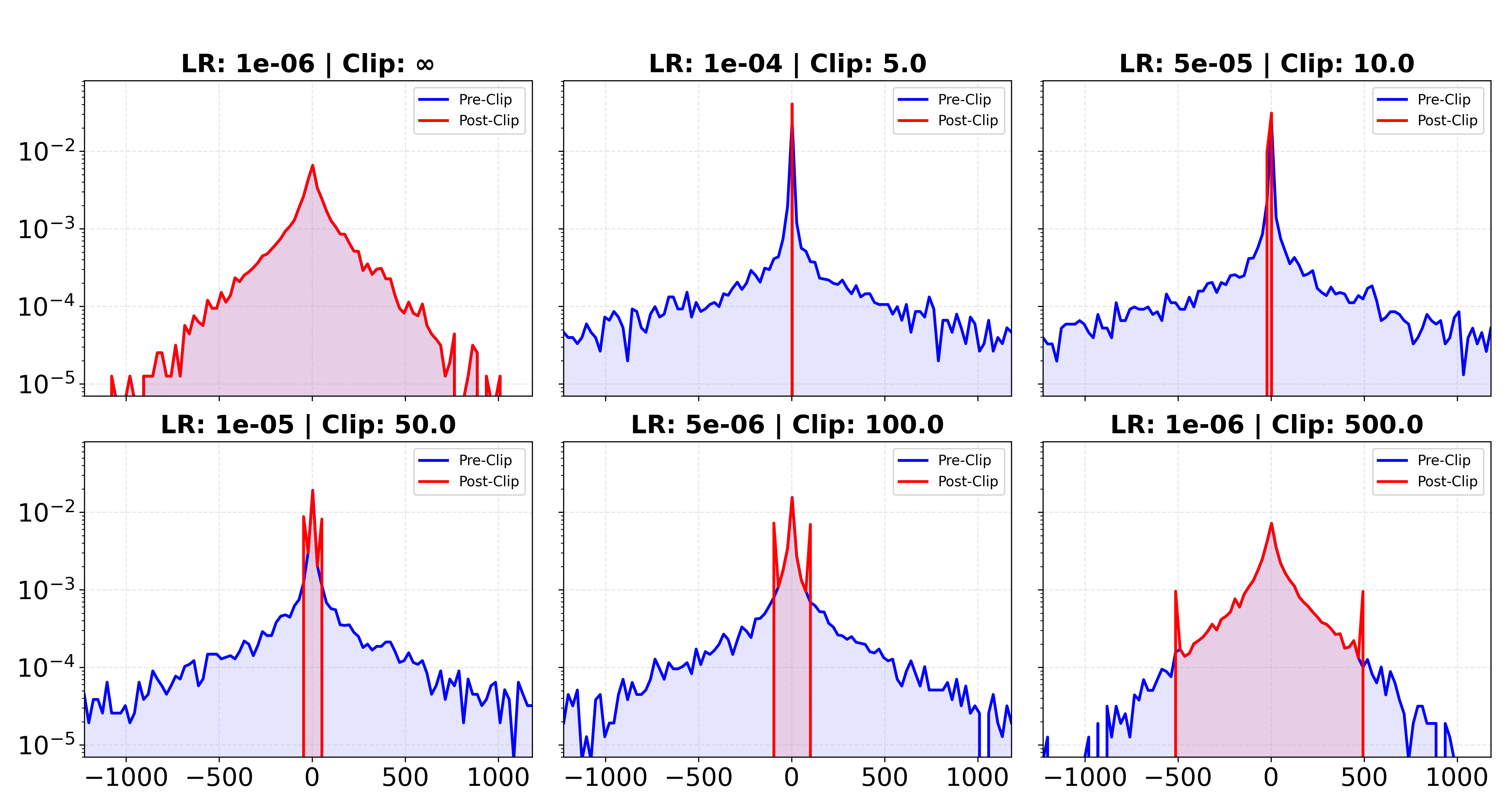}
    \put(-215,-7){Density (Log Scale)}
    \put(-370,95){\rotatebox[origin=c]{90}{Magnitude}}
    
\end{center}
\caption{Histogram depicting the spread of the gradient estimator before and after clipping}
\label{fig:histogram}
\end{figure}

\begin{figure}[h]
\begin{center}
    \includegraphics[width=0.9\columnwidth]{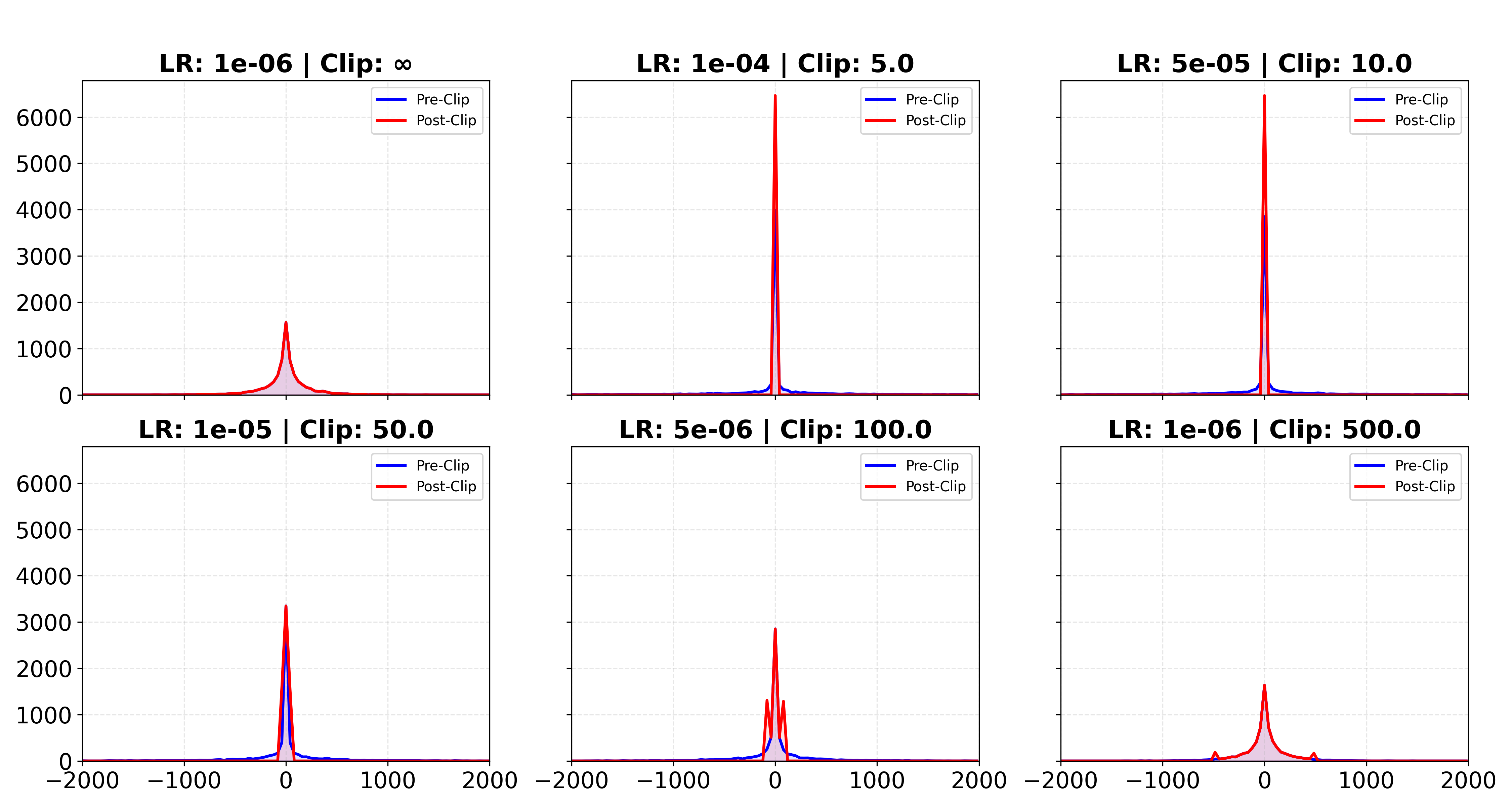}
    \put(-215,-7){Estimator Magintude}
    \put(-370,95){\rotatebox[origin=c]{90}{Count}}
    
\end{center}
\caption{Histogram depicting raw counts of gradient estimator before and after clipping}
\label{fig:histogram_raw}
\end{figure}

\label{app:estimator_distribution}

To physically observe how the clipping threshold controls the optimization step, we visualize the empirical distribution of the per-sample zeroth-order gradient estimators before and after clipping. Figure~\ref{fig:histogram} and Figure~\ref{fig:histogram_raw} display these distributions across six parameterizations sampled from the optimal diagonal, utilizing both raw counts to show the central mass and log-scaled density to highlight the tail behavior.

Examining the linear-scale count distributions highlights a distinct shift in the central mass of the estimator. While unclipped ZO estimation ($c=\infty$) naturally produces a smoothly decaying, wide-shouldered distribution around zero, the introduction of clipping severely distorts this geometry. Specifically, strict clipping constraints visually stretch the limits of the effective distribution while simultaneously inducing significantly thinner, sharper central peaks. This geometric transformation occurs because the clipping function forcibly redistributes the probability mass from the continuous heavy tails, pushing the values inward to aggregate exactly at the structural limits ($\pm c$). Consequently, the gradient estimator transitions from a broad, continuous spread into a highly concentrated, truncated structure where the update signal is strictly confined to the immediate center and the absolute clipping boundaries.

\section{Additional experiments from Section~\ref{sec:analysis}}
\label{sec:additional_analysis}

\subsection{Injected Label Noise}
\begin{figure}[h]
\vskip -0.1in
\begin{center}
\includegraphics[width=0.48\columnwidth]{images/iclr/SNLI_TEST_Accuracy_1e-6_vs_1e-5.png} 
    \put(-198,50){\rotatebox[origin=c]{90}{Accuracy}}
    \put(-130,110){SNLI Test Accuracy/Loss}
    \put(-142,100) {Comparison Mismatched Data}
    \includegraphics[width=0.48\columnwidth]{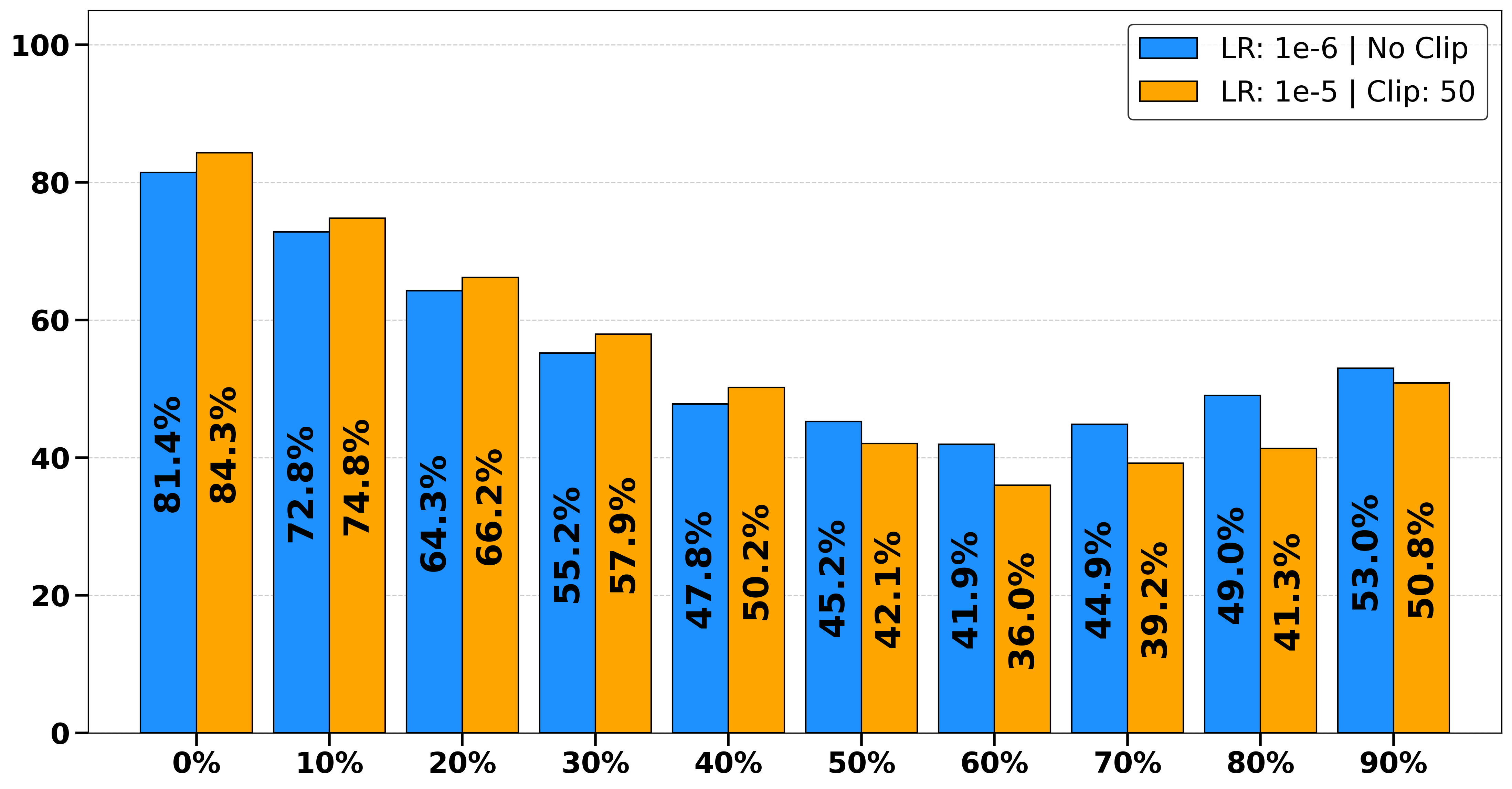} 
    \put(-130,110){SNLI Train Accuracy/Loss}
    \put(-142,100) {Comparison Mismatched Data}
    \hspace{0.2cm}
    \includegraphics[width=0.48\columnwidth]{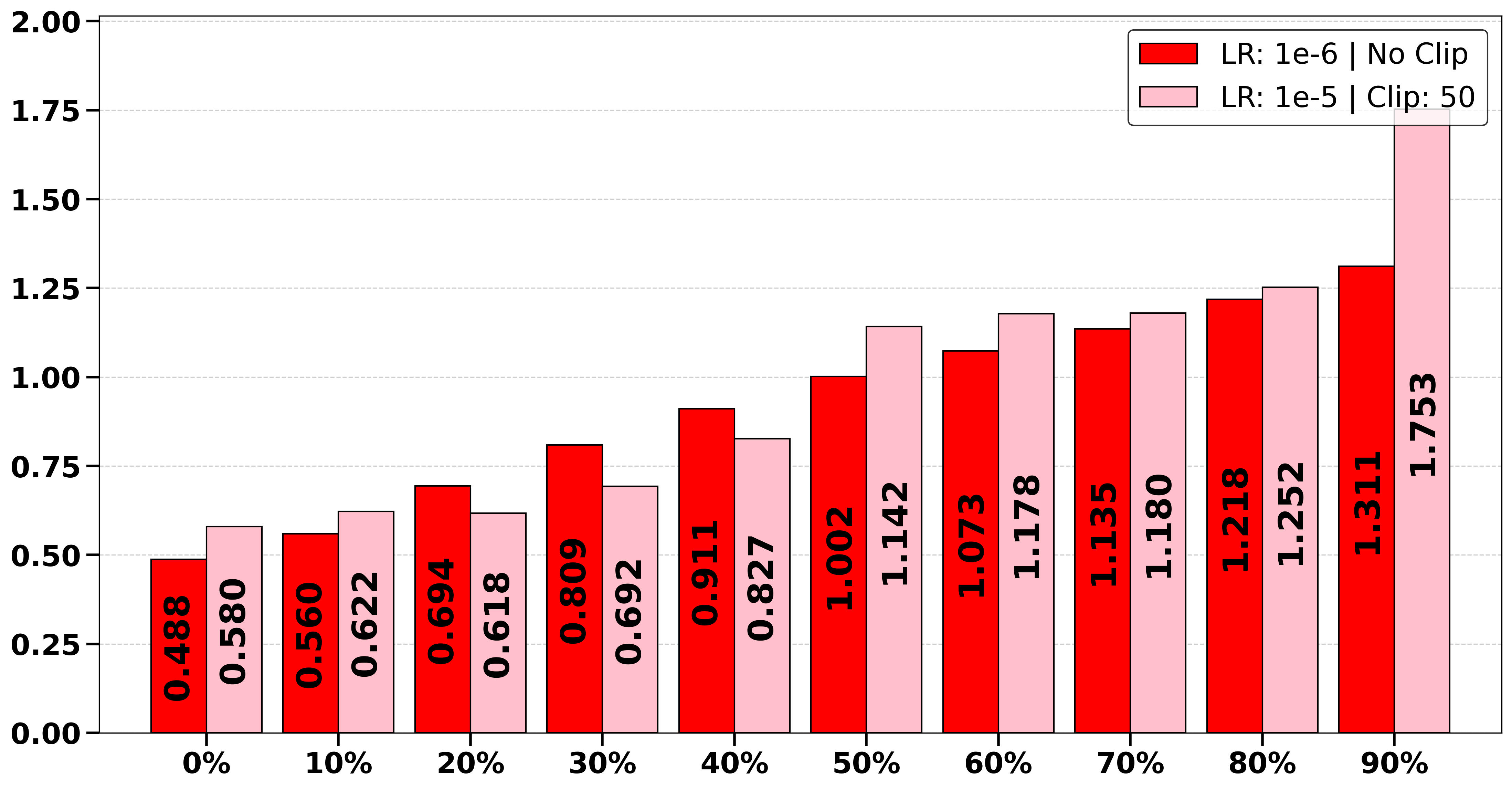} 
    \put(-130,-7){Data Mismatch Rate}
    \put(-198,55){\rotatebox[origin=c]{90}{Loss}}
    \includegraphics[width=0.48\columnwidth]{images/iclr/SNLI_TRAIN_Loss_1e-6_vs_1e-5.png} 
    \put(-130,-7){Data Mismatch Rate}

\end{center}
\vspace{-3mm}
\caption{Comparison of test and training accuracy, as well as cross-entropy loss, between clipped (\(\eta=1\times 10^{-5}, c=50.0\)) and unclipped (\(\eta=1\times 10^{-6}\)) ZO-SGD models across varying rates of injected label mismatch of SNLI training data set. The clipped model demonstrates increased robustness to label noise up to a 50\% corruption threshold, maintaining higher test accuracy despite systematically higher training losses.}
    \label{fig:mismatch_acc_loss_app}
\label{fig:accuracyloss}
\end{figure}

To further investigate the robustness of clipping, we introduce artificial label noise by randomly mislabeling subsets of the training data at varying corruption levels, ranging from 0\% to 90\% of the SNLI training set. Figure~\ref{fig:mismatch_acc_loss_app} demonstrates that in the regime where the injected noise fraction is between 0\% and 40\%, the clipped model (\(\eta=1\times 10^{-5}, c=50.0\)) is considerably more robust and maintains a higher test accuracy than the unclipped baseline. For example, at a 30\% corruption rate, the clipped model achieves a 73.7\% test accuracy compared to the unclipped model's 68.1\%. Generally, this robustness gap between clipped and unclipped ZO-SGD progressively widens as we inject more noisy labels, up until the corruption rate reaches 50\%. Once the injected noise reaches 50\% of the training data, the ground-truth labels become fundamentally ambiguous, causing both noise-injected and clean examples to behave similarly during optimization. Conversely, the training loss is consistently higher for clipped training across all corruption levels. This inflated loss may be occurring because the clipping mechanism essentially waters down the effect of poorly fitted samples. By explicitly capping the influence of mislabeled or inherently ambiguous examples, the mechanism seems introduces a systematic bias that prevents the model from minimizing the training loss through the memorization of bad data.

\subsection{Clipping leads to miscalibration} 
\label{sec:additional_analysis2}

\begin{figure}[h]

\begin{center}
\includegraphics[width=0.48\columnwidth]{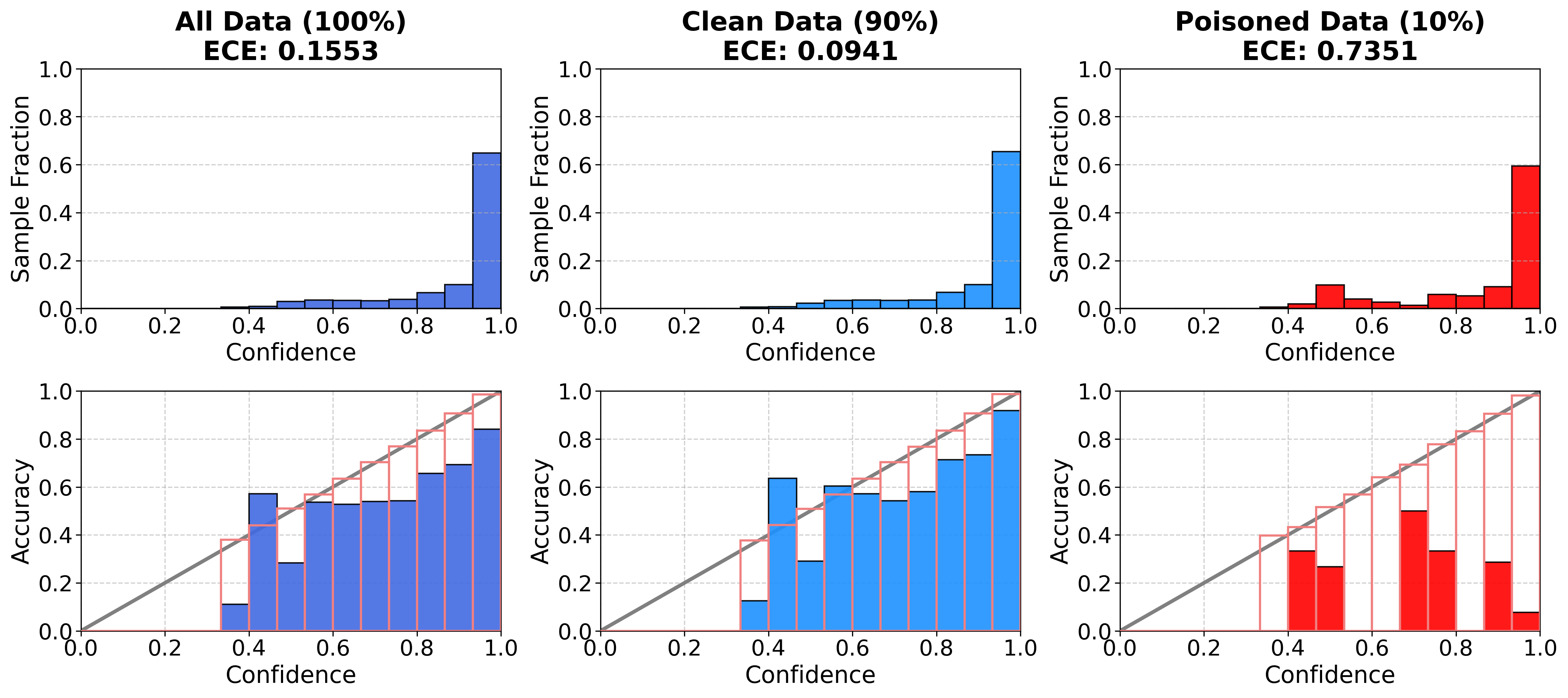}
\put(-150,100){SNLI Calibration Analysis 10\%} 
\put(-150,90) {Data Mismatch LR:1e-5, Clip:50}
\includegraphics[width=0.48\columnwidth]{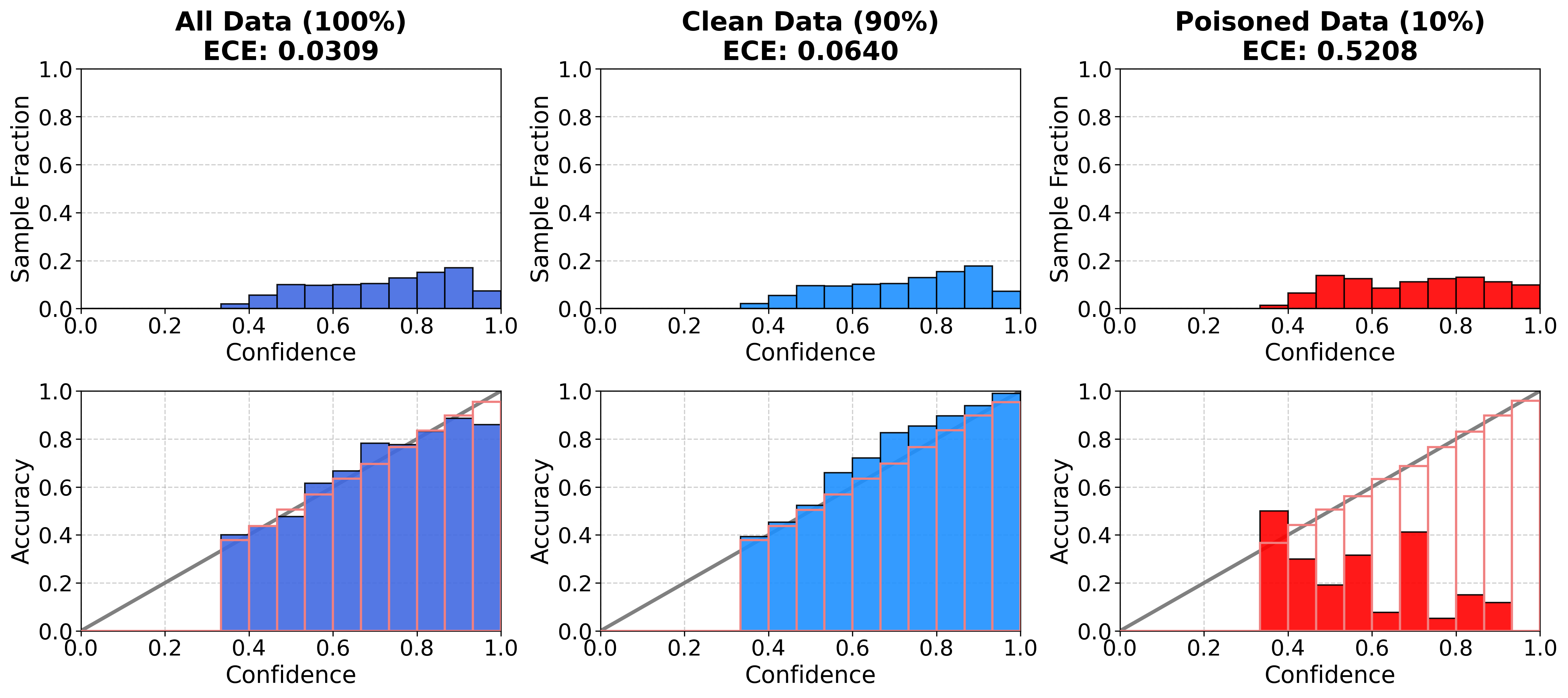} 
\put(-150,100){SNLI Calibration Analysis 10\%} 
\put(-150,90) {Data Mismatch LR:1e-6, Clip:0}
\end{center}
\vspace{-3mm}
\caption{Model Calibration for SNLI fine-tuning under 10\%, injected label noise. The plots compare the Expected Calibration Error (ECE) and confidence distributions for the clipped model (\(\eta=1\times 10^{-5}, c=50.0\)) \textbf{(left)} versus the unclipped baseline (\(\eta=1\times 10^{-6}, c=\infty\)) \textbf{(right)} across all data, clean data, and poisoned data subsets.}
\label{fig:calibration10}
\end{figure}

\begin{figure}[h]
\vskip -0.1in
\begin{center}
\includegraphics[width=0.48\columnwidth]{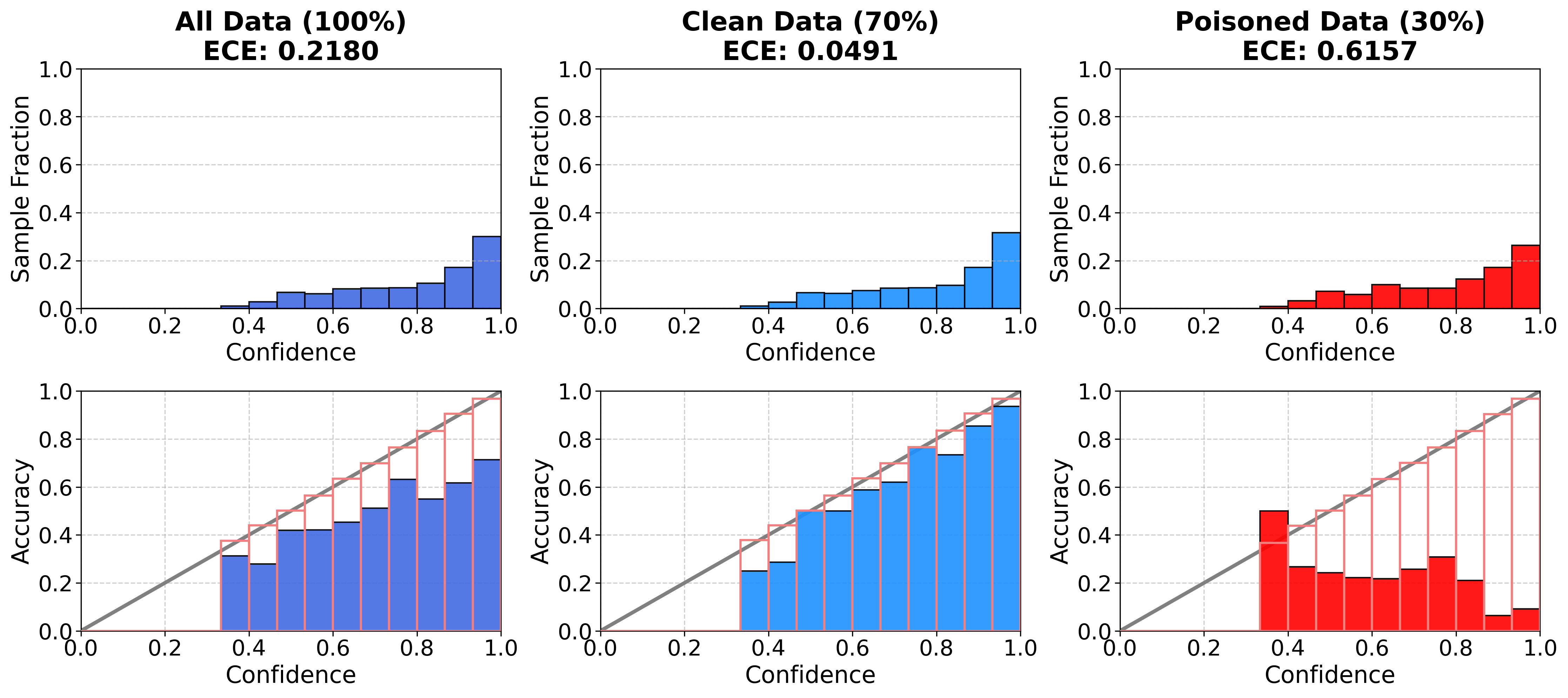}
\put(-150,100){SNLI Calibration Analysis 30\%} 
\put(-150,90) {Data Mismatch LR:1e-5, Clip:50}
\includegraphics[width=0.48\columnwidth]{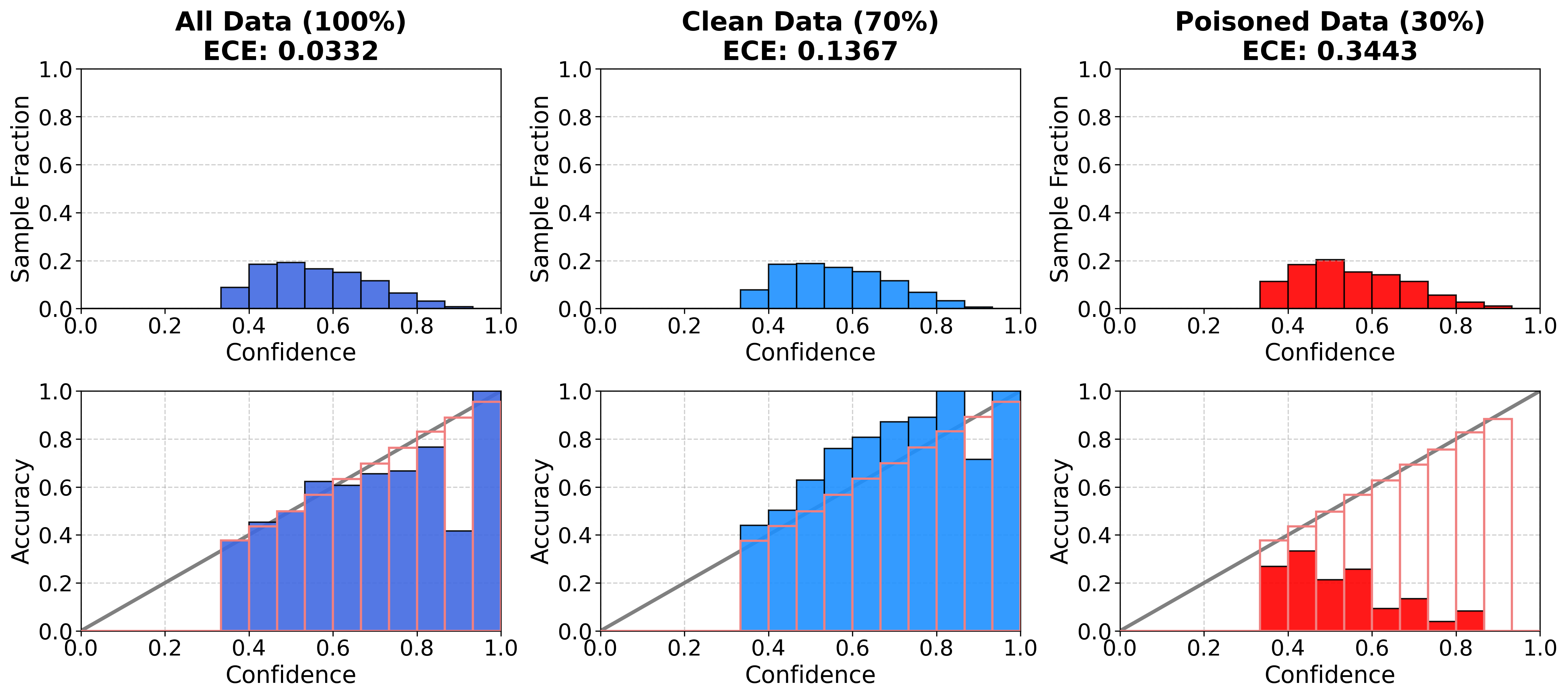} 
\put(-150,100){SNLI Calibration Analysis 30\%} 
\put(-150,90) {Data Mismatch LR:1e-6, Clip:0}
\end{center}
\vspace{-3mm}
\caption{Model Calibration for SNLI fine-tuning under 30\%, injected label noise. The plots compare the Expected Calibration Error (ECE) and confidence distributions for the clipped model (\(\eta=1\times 10^{-5}, c=50.0\)) \textbf{(left)} versus the unclipped baseline (\(\eta=1\times 10^{-6}, c=\infty\)) \textbf{(right)} across all data, clean data, and poisoned data subsets.}
\label{fig:calibration30}
\end{figure}

\begin{figure}[h]
\vskip -0.1in
\begin{center}
\includegraphics[width=0.48\columnwidth]{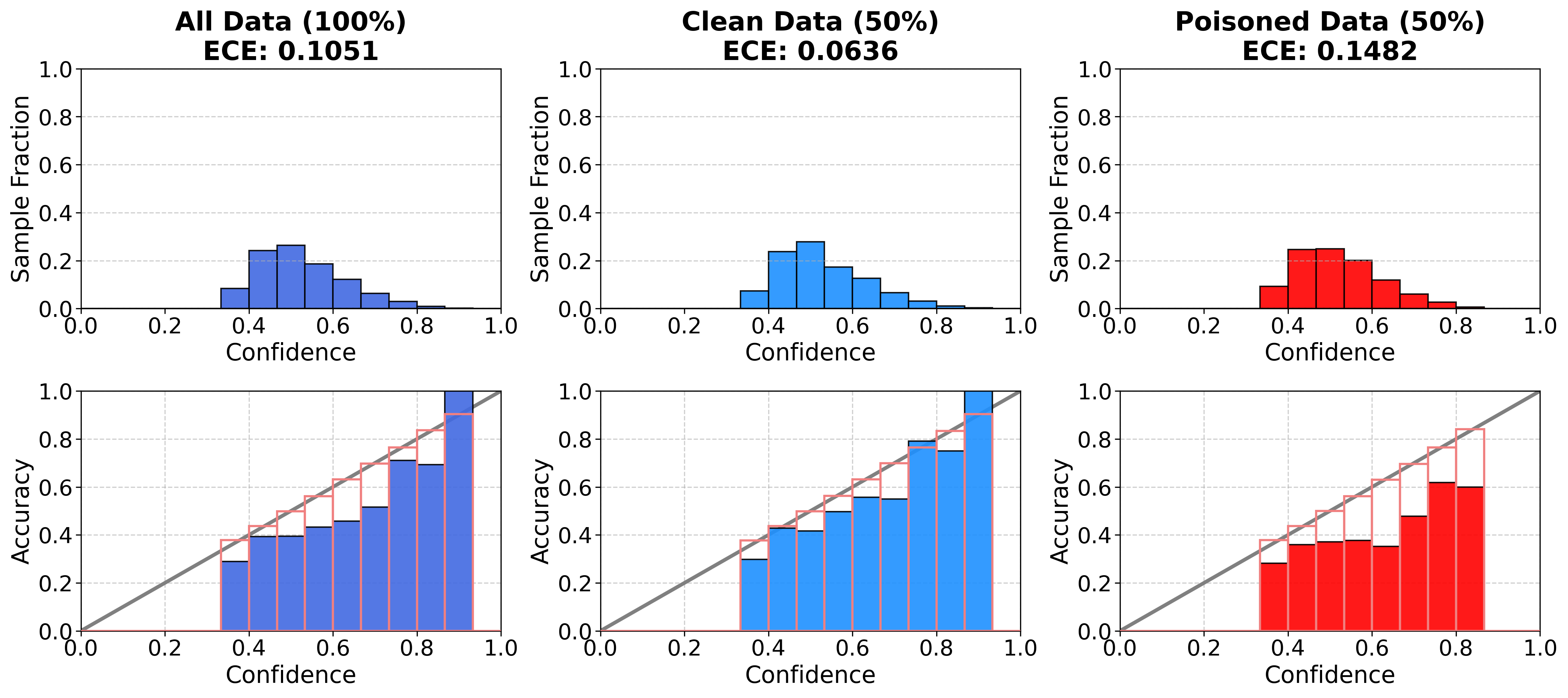}
\put(-150,100){SNLI Calibration Analysis 50\%} 
\put(-150,90) {Data Mismatch LR:1e-5, Clip:50}
\includegraphics[width=0.48\columnwidth]{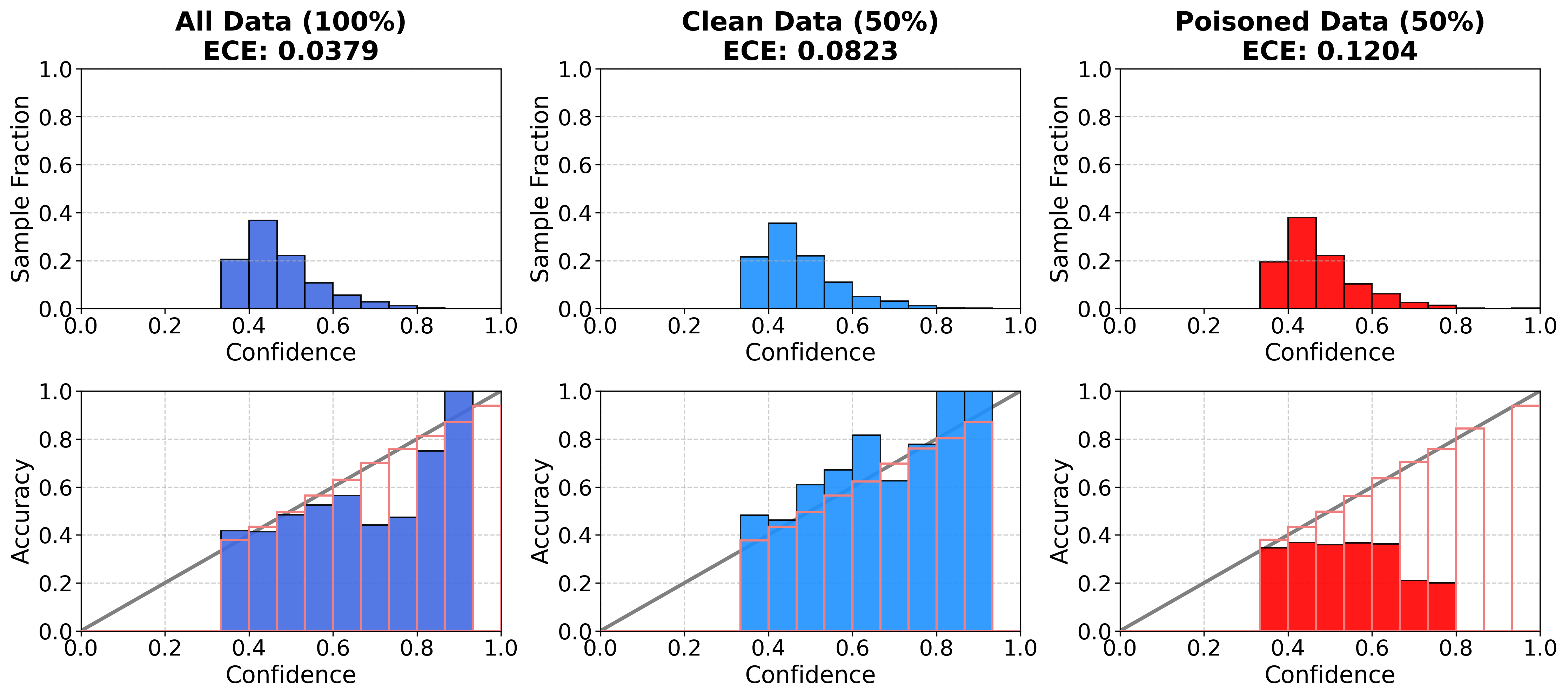} 
\put(-150,100){SNLI Calibration Analysis 50\%} 
\put(-150,90) {Data Mismatch LR:1e-6, Clip:0}
\end{center}
\vspace{-3mm}
\caption{Model Calibration for SNLI fine-tuning under 50\%, injected label noise. The plots compare the Expected Calibration Error (ECE) and confidence distributions for the clipped model (\(\eta=1\times 10^{-5}, c=50.0\)) \textbf{(left)} versus the unclipped baseline (\(\eta=1\times 10^{-6}, c=\infty\)) \textbf{(right)} across all data, clean data, and poisoned data subsets.}
\label{fig:calibration50}
\end{figure}

To provide a deeper empirical understanding of the predictive uncertainty discussed in Section 4.2, Figures~\ref{fig:calibration10}, \ref{fig:calibration30}, and \ref{fig:calibration50} detail the segmented calibration behaviors of the clipped and unclipped models under injected label noise. By partitioning the reliability diagrams into clean and poisoned subsets, we can isolate exactly how the clipping mechanism affects confidence and accuracy.

At a 10\% data mismatch rate, the clipped model maintains a low Expected Calibration Error (ECE) of 0.0941 on the clean data, but exhibits a massive ECE spike of 0.7351 specifically on the poisoned data. This divergence is even more pronounced at the 30\% mismatch rate, where the clipped model achieves a highly calibrated ECE of 0.0491 on clean data, while the poisoned ECE remains heavily inflated at 0.6157. Furthermore, the histograms reveal the clipped model concentrates the vast majority of its prediction mass in the highest confidence bin (\(>0.9\)) across both clean and poisoned samples, creating a severe right-skewed distribution. Because the model correctly resists memorizing the poisoned labels, its empirical accuracy relative to those incorrect labels is exceptionally low. This combination of maximum confidence and minimal empirical accuracy pulls the reliability bars far below the perfect calibration diagonal, forming a stark horizontal flatline near zero accuracy in the reliability diagram. This mathematically explains the massive ECE spikes on the rejected poisoned subset.

Conversely, the unclipped baseline exhibits seemingly better calibration on the poisoned data at the 30\% mismatch rate, achieving an ECE of 0.3443 compared to the clipped model's 0.6157. Visually, the unclipped model's sample fraction histogram is much broader and flatter, distributing prediction mass across intermediate confidence bins rather than strictly at the extremes. However, this lower ECE is a symptom of optimization failure rather than generalization success, as the unclipped model achieves better calibration here because it actively memorizes the noisy labels. Its reliability bars form a gradual, upward-sloping staircase that tracks closer to the diagonal, bringing its predictions into closer alignment with the incorrect poisoned targets. Once the data corruption rate reaches the 50\% threshold, the ground-truth signals become fundamentally ambiguous, and the robust decision boundary of the clipped model breaks down. At this stage, the clipped model loses its high-confidence predictive structure across all subsets, with its sample fraction mass shifting abruptly away from the right edge and dispersing toward the center, heavily resembling the flatter shape of the unclipped baseline. As a result, the poisoned ECE of the clipped model drops to 0.1482, converging toward the unclipped model's ECE of 0.1204. At this point, the reliability diagrams for both models exhibit similarly flattened and dispersed prediction distributions.

{\color{black}
\subsection{Separating Mislabeled Examples vs.\ Difficult Clean Examples on Accuracy and Calibration.}
\label{sec:abc_disentangle}

To isolate whether the accuracy gains of clipped ZO-SGD stem from suppressing mislabeled outliers or from downweighting correctly labeled but difficult boundary examples, we partition the $10\%$-corrupted SNLI training set ($N=1{,}536$) into three disjoint subsets based on per-sample clipping frequencies recorded during clipped ZO-SGD training ($\eta=10^{-5}, c=50$):
\begin{enumerate}
    \item \textbf{Subset $A$ - Synthetic Mislabeled Data, $\sim10\%$:} Training examples with intentionally flipped labels.
    \item \textbf{Subset $B$ - Natural Difficult / High-Clip Data, $\sim15\%$:} Uncorrupted examples whose per-sample directional derivative exceeded the clipping threshold $c=50$ more than $\tau=375$ times across $10{,}000$ steps. 
    \item \textbf{Subset $C$ Clean Low-Clip Data, $\sim75\%$:} The remaining uncorrupted examples ($\text{clip\_count} \le 375$).
\end{enumerate}

\sg{We train both unclipped ZO-SGD ($\eta=10^{-6}$) and clipped ZO-SGD ($\eta=10^{-5}, c=50$) from scratch on four subsets, $C$, $C+B$, $C+A$, and $C+B+A$, for $10{,}000$ steps (Table~\ref{tab:abc_disentangle}). We focus on these four configurations rather than degenerate combinations such as $A$, $B$, or $A+B$, which exclude the core clean distribution $C$ required for learning the task and contain only small fractions of mislabeled or ambiguous examples. Evaluating the two parallel nested progressions, $C \subset C+B \subset C+B+A$ and $C \subset C+A \subset C+B+A$, allows us to independently measure the impact of each data type.}

\begin{table}[h]
\caption{\sg{Test accuracy ($\%$) of unclipped and clipped ZO-SGD across nested SNLI data subsets $C$, $C+B$, $C+A$, and $C+B+A$. The results demonstrate that unclipped ZO-SGD performs worse as natural difficult examples ($B$) and synthetic mislabeled examples ($A$) are added, whereas clipped ZO-SGD remains completely invariant to natural difficult data ($81.1\% \to 81.3\%$) and highly resilient to synthetic label noise ($79.4\%$). Crucially, when both $A$ and $B$ are manually pruned (Subset $C$), unclipped ZO-SGD ($81.5\%$) fully recovers and matches clipped ZO-SGD ($81.1\%$), confirming that clipping improves generalization by acting as an automatic outlier filter against both natural and synthetic high-loss gradients.}}
\centering
\label{tab:abc_disentangle}
\resizebox{0.85\textwidth}{!}{%
\begin{tabular}{lcccc}
\toprule
\textbf{Method} & \textbf{Subset $C$} & \textbf{Subset $C+B$} & \textbf{Subset $C+A$} & \textbf{Subset $C+B+A$} \\
\midrule
\textbf{Unclipped ZO-SGD} ($\eta=10^{-6}$) & ${81.5\%}$ & $80.4\%$ & $78.3\%$ & $77.6\%$ \\
\midrule
\textbf{Clipped ZO-SGD} ($\eta=10^{-5}, c=50$) & $81.1\%$ & ${81.3\%}$ & ${79.1\%}$ & ${79.4\%}$ \\
\bottomrule
\end{tabular}%
}
\vspace{2pt}
\end{table}

The results in Table~\ref{tab:abc_disentangle} separate the sources of accuracy improvement across the two optimization regimes by isolating the impact of natural difficult examples and synthetic mislabeled examples. First, comparing subset $C$ against $C+B$ reveals that adding naturally difficult, high-clip examples ($B$) to the clean subset ($C$) reduces unclipped ZO-SGD test accuracy by $1.1\%$ ($81.5\% \to 80.4\%$). In contrast, clipped ZO-SGD is unaffected by the inclusion of $B$, maintaining its accuracy ($81.1\%$ on $C$ vs.\ $81.3\%$ on $C+B$). This indicates that naturally ambiguous or noisy examples in the original dataset destabilize unclipped zeroth-order updates, whereas per-sample clipping bounds their influence so they do not degrade the learned decision boundary. Second, introducing the $10\%$ synthetically corrupted subset $A$ on top of the uncorrupted data ($C+B \to C+B+A$) causes a sharp accuracy drop from $80.4\% \to 77.6\%$ in unclipped ZO-SGD as the model memorizes poisoned labels. Under clipped ZO-SGD, accuracy drops from $81.3\% \to 79.4\%$, outperforming unclipped ZO-SGD on the full corrupted dataset by almost $2$\%. To verify this effect independently, the $C+A$ subset acts as a control configuration to measure the injection of synthetic noise without the confounding presence of natural boundary examples. Adding $A$ directly to the clean subset ($C \to C+A$) confirms this robustness, triggering a $3.2\%$ drop in unclipped ZO-SGD ($81.5\% \to 78.3\%$) while clipped ZO-SGD remains highly resilient with only a $2.0\%$ drop ($81.1\% \to 79.1\%$). Moreover, when both synthetic mislabeled examples ($A$) and natural difficult examples ($B$) are removed, unclipped ZO-SGD ($81.5\%$) completely closes the performance gap and matches clipped ZO-SGD ($81.1\%$). Together, the $+2.8\%$ gain from removing $A$ and the $+1.1\%$ gain from removing $B$ in unclipped ZO-SGD demonstrate that per-sample clipping improves accuracy by simultaneously suppressing gradient spikes from both mislabeled outliers and naturally difficult examples without requiring offline data pruning.

\begin{figure}[!b]
\vspace{-2mm}
    \centering
    \includegraphics[width=0.48\textwidth]{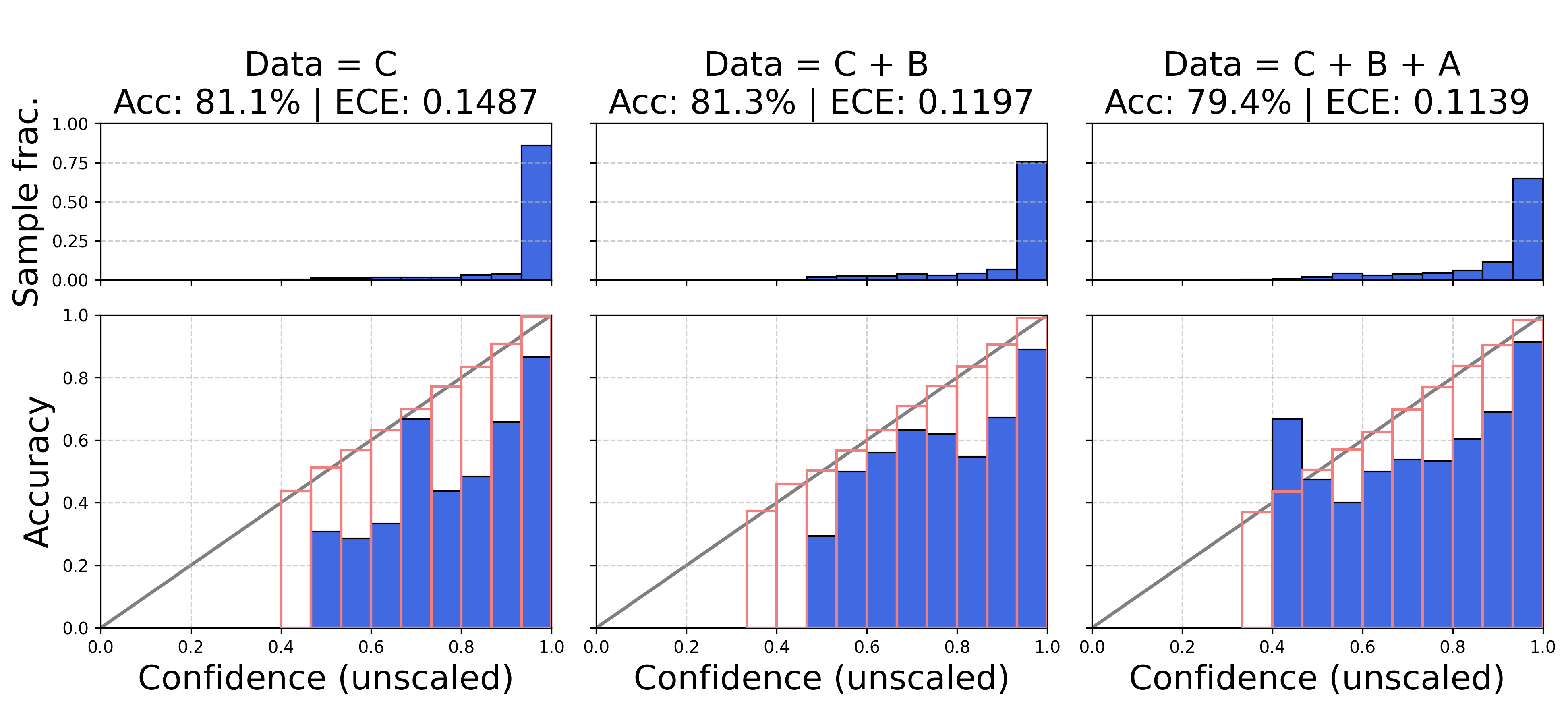}
    \put(-150,90){Clipped ZO-SGD (LR=1e-5, c=50)} 
    \includegraphics[width=0.48\textwidth]{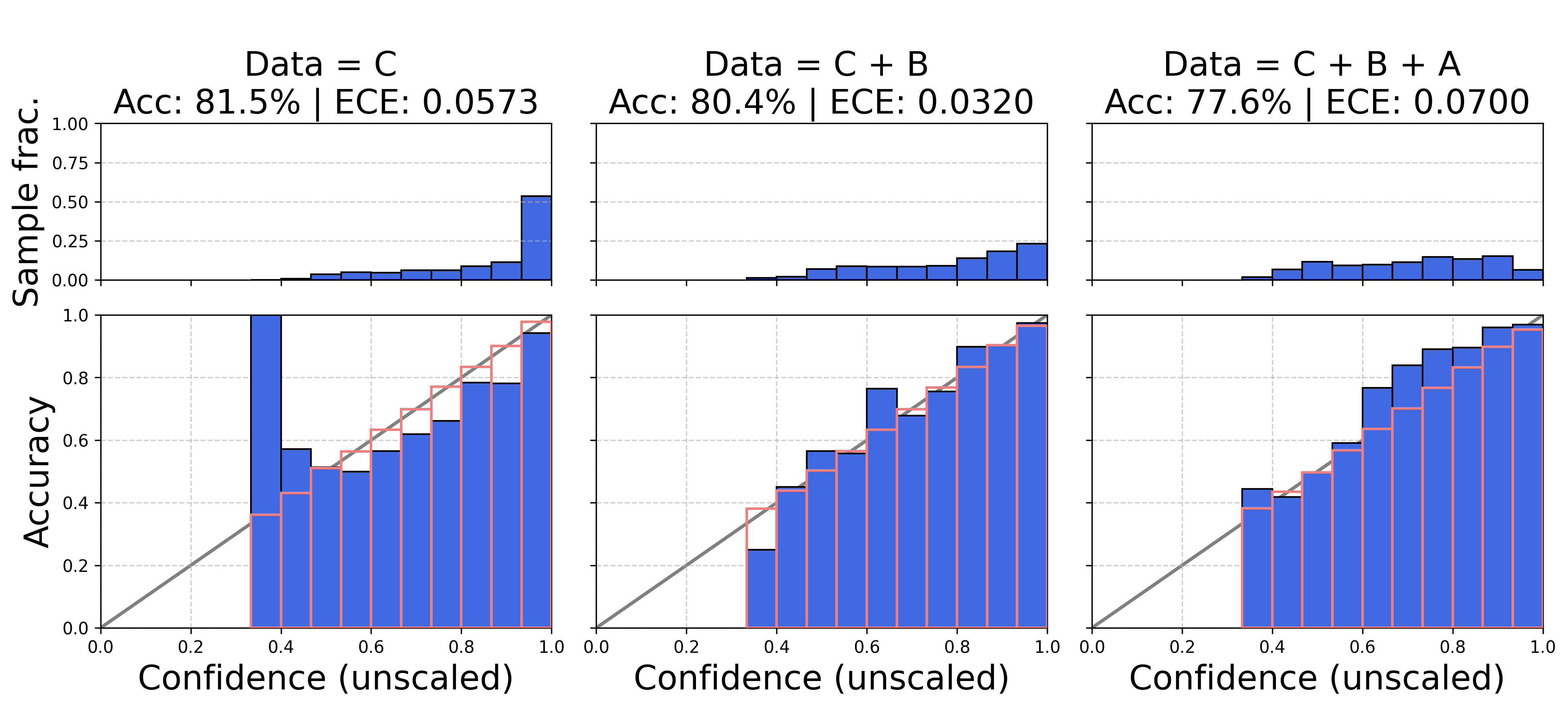}
    \put(-150,90){Unclipped ZO-SGD (LR=1e-6, c=$\infty$)} 
    \caption{\sg{Reliability diagrams and confidence histograms on SNLI across nested subsets $C$, $C+B$, and $C+B+A$ for Clipped ZO-SGD ($\eta=10^{-5}, c=50$) and Unclipped ZO-SGD ($\eta=10^{-6}, c=\infty$). {Left:} Clipped ZO-SGD is consistently overconfident (red confidence bars exceed blue accuracy bars), with top-bin ($>0.93$) concentration and ECE increasing monotonically as $A$ and $B$ are removed ($0.1139 \to 0.1197 \to 0.1487$).{Right:} Unclipped ZO-SGD transitions from \emph{underconfident} on $C+B+A$ ($\text{ECE}=0.0700$, blue accuracy bars exceed red confidence bars due to memorizing corrupted labels in $A$), to \emph{well-calibrated} on $C+B$ ($\text{ECE}=0.0320$), to \emph{overconfident} on clean low-clip data $C$ ($\text{ECE}=0.0573$, with $>53\%$ of predictions concentrating in the highest confidence bin), demonstrating that removing difficult boundary examples ($B$) directly inflates model confidence.}}
    \label{fig:abc_calibration}
\end{figure}

Figure~\ref{fig:abc_calibration} compares the calibration diagrams across $C$, $C+B$, and $C+B+A$ to investigate how mislabeled examples ($A$) versus difficult clean examples ($B$) affect calibration. First, comparing $C+B$ to $C+B+A$ shows that mislabeled examples ($A$) pull unclipped ZO-SGD toward underconfidence. When unclipped ZO-SGD is trained on $C+B+A$ (Figure~\ref{fig:abc_calibration}), memorizing the contradictory labels in $A$ flattens the output logits across classes, leaving the model systematically underconfident (blue accuracy bars exceed red confidence bars across intermediate bins, $\text{ECE}=0.0700$) and reducing test accuracy to $77.6\%$. Removing $A$ ($C+B$) eliminates this label contradiction and restores calibration ($\text{ECE}=0.0320$). Under clipped ZO-SGD, per-sample clipping caps the opposing gradient spikes from $A$, preventing them from flattening the model's confidence ($\text{ECE}$ remains stable at $0.1139$ vs.\ $0.1197$).

Second, comparing $C+B$ to $C$ reveals that suppressing difficult clean boundary examples ($B$) seems to drive overconfidence. On $C+B$, the high-loss boundary examples in $B$ exert a strong gradient penalty against overconfident misclassifications, keeping unclipped ZO-SGD well-calibrated ($\text{ECE}=0.0320$, with only $\sim 23\%$ of predictions in the top $>0.93$ confidence bin). When $B$ is pruned ($C+B \to C$), even \emph{unclipped} ZO-SGD becomes systematically overconfident: the fraction of predictions in the highest confidence bin ($>0.93$) more than doubles to $>53\%$, red confidence bars rise above blue accuracy bars across bins $0.55$--$0.95$, and ECE nearly doubles from $0.0320$ to $0.0573$.

Finally, comparing unclipped and clipped ZO-SGD on Subset $C$ ($0.0573$ vs.\ $0.1487$) shows that per-sample clipping further increases this confidence concentration through the continuous reweighting of Proposition~\ref{propo:clippedgradient}. Because Subset $C$ retains examples clipped up to $375$ times, clipped ZO-SGD ($c=50$) continues to downweight moderate-gradient boundary examples within $C$ by $h_c(\|\nabla f_i(w)\|) = 2\Phi(c/\|\nabla f_i(w)\|)-1$ while updating along the majority vote of clean examples \eqref{eq:fullyclipped}, pushing $\sim 86\%$ of predictions into the $>0.93$ confidence bin.

}

\vspace{-2mm}
\section{Connections to Edge-of-Stability (EoS)}
\label{app:eos}
\begin{figure}[h]
\vskip -0.1in
\begin{center}
\includegraphics[width=0.31\columnwidth]{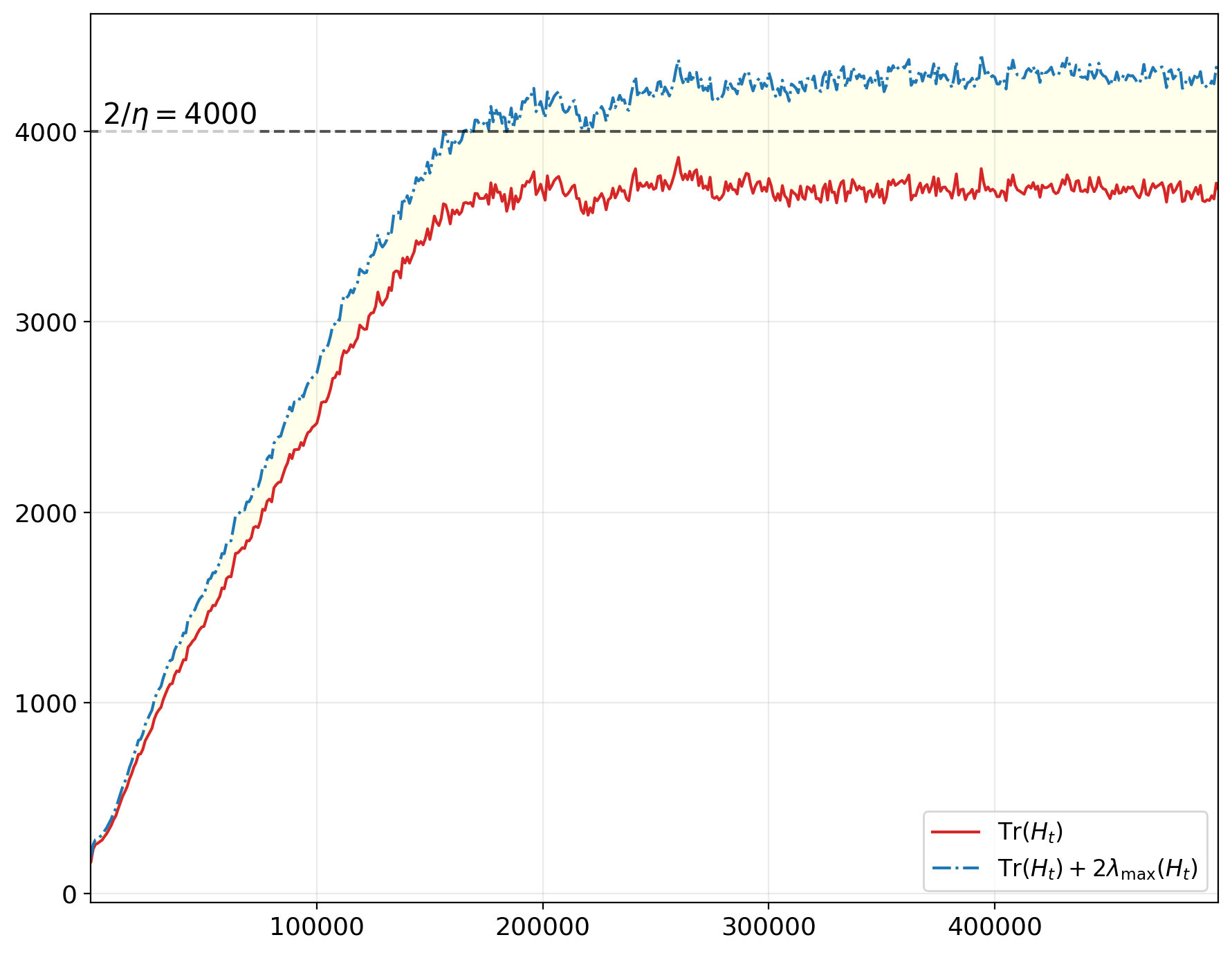}
    \put(-135,45){\rotatebox[origin=c]{90}{Hessian trace}}
    \hspace{-0.2cm}
\includegraphics[width=0.31\columnwidth]{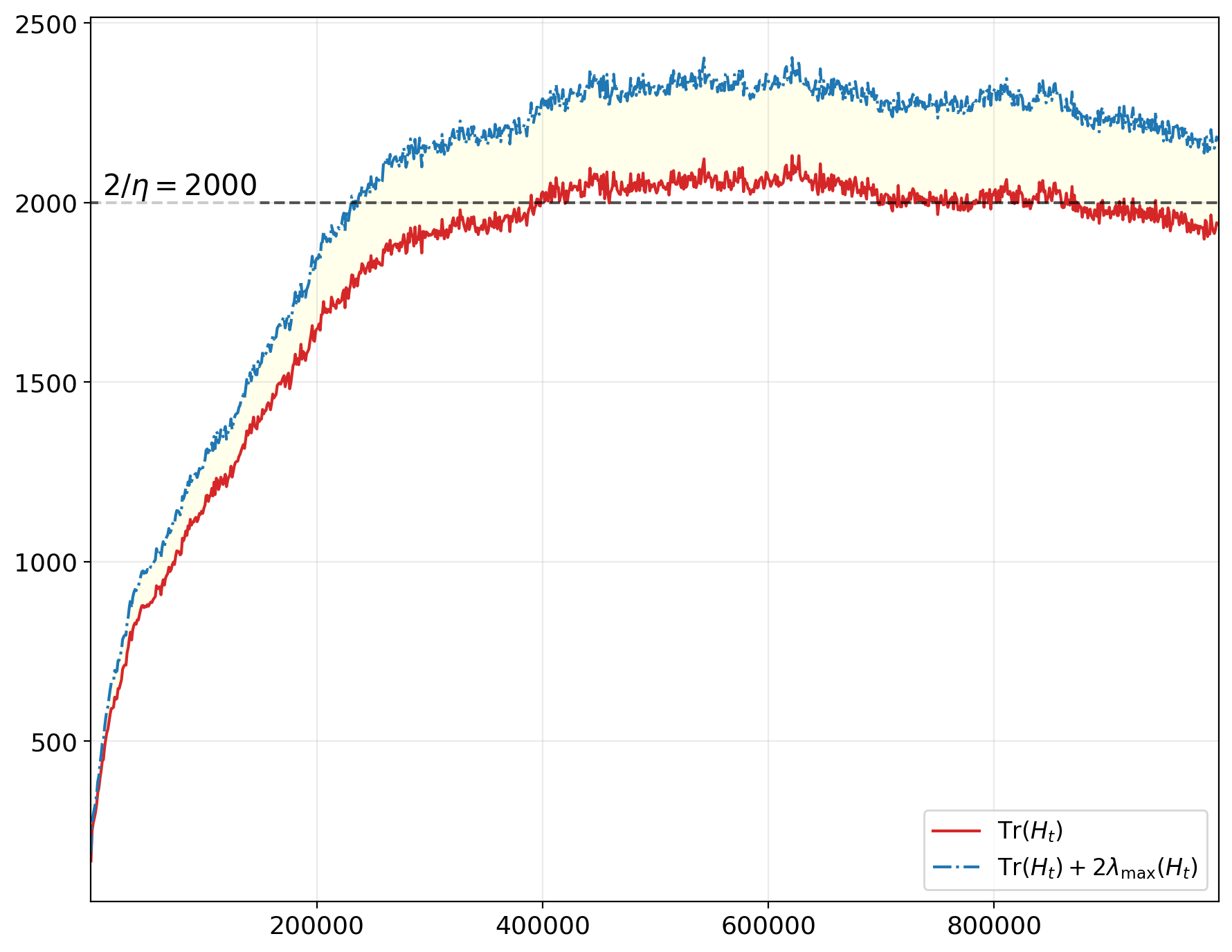}
    \put(-85,-10){iterations $t$}
\includegraphics[width=0.31\columnwidth]{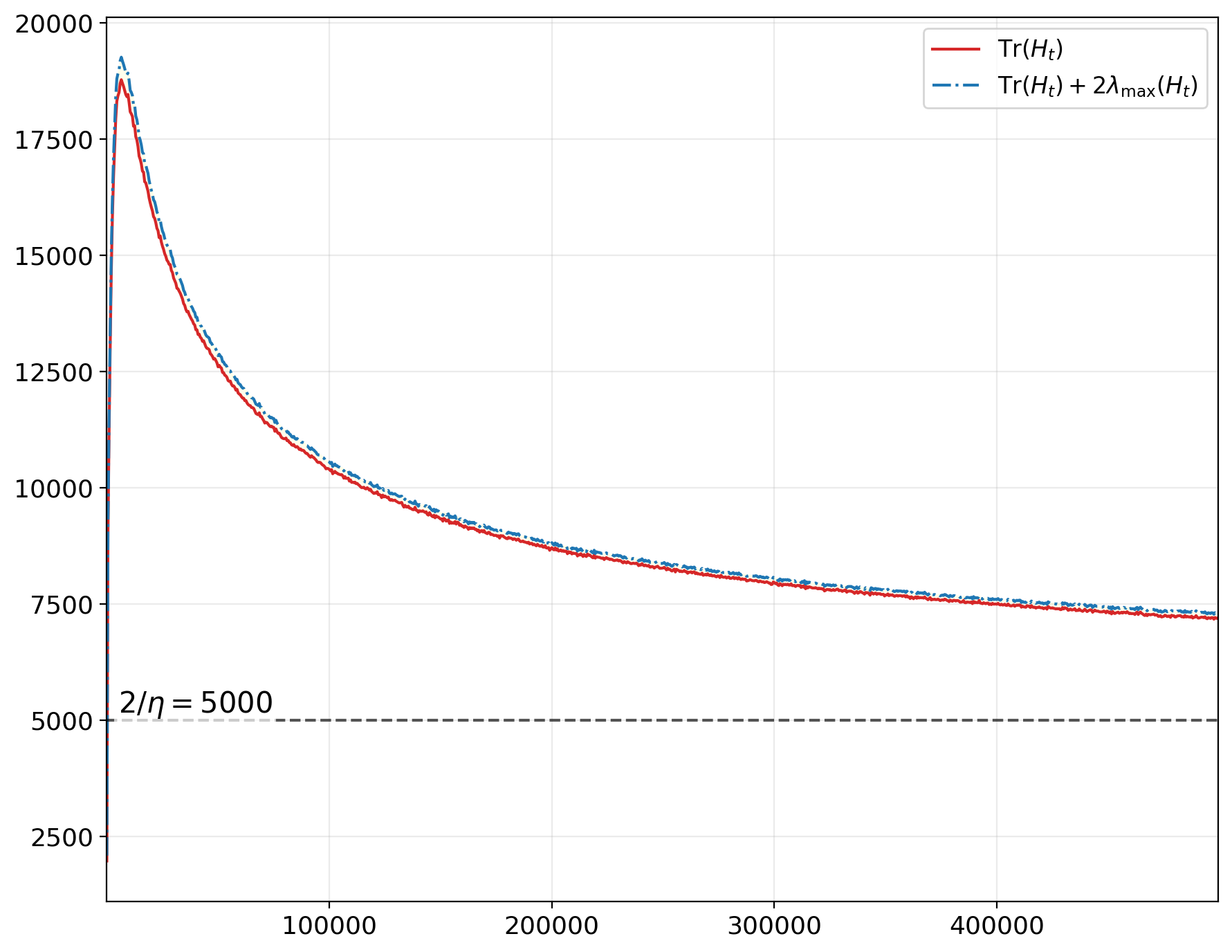}
\end{center}
    \caption{(left) We reproduce the training from random initialization on a subset of CIFAR-10 to demonstrate the EoS phenomenon reported in \cite{song2026zeroth}. (middle) We repeat the experiment shown in the left panel with per-sample clipping. (right) We reproduce the EoS phenomenon reported in the leftmost panel using ImageNet-pretrained ResNet weights.}
    \label{fig:eos}
    \vskip -0.1in
\end{figure}

When full batch gradient descent (GD) is applied to training deep neural networks, \cite{cohen2021gradient} discovered that the largest eigenvalue, \(\lambda_1(H)\), of the Hessian hovers around the threshold of linear stability, i.e., \(2/\eta\), where \(\eta\) is the step size. This is known as Edge-of-Stability (EoS). In full batch zeroth-order gradient descent, EoS happens around \({\rm Trace}(H) \leq 2/\eta \leq {\rm Trace}(H) + 2\lambda_1(H)\), as empirically shown by \citet{song2026zeroth}. We first reproduce this EoS result by training a CNN on CIFAR-10 from random initialization (Figure~\ref{fig:eos}, left), where the $2/\eta$ line is sandwiched between the predicted upper and lower bounds. With per-sample clipping, the curvature bounds become noisier and exhibit larger oscillations before eventually converging to the EoS threshold (Figure~\ref{fig:eos}, middle). However, the emergence of EoS appears to depend on the initialization regime. When the same model is initialized with ImageNet-pretrained weights, it does not reach the EoS threshold (Figure~\ref{fig:eos}, right). Hence, we should not expect EoS in our setting of fine-tuning LLMs, which is discussed in our experiments below.

\subsection{The Impact of Clipping}
We measure the Hessian trace \(\text{Tr}(H_t)\) and its upper bound \(\text{Tr}(H_t) + 2\lambda_{\max}(H_t)\) to compare random and pretrained initializations, both with and without gradient clipping.

\begin{figure}[h]
\vskip -0.1in
\begin{center}
\includegraphics[width=0.31\columnwidth]{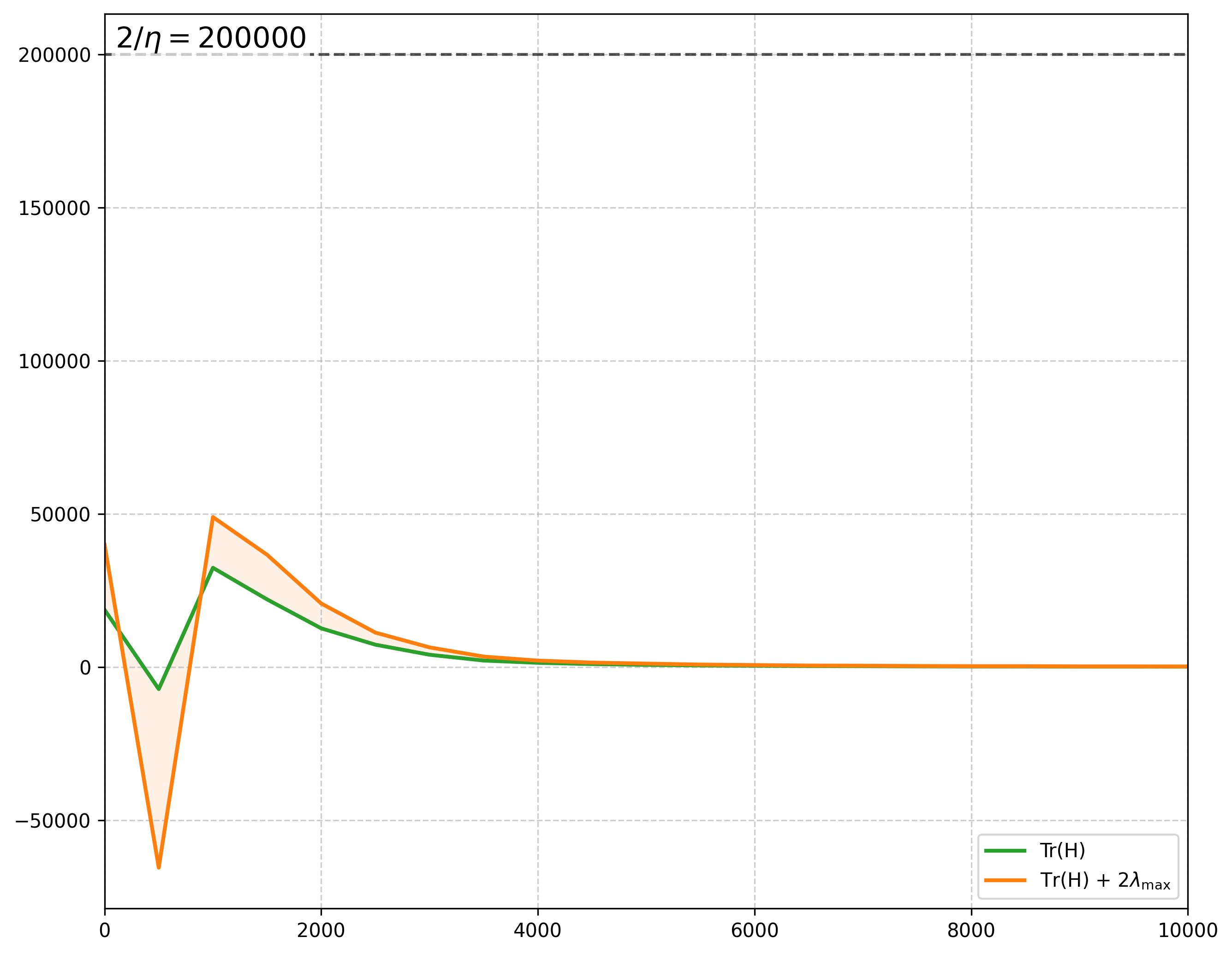}
    \put(-135,45){\rotatebox[origin=c]{90}{Hessian Trace}}
    \hspace{-0.2cm}
\includegraphics[width=0.31\columnwidth]{images/iclr/5e-4_0_hessian_plot.png}
    \put(-85,-10){Iterations $t$}
\includegraphics[width=0.31\columnwidth]{images/iclr/4e-4_0_hessian_plot_pt.png}
\end{center}
    \caption{(left) We measure standard spectral features of the Hessian matrix over training iterations in fine-tuning RoBERTa model on SST-2 dataset. Notice that there is no EoS phenomenon. (middle) We reproduce the training from random initialization on a subset of CIFAR-10 to demonstrate  EoS phenomenon as reported in \cite{song2026zeroth}. (right) We replicate EoS phenomenon as reported in \cite{song2026zeroth} but with ImageNet pretrained ResNet weights.}
    \label{fig:clip_nos}
\end{figure}

\paragraph{Without Clipping.} Without clipping, the randomly initialized CIFAR-10 model exhibits classical EoS: the trace rises and plateaus exactly at the \(2/\eta = 4000\) threshold (\ref{fig:clip_nos}, \textbf{middle}). Conversely, the pretrained ResNet does not converge to its \(2/\eta = 5000\) threshold; its trace starts near 17,500 and smoothly decays to roughly 7,500 (\ref{fig:clip_nos}, \textbf{left}). The DPZero-RoBERTa model also shows no EoS; the trace briefly increases to 30,000 before flattening to 0 (\ref{fig:clip_nos}, \textbf{right}).

\begin{figure}[h]
\vskip -0.1in
\begin{center}
\includegraphics[width=0.31\columnwidth]{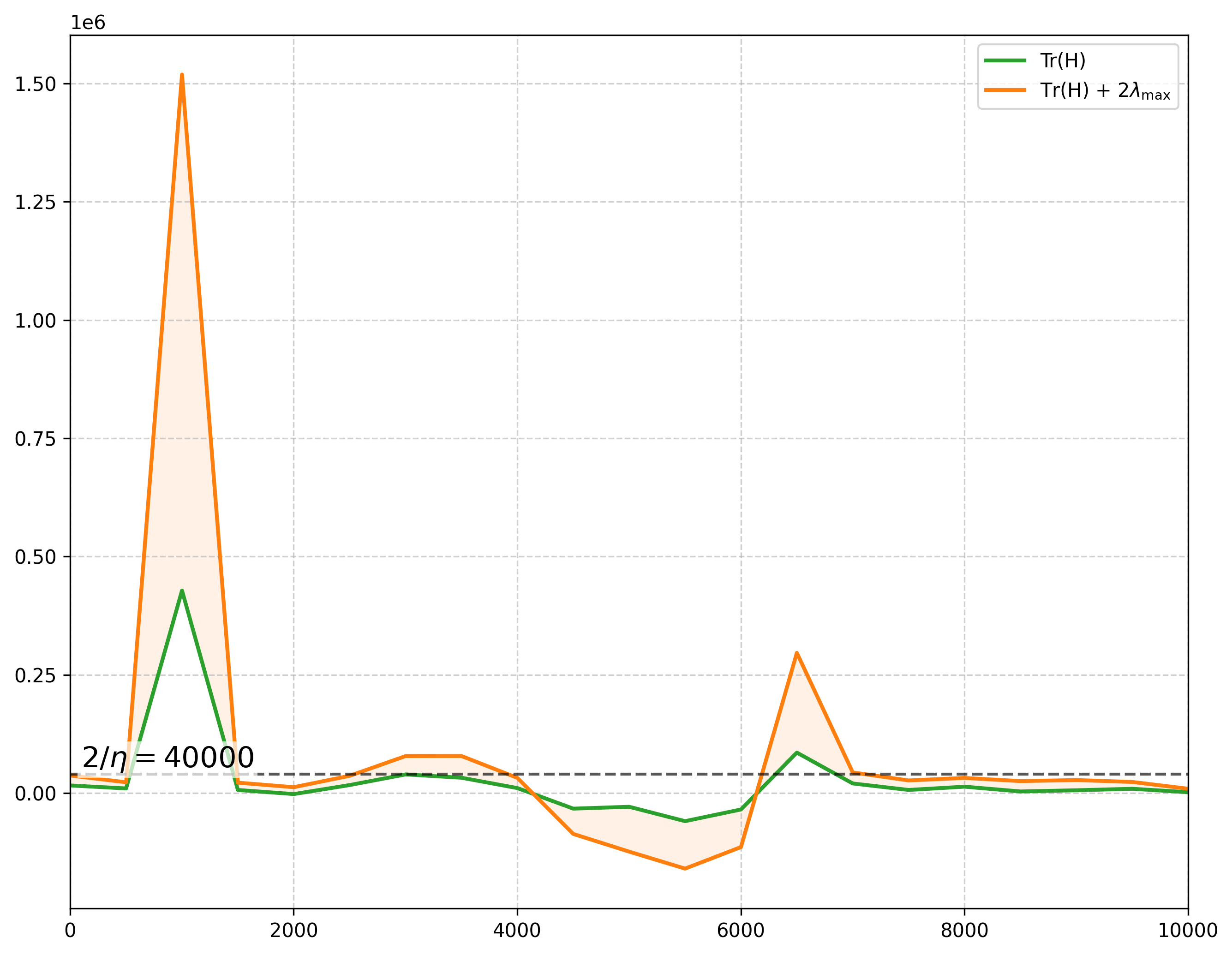}
    \put(-135,45){\rotatebox[origin=c]{90}{Hessian Trace}}
    \hspace{-0.2cm}
\includegraphics[width=0.31\columnwidth]{images/iclr/1e-3_5_hessian_plot_1M.png}
    \put(-85,-10){Iterations $t$}
\includegraphics[width=0.31\columnwidth]{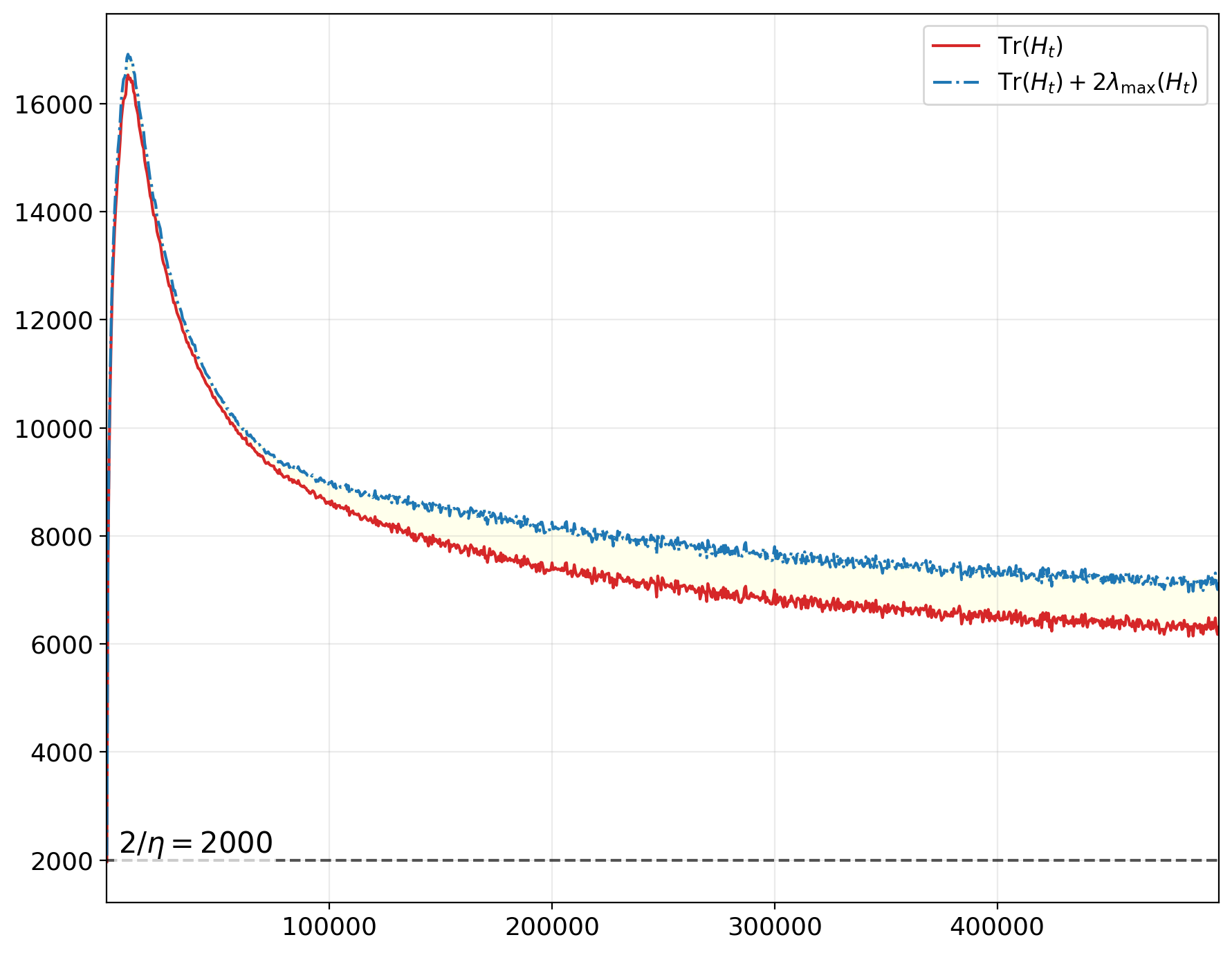}
\end{center}
    \caption{We repeat the experiments in \ref{fig:clip_nos} however we add per sample clipping. (left) We measure standard spectral features of the Hessian matrix over training iterations in fine-tuning RoBERTa model on SST-2 dataset. Notice that there is no EoS phenomenon.  (middle) We reproduce the training from random initialization on a subset of CIFAR-10 to demonstrate  EoS phenomenon as reported in \cite{song2026zeroth}. (right) We replicate EoS phenomenon as reported in \cite{song2026zeroth} but with pretrained ResNet Weights.}
    \label{fig:clip_eos}
    \vskip -0.2in
\end{figure}

\paragraph{With Clipping.} With clipping, the divergence between random and pretrained initializations remains. For the randomly initialized CIFAR-10 model, clipping preserves the EoS phenomenon (\ref{fig:clip_eos}, \textbf{middle}). The trace rises to the \(2/\eta = 2000\) threshold and oscillates along it, widening the gap between \(\operatorname{Tr}(H_t)\) and its upper bound before eventually converging at the EOS threshold. However, this convergence takes \(1{,}000{,}000\) steps in our experimental setup, compared with \(250{,}000\) steps in the setup without clipping. For the pretrained ResNet, clipping does not induce EoS (\ref{fig:clip_eos}, \textbf{right}); the trace starts above 16,000 and decays toward 6,000. For DPZero-RoBERTa, clipping increases training volatility (\ref{fig:clip_eos}, \textbf{left}). The trace misses the \(2/\eta = 40000\) threshold, exhibiting erratic spikes up to \(1.5 \times 10^6\) and negative values before converging to 0.

\subsection{Lack of EoS in Pretrained Models}
Standard EoS features curvature that rises, fluctuates near the stability threshold, and converges along \(2/\eta\). In contrast, neither pretrained model converges to this threshold during fine-tuning (\ref{fig:clip_nos} and \ref{fig:clip_eos}). We suspect this occurs because pretrained models are already situated in basins that are flat relative to the fine-tuning objective, allowing optimization to proceed without escaping the initial basin \cite{neyshabur2021transferredtransferlearning}.

\subsection{Experimental Limitations of EOS in LLM Fine-Tuning}
Finally, we note that the RoBERTa EOS evaluations exhibit greater variability than the other settings. Due to compute limits and other practical constraints, the fine-tuning trajectory was relatively volatile, making it difficult to establish a clear EoS trend. The resulting spiking behavior and absence of a clear EoS are nevertheless informative, suggesting that LLM fine-tuning with ZO-Roberta captures substantially different dynamics in this setting and motivating further investigation under more stable fine-tuning conditions.

\vskip -0.1in
\section{Additional Mismatched/Ambiguous Examples from SNLI training data}
\label{app:mismatch_examples}

Additional Mismatched/Ambiguous Examples from SNLI training data that are overcome due to the robustness of clipping.

\begin{figure}[!h]
\begin{tcolorbox}[title={Mislabeled example, ID: 174}]
{\small \textbf{Premise:} Bicyclists waiting at an intersection.\\
\textbf{Hypothesis:} Bikers stop and wait for traffic at the intersection.\\
\textbf{Original SNLI label:} Entailment \\
\textbf{Model predicted:} Contradiction\\
\textbf{Clipping stats:} 388 clips, {\bf Avg magnitude of gradient estimate:} 563.87}
\end{tcolorbox}
\vskip -0.2in
\end{figure}

\begin{figure}[!h]
\begin{tcolorbox}[title={Mislabeled example, ID: 433}]
{\small \textbf{Premise:} A bearded man in a black t-shirt sits in front of a desk holding a computer.\\
\textbf{Hypothesis:} A man waiting with his computer.\\
\textbf{Original SNLI label:} Entailment \\
\textbf{Model predicted:} Contradiction\\
\textbf{Clipping stats:} 394 clips, {\bf Avg magnitude of gradient estimate:} 482.56}
\end{tcolorbox}
\vskip -0.2in
\end{figure}

\begin{figure}[!h]
\begin{tcolorbox}[title={Mislabeled example, ID: 1147}]
{\small \textbf{Premise:} A boy in a blue, yellow, and orange shirt holding his arms out from his sides.\\
\textbf{Hypothesis:} The shirt is blue.\\
\textbf{Original SNLI label:} Contradiction\\
\textbf{Model predicted:} Neutral\\
\textbf{Clipping stats:} 384 clips, {\bf Avg magnitude of gradient estimate:} 432.83}
\end{tcolorbox}
\vskip -0.2in
\end{figure}

\begin{figure}[!h]
\begin{tcolorbox}[title={Mislabeled example, ID: 17}]
\textbf{Premise:} A few people in a restaurant setting, one of them is drinking orange juice.\\
\textbf{Hypothesis:} The diners are at a restaurant.\\
\textbf{Original SNLI Label:} Entailment\\
\textbf{Model Predicted:} Neutral\\
\textbf{Clipping Stats:} 404 clips, {\bf Avg magnitude of gradient estimate:} 816.07\\
\end{tcolorbox}
\vskip -0.2in
\end{figure}

\begin{figure}[!h]
\begin{tcolorbox}[title={Ambiguous example, ID: 1427}]
\textbf{Premise:} Two women, each embracing a little girl, catch up at a small family gathering.\\
\textbf{Hypothesis:} The women left their children behind.\\
\textbf{Original SNLI Label:} Contradiction\\
\textbf{Model Predicted:} Entailment\\
\textbf{Clipping Stats:} 399 clips, {\bf Avg magnitude of gradient estimate:} 684.28\\
\end{tcolorbox}
\vskip -0.2in
\end{figure}

\begin{figure}[!h]
\begin{tcolorbox}[title={Ambiguous example, ID: 189}]
\textbf{Premise:} People waiting at a light on bikes.\\
\textbf{Hypothesis:} The people are riding their bikes to the gym.\\
\textbf{Original SNLI Label:} Neutral\\
\textbf{Model Predicted:} Entailment\\
\textbf{Clipping Stats:} 392 clips, {\bf Avg magnitude of gradient estimate:} 759.27\\
\end{tcolorbox}
\end{figure}

\begin{figure}[!h]
\begin{tcolorbox}[title={Ambiguous example, ID: 1385}]
{\small \textbf{Premise:} Two people standing in front of a large statue of a woman, other statues and busts visible in the background.\\
\textbf{Hypothesis:} Two people standing standing near a large statue, with other states nearby.\\
\textbf{Original SNLI label:} Entailment\\
\textbf{Model predicted:} Contradiction\\
\textbf{Clipping stats:} 389 clips, {\bf Avg magnitude of gradient estimate:} 544.60}
\end{tcolorbox}
\end{figure}

\begin{figure}[!h]
\begin{tcolorbox}[title={Ambiguous example, ID: 1266}]
{\small \textbf{Premise:} Girl is blowing to a butterfly.\\
\textbf{Hypothesis:} The girl blows a butterfly.\\
\textbf{Original SNLI label:} Entailment\\
\textbf{Model predicted:} Contradiction\\
\textbf{Clipping stats:} 366 clips, {\bf Avg magnitude of gradient estimate:} 524.82}
\end{tcolorbox}
\end{figure}

\end{document}